\documentclass[journal]{IEEEtran}
\usepackage{epsfig}
\usepackage{graphicx}
\usepackage{amsmath}
\usepackage{amssymb}

\usepackage{bbm}
\usepackage{color}
\usepackage{xcolor}
\usepackage{textcomp,booktabs}
\usepackage{multirow}
\usepackage{colortbl}
\usepackage{setspace}
\usepackage{array,caption}
\usepackage{epstopdf}
\usepackage{paralist}
\usepackage{setspace}
\usepackage{latexsym,amssymb,amsmath}
\usepackage{algorithm}
\usepackage{algorithmic}
\usepackage{subcaption}
\usepackage{caption}
\usepackage{ulem}

\ifCLASSINFOpdf
\else
\fi
\begin{document}
%\title{Nearest Neighborhood-Based Self-Supervised Learning for Source Data-absent Unsupervised Domain Adaptation}
%\title{Nearest Neighborhood-Based Deep Clustering for Source Data-absent Unsupervised Domain Adaptation}
\title{Nearest Neighborhood-Based Deep Clustering for Source Data-absent Unsupervised Domain Adaptation}
%\title{Nearest Neighborhood-Regularized Deep Clustering for Source Data-absent Unsupervised Domain Adaptation}

\author{Song~Tang,~\IEEEmembership{Member,~IEEE,}
		Yan~Yang,
		Zhiyuan~Ma,~\IEEEmembership{Member,~IEEE,}
		Norman~Hendrich,
		Fanyu~Zeng,~\IEEEmembership{Member,~IEEE,}
        Shuzhi~Sam~Ge,~\IEEEmembership{Fellow,~IEEE,}
        Changshui~Zhang,~\IEEEmembership{Fellow,~IEEE,}
        and~Jianwei~Zhang,~\IEEEmembership{Member,~IEEE}% <-this % stops a space
        
\thanks{\textit{This work has been submitted to the IEEE for possible publication. Copyright may be transferred without notice, after which this version may no longer be accessible.}}        
\thanks{
This work is partly funded by the German Research Foundation and the National Natural Science Foundation of China in the Crossmodal Learning project under contract Sonderforschungsbereich Transregio 169, the Hamburg Landesforschungsf{\"o}rderungsprojekt Cross, the National Natural Science Foundation of China (61773083); Horizon2020 RISE project STEP2DYNA (691154); the National Key R\&D Program of China (2018YFE0203900, 2020YFB1313600); the National Natural Science Foundation of China (U1813202, 61773093); the Shanghai Artificial Intelligence Innovation Development Special Support Project, R \& D and Industrialization (3920365001); the Sichuan Science and Technology Program (2020YFG0476); Open Project of State Key Lab. for Novel Software Technology, Nanjing University, Nanjing, China (KFKT2021B39). \textit{(Corresponding authors: Jianwei Zhang.)}} %Important Science and Technology Innovation Projects in Chengdu (2018-YF08-00039-GX)
\thanks{Song~Tang are with the Institute of Machine Intelligence, University of Shanghai for Science and Technology, Shanghai, China; the State Key Laboratory of Electronic Thin Films and Integrated Devices, University of Electronic Science and Technology of China, Chengdu, China; the Technical Aspects of Multimodal Systems~(TAMS) Group, Department of Informatics, Universität Hamburg, Hamburg, Germany. (e-mail: tntechlab@hotmail.com)}
\thanks{Yan~Yang are with the Institute of Machine Intelligence, University of Shanghai for Science and Technology, Shanghai, China}
\thanks{Zhiyuan~Ma are with the Institute of Machine Intelligence, University of Shanghai for Science and Technology, Shanghai, China; the State Key Lab. for Novel Software Technology, Nanjing University, Nanjing, China.}
\thanks{Fanyu~Zeng is with the Engineering Research Center of Wideband Wireless Communication Technology, Ministry of Education, Nanjing University of Posts and Telecommunications, Nanjing, China.}
\thanks{Shuzhi~Sam~Ge is with the Department of Electrical and Computer Engineering, National University of Singapore, Singapore.}% ; the Centre for Robotics, University of Electronic Science and Technology of China, Chengdu, China
%Centre for Robotics, University of Electronic Science and Technology of China, Chengdu, China.
\thanks{Changshui~Zhang is with the Department of Automation, Tsinghua University, Beijing, China.}% <-this % stops a space
\thanks{Norman~Hendrich and Jianwei~Zhang are with the Technical Aspects of Multimodal Systems~(TAMS) Group, Department of Informatics, Universität Hamburg, Hamburg, Germany.}% <-this % stops a space

}

\markboth{SUBMITTED TO IEEE TRANSACTIONS ON CYBERNETICS}%
{Shell \MakeLowercase{\textit{et al.}}: Bare Demo of IEEEtran.cls for IEEE Journals}



\maketitle
\begin{abstract}
In the classic setting of unsupervised domain adaptation (UDA), the labeled source data are available in the training phase. However, in many real-world scenarios, owing to some reasons such as privacy protection and information security, the source data is inaccessible, and only a model trained on the source domain is available. 
This paper proposes a novel deep clustering method for this challenging task. Aiming at the dynamical clustering at feature-level, we introduce extra constraints hidden in the geometric structure between data to assist the process. Concretely, we propose a geometry-based constraint, named semantic consistency on the nearest neighborhood (SCNNH), and use it to encourage robust clustering. To reach this goal, we construct the nearest neighborhood for every target data
and take it as the fundamental clustering unit by building our objective on the geometry. 
Also, we develop a more SCNNH-compliant structure with an additional semantic credibility constraint, named semantic hyper-nearest neighborhood (SHNNH). After that, we extend our method to this new geometry. 
Extensive experiments on three challenging UDA datasets indicate that our method achieves state-of-the-art results. The proposed method has significant improvement on all datasets (as we adopt SHNNH, the average accuracy increases by over 3.0\% on the large-scaled dataset). Code is available at https://github.com/tntek/N2DCX.
\end{abstract}

\begin{IEEEkeywords}
Nearest neighborhood, Deep clustering, Semantic consistency, Classification, Unsupervised domain adaptation.
\end{IEEEkeywords}

% \fi
\IEEEpeerreviewmaketitle

%\vspace{-2em}
\section{Introduction}
%\IEEEPARstart{T}{his} demo file is intended to serve as a ``starter file''
%for IEEE journal papers produced under \LaTeX\ using
%IEEEtran.cls version 1.8b and later.
\IEEEPARstart{B}{eing} a branch of transfer learning~\cite{pan2009survey}, unsupervised domain adaptation (UDA)~\cite{wilson2020survey} intends to perform an accurate classification on the unlabeled test set given a labeled train set. 
In UDA, we specialize the train and test sets with different probability distributions as the source domain and target domain, respectively. 
During the transfer (training) process, we assume the labeled source data is to be accessible in the problem setting of UDA. 

The key to solving UDA is to reduce the domain drift. Because of data available from both domains, the existing methods mainly convert UDA to probability distribution matching problems, i.e., domain alignment, where the domain data represent the corresponding domain's probability distribution.
However, access to the source data is becoming extremely difficult. 
First, as the evolution of algorithms begins to wane, the performance improvements primarily rely on the increase of large-scale labeled data with high quality. 
It is hard to obtain data, deemed as a vital asset, at a low cost.  Companies or organizations may release learned models but cannot provide their customer data due to data privacy and security regulations.
Second, in many application scenarios, the source datasets (e.g., videos or high-resolution images) are becoming very large. It will often be impractical to transfer or retrain them to different platforms. 

Therefore, the so-called \textit{source data-absent UDA (SAUDA)} problem considers the scenario where only a source model pre-trained on the source domain and the unlabeled target data are available for the transfer (training) phase. Namely, we can only use the source data for the source model training. 
As most UDA methods cannot support this tough task due to their dependence on source data to perform distribution matching, this challenging topic has recently attracted a lot of research~\cite{kim2020domain,2020MA,2020shot,yang2020unsupervised,2019Distant}.

%At present, there are main two main ideas to solve SAUDA. One convets SAUDA to UDA by faking a auxiliary source domain using generative model~\cite{ma}. Another follows the clue of the self-supervised learning, deeming SAUDA as a model transfer problem from the source domain to the target domain in an unsupervised manner. For instance,    
%Xiaosong Wang, Le Lu, Hoo-Chang Shin, Lauren Kim, Mohammadhadi Bagheri, Isabella Nogues, Jianhua Yao, and Ronald M Summers. Unsupervised joint mining of deep features and image labels for large-scale radiology image categorization and scene recognition. In Winter Conference on Applications of Computer Vision (WACV), 2017.

\begin{figure}[t]
	\begin{center}
		\includegraphics[width=0.85\linewidth]{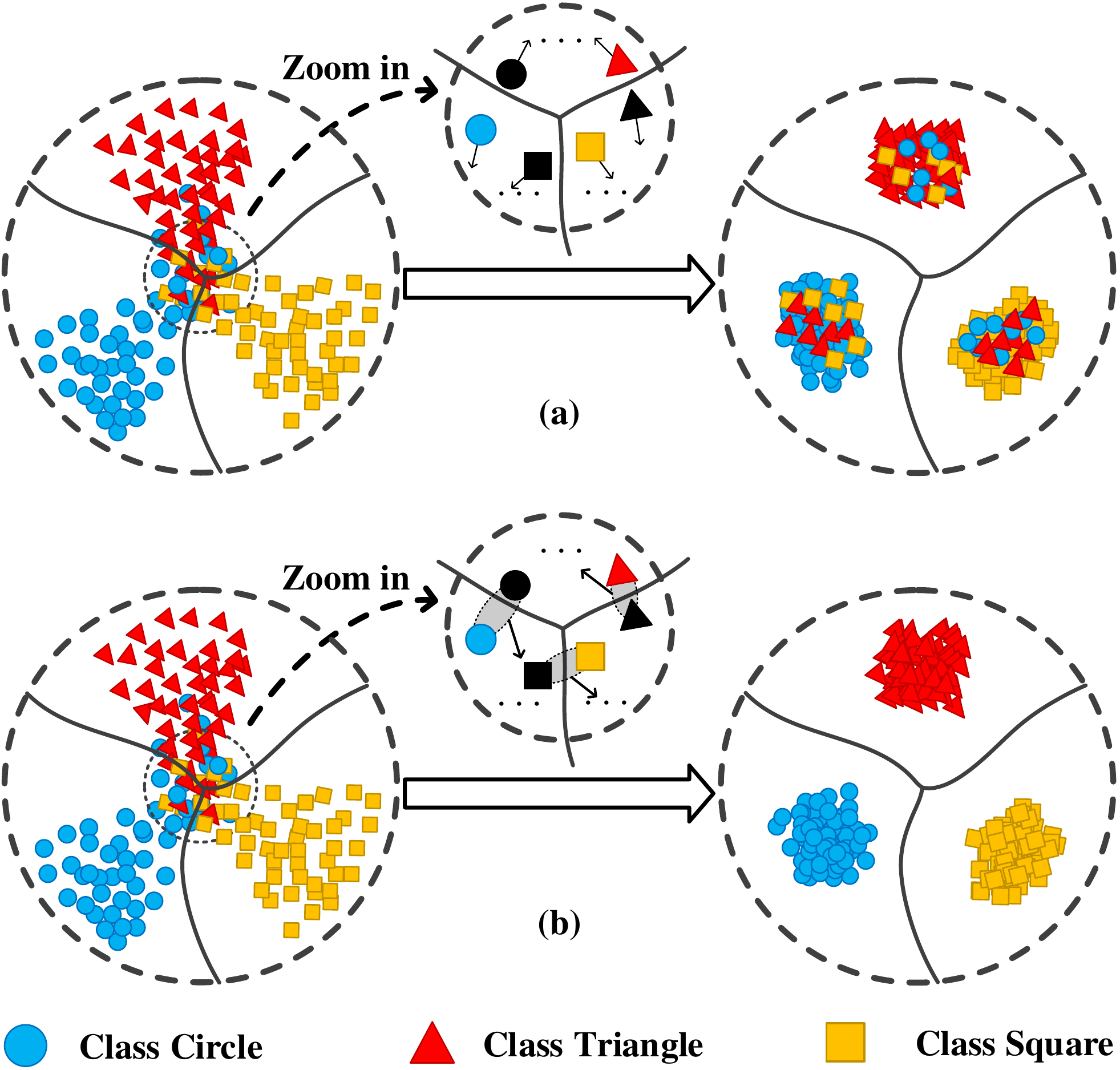}
	\end{center}
	%\vspace*{-2mm}
	\caption{Illustration of the NNH-based deep clustering. (a) and (b) present a deep clustering taking individual data and the nearest neighborhood~(NNH) of single data as fundamental clustering units, respectively.}
	\label{fig:mov}
\end{figure}

Due to its independence from given supervision information, self-supervised learning becomes a central concept for solving unsupervised learning problems. 
As an important self-supervised scheme, deep clustering has made progress in many unsupervised scenarios. %~\cite{uda,Xiaosong Wang}. 
For example, \cite{caron2018deep} developed an end-to-end general method with pseudo-labels from self-labelling for unsupervised learning of visual features. 
Very recently, \cite{2020shot} extended the work proposed in \cite{caron2018deep} to SAUDA, following the hypothesis transfer framework~\cite{ao2017effective}. 
Essentially, these methods equivalently implement a deep clustering built on individual data, as shown in Fig.~\ref{fig:mov}(a). 
Although achieving excellent results, this individual-based clustering process is susceptible to external factors, such as pseudo-label errors. 
As a result, some samples move towards a wrong cluster~(see the middle and the right subfigure in Fig.~\ref{fig:mov}(a)). 

%However, due to errors in pseudo-labels, the global geometric constraint, represented by pseudo-labels, is not sufficient to ensure robust clustering. 

Holding the perspective of robust clustering, in this paper, we intend to mine a correlation~(constraint) from the local geometry of data for more robust clustering. 
If we take the nearest neighborhood~(NNH) of individual data as the fundamental unit, the clustering may group more robustly.
As shown in the middle
subfigure of Fig.~\ref{fig:mov}(b), the black circle~(misclassified sample) moves to the wrong cluster. 
However, the gray oval (the nearest neighborhood) may move to the correct cluster with the help of an adjustment from the blue circle~(the nearest neighbor that is correctly classified).
Maintaining this trend, the gray oval, including the black circle~(misclassified sample), can eventually cross the classification plane and reach the correct cluster. 

Inspired by the idea above, this paper proposes a new deep clustering-based SAUDA method. 
The key of our method is to encourage NNH geometry, i.e., the gray oval in Fig.~\ref{fig:mov}(b), to move correctly instead of moving the individual data. 
We achieve this goal in two ways.
Firstly, we propose a new constraint on the local geometry, named semantic consistency on nearest neighborhood~(SCNNH). %: \textit{the samples in NNH should have the same semantic representation}. 
A law of cognition inspires it: the most similar objects are likely to belong to the same category, discovered by research on infant self-learning~\cite{markman2003category}.
Secondly, based on the backbone in~\cite{caron2018deep,2020shot}, we propose a novel framework to implement the clustering on NNH. 
Specifically, besides integrating a geometry constructor to build NNH for all data, we generate the pseudo-labels based on a semantic fusion on NNH. Also, we give a new SCNNH-based regularization to regulate the self-training. 

Moreover, we give an advanced version of our method. This method proposes a new, more SCNNH-compliant implementation of NNH, named semantic hyper-nearest neighborhood~(SHNNH), and extends the proposed adaptation framework to this geometry. 
The contributions of this paper cover the following three areas.
%Besides developing the method built on classic NNH constructed from spatial information, we propose a new, more SCNNH-compliant implementation of HNN, named semantic hyper-nearest neighborhood~(SHNNH), and extend our method to this geometry.
%Moreover, we propose a more SCNNH-compliant implementation of HNN, namely semantic hyper-nearest neighborhood~(SHNNH), and extend our method to this geometry.

%In this paper, we refer to this behavior as the nearest neighborhood orientation.
%The three main contributions of this paper are:

\begin{itemize}
	\item We exploit a new way to solve SAUDA. 
	We introduce the semantic constraint hidden in the local geometry of individual data, i.e., NNH, to  
	encourage robust clustering on the target domain. Correspondingly, we propose a new semantic constraint  upon NNH, i.e., SCNNH.
	\item Based on the SCNNH constraint, we propose a new deep clustering-based adaptation method
	building on NNH instead of individual data.  
	Moreover, we extend our method to a newly designed geometry, i.e., SHNNH, which expresses SCNNH more reasonably. Different from the construction strategy only based on spatial information, we additionally introduce semantic credibility constraints in the new structure.
	\item We perform extensive experiments on three challenging datasets. The results of the experiment indicate that our approach achieves state-of-the-art performance on both developed geometries. Also, except for the ablation study to explore the effect of the components in our method, a careful investigation is conducted in the analysis part. 
\end{itemize}

The remainder of this paper is organized as follows.
Section~\ref{sec:rework} introduces the related work, followed by the preliminary work as Section~\ref{sec:pre}. 
Section~\ref{sec:method} details the proposed method, while Section~\ref{sec:method-ext} extends our approach to a semantic credibility-based NNH. 
Section~\ref{sec:rlt} gives the experimental results and related analyses. 
In the end, we present the conclusion in Section~\ref{sec:con}.

\section{Related Work}\label{sec:rework}
\subsection{Unsupervised Domain Adaptation}
%The key to UDA is to reduce the domain drift. 
At present, UDA methods are widely used in scenarios such as medical image diagnosis~\cite{zhang2020collaborative}, semantic segmentation~\cite{wang2020differential}, and person re-identification~\cite{yang2020part}. 
The existing methods mainly rely on probability matching to reduce the domain drift, i.e., diminishing the probability distributions' discrepancies between the source and target domains. 
Based on whether to use deep learning algorithms, current methods can be divided into two categories.
%At present, UDA methods are widely used in scenarios such as medical image diagnosis~\cite{zhang2020collaborative}, semantic segmentation~\cite{wang2020differential}, and person re-identification~\cite{yang2020part}. 
In the first category (i.e., deep learning-based methods), researchers rely on techniques such as metric learning to reduce the domain drift~\cite{long2015learning,long2018conditional,2016Deepjan}. In these methods, an embedding space with unified probability distribution is learned by minimizing certain statistical measures % FNH: MMD not explained 
(e.g.,~maximum mean discrepancy), which are used to evaluate the discrepancy of the domains. Also, adversarial learning has been another popular framework due to its capability of aligning the probabilities of two different distributions~\cite{hoffman2018cycada, 2019Domain, 2020Multi}. 
As for the second class, the non-deep-learning methods reduce the drift in diverse manners. 
From the aspect of geometrical structure, \cite{gopalan2011domain, gong2012geodesic, caseiro2015beyond} model the transfer process from the source domain to the target one based on the manifold of data. \cite{wang2018visual} perform the transfer via manifold embedded distribution alignment. 
%Our previous work
\cite{2019visualtang} develop an energy distribution-based classifier by which the confidence target data are detected.    
% here1: Visual Domain Adaptation with Manifold Embedded Distribution Alignment
In all the aforementioned methods, the source data is indispensable because the labeled samples are used to formulate domain knowledge explicitly (e.g., probability, geometrical structure, or energy). When the labeled source domain data are not available, these traditional UDA methods fail. 

\subsection{Source Data-absent Unsupervised Domain Adaptation}
Current solutions for the SAUDA problem mainly follow three clues. 
The first one is to convert model adaptation without source data to a classic UDA setting by faking a source domain.
\cite{2020MA} incorporated a conditional generative adversarial net to explore the potential of unlabeled target data. 
The second focuses on mining transferable factors that are suitable for both domains. 
\cite{tang2019adaptive} supposed that a sample and its exemplar classifier (SVM) satisfy a certain mapping relationship. Following this idea, this method learned the mapping on the source domain and predicted the classifier for each target sample to perform an individual classification. \cite{2019Distant} used the nearest centroid classifier to represent the subspace where the target domain can be transferred from the source domain in a moderate way. 
As it features no end-to-end training, this kind of method may not work well enough in practice.
The third provides the end-to-end solution. 
This kind of method performs self-training with a pre-trained source model to bypass the absence of the source domain and the label information of the target domain. 
\cite{2020shot} developed a general end-to-end method following deep clustering and a hypothesis transfer framework to implement an implicit alignment from the target data to the probability distribution of the source domain. 
In the method, information maximization (IM)~\cite{krause2010discriminative} and pseudo-labels, generated by self-labelling, were used to supervise the self-training.
\cite{yang2020unsupervised} canceled the self-labelling operation and newly added a classifier to offer the semantic guidance for the right move. 
These two methods obtained outstanding results, however, they ignored the fact that the geometric structure between data can provide meaningful context. 

\subsection{Deep Clustering}
%\subsection{Hypothesis Transfer Learning}

%Among all the transfer learning frameworks, Hypothesis Transfer Learning (HTL) seems to be the most reasonable and realistic way to solve the above problem~\cite{perrot2015theoretical}. In HTL, the target model has no direct access to the source domain. Instead, it operates on the source hypothesis~\cite{kuzborskij2013stability}. However, some traditional HTL methods require some labeled target samples or multiple source domain hypotheses~\cite{ao2017effective, mansour2009domain}, and others focus on a single source domain and modify models from feature level~\cite{chidlovskii2016domain, 2019Distant}. Besides, some researchers like~\cite{chidlovskii2016domain} incorporated feature corruption and marginalization in both supervised and unsupervised settings. Most recently, Liang et al.~\cite{2019Distant} used the nearest centroid classifier to represent the subspace where the target domain can be transferred from the source domain in a moderate way. Although the methods proved to be effective, the training is not end-to-end, which may not work well enough in practice.

Deep clustering (DC)~\cite{bojanowski2017unsupervised,liao2016learning,caron2019unsupervised,xie2016unsupervised} performs deep network learning together with discovering the data labels unlike conventional clustering performed on fixed features~\cite{zhan2018consensus,yang2019learning}. 
Essentially, it is a process of simultaneous clustering and representation learning.
Combining cross-entropy minimization and a k-means clustering algorithm, the recent DeepCluster~\cite{caron2018deep} method first proposed a simple but effective implementation.
Most recently, ~\cite{asano2019self} and ~\cite{chen2020unsupervised} boosted the framework developed by DeepCluster.
~\cite{asano2019self} introduced a data equipartition constraint to address the problem of all data points mapped to the same cluster.
~\cite{chen2020unsupervised} provided a concise form without k-means-based pseudo-label where the augmentation data's logits are used as the self-supervision.
Besides the unsupervised learning problem on the vast dataset, such as ImageNet,  various attempts have been made to solve the UDA problem.
~\cite{kang2019contrastive} leveraged spherical k-means clustering to improve the feature alignment. The work in~\cite{tang2020unsupervised} introduced an auxiliary counterpart to uncover the intrinsic discrimination among target data to minimize the KL divergence between the introduced one and the predictive label distribution of the network. 
For SAUDA, DC also achieved excellent results, for example, \cite{2020shot} and \cite{yang2020unsupervised} that we review in the last part.
%Similar to the existing SAUDA methods aforementioned, 
All of these methods mentioned above only focus on learning the representation of data from single samples. The extra constraints hidden in the geometric structure between data were not well exploited.

\section{Preliminary}\label{sec:pre}
\subsection{Problem Formulation}\label{sec:pf}
Given two domains with different probability distributions, i.e., source domain $\mathcal{S}$ and target domain $\mathcal{T}$, 
where $\mathcal{S}$ contains $n_s$ labeled samples while $\mathcal{T}$ has $n_t$ unlabeled data. Both labeled and unlabeled samples share the same $K$ categories. Let 
$\mathcal{X}_s=\{\boldsymbol{x}_{i}^s\}_{i=1}^{n_s}$ and $\mathcal{Y}_s=\{{y}_{i}^s\}_{i=1}^{n_s}$ be the source samples and their labels where $y_i^s$ is the label of $\boldsymbol{x}_i^s$. Let $\mathcal{X}_t=\{{\boldsymbol{x}_{i}^t\}_{i=1}^{n_t}}$ and $\mathcal{Y}_t=\{{y}_{i}^t\}_{i=1}^{n_t}$ be the target samples and their labels. 

Traditional UDA intends to conduct a $K$-way classification on the target domain with the labeled source data and the unlabeled target data. In contrast, SAUDA tries to build a target function~(model) $f_t: \mathcal{X}_t \to \mathcal{Y}_t$ for the classification task, while only $\mathcal{X}_t$ and a pre-obtained source function~(model) $f_s: \mathcal{X}_s \to \mathcal{Y}_s$ are available.
\begin{figure*}[t]
	\begin{center}
		\includegraphics[width=7.0in,angle=0]{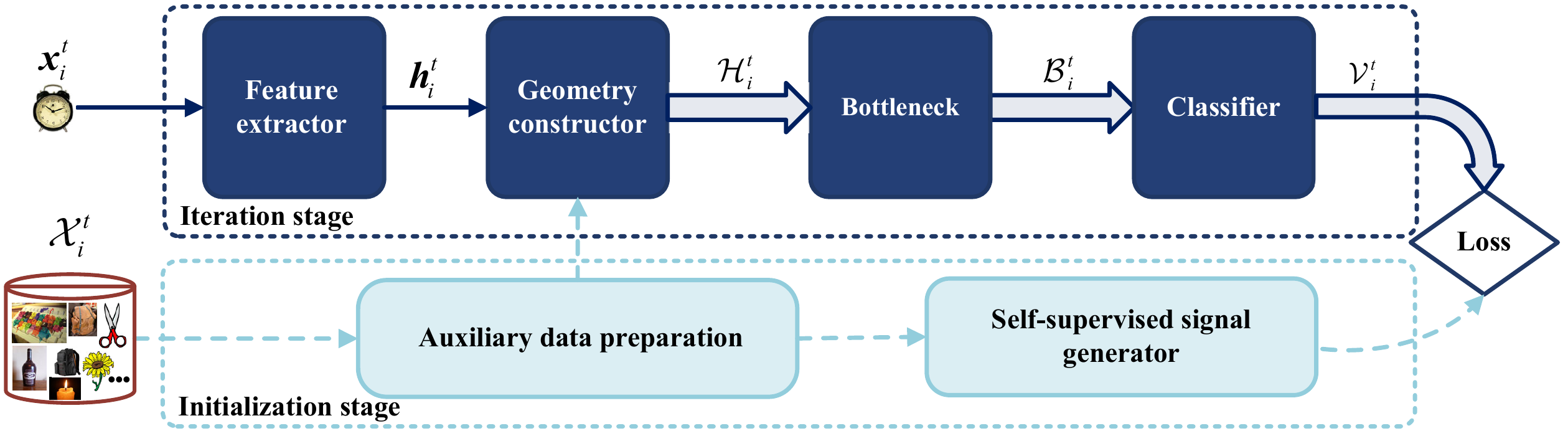}
	\end{center}
	%\vspace{-5mm}
	\caption{Our pipeline for model adaptation. 
		\textbf{Top}: This branch contains the target model with a geometry construction module, i.e., geometry constructor, and works in the iteration stage of an epoch. 
		\textbf{Bottom}: This branch includes the two supporting modules working in the initialization stage. To assist the self-training in the target model, the generated auxiliary data help the geometry building and the self-supervised signal generation. And the generated self-supervised signal offers semantic regulation.} 
	\label{fig:fw}
\end{figure*}

\subsection{Semantic Consistency on Nearest Neighborhood (SCNNH)}\label{sec:scnnh}
Much work has shown that the geometric structure between data is beneficial for unsupervised learning. An example of this are the pseudo-labels generated by clustering with global geometric information. 
The local geometry and semantics (class information) of data are closely related. 
Self-learning is known as an important way for babies to gain knowledge through experience. 
Research has found that babies will use a simple strategy, named category learning~\cite{ashby2005human,deng2015development}, to supervise their self-learning. 
Specifically, babies tend to classify a new object into the category that its most similar object belongs. 
When the semantics of the most similar object is reliable, babies can correctly identify the new object by this strategy. 

Inspired by the cognition mechanism introduced above, we propose a new semantic constraint, named semantic consistency on the nearest neighborhood~(SCNNH): \textit{the samples in NNH should have the same semantic representation as close to the true category as possible}. 
The constraint can promote the robust clustering pursued by this paper in two folds: \textbf{(\romannumeral1)} the constraint is confined to the local geometry of NNH that helps us to carry out clustering taking NNH as the basic clustering unit, and \textbf{(\romannumeral2)} the consistency makes the NNH samples all move to the same cluster center . 
%The experiment results confirm our expectations.
To implement the SCNNH constraint, we should address two essentials. One is to construct proper NNH, while another is to model the semantic consistency.  Focusing on these two issues, we develop the adaptation method, as shown in the next section.

\section{Methodology}\label{sec:method}
In this section, we introduce the framework of the proposed method, followed by the details of the modules in the framework and the regularization method.

\subsection{Model Adaptation Framework}
According to the manner of a hypothesis transfer framework, our solution for SAUDA consists of two phases. 
The first one is the pre-training phase to train the source model, and the second is the adaptation phase to transfer the obtained source model to the target domain.
%\IEEEpubidadjcol

\textbf{Pre-training Phase}. 
We take a deep network as the source model. Specifically, we parameterize a feature extractor $u_s\left(\cdot;\theta_s\right)$, a bottleneck $g_s\left(\cdot;\phi_s\right)$, and a classifier $c_s\left(\cdot;\psi_s\right)$ as the network where $\{\theta_s,\phi_s,\psi_s\}$ collects the network parameters. We also use $c_s \circ g_s \circ u_s$ to denote the source model $f_s$. For input instance ${\boldsymbol{x}}_{i}^{s}$, this network finally outputs a probability vector $\boldsymbol{p}_i^s=\mathrm{softmax}(c_s \circ g_s \circ u_s({\boldsymbol{x}}_{i}^{s})) \in [0,1]$.

In this model, the feature extractor is a deep architecture initiated by a pre-trained deep model, for example, ResNet~\cite{he2016deep}.   
The bottleneck consists of a batch-normalization layer and a fully-connect layer, while the classifier comprises a weight-normalization layer and a fully-connect layer 
%in which $K$ differs from one dataset to another. 
%with a size of $2048\times 256$,with a size of $256\times K$ 
We train the source model via optimizing the objective as follows. 
\begin{equation}
	\label{equ:opt_source}
	\mathop{\min}_{\{\theta_s,\phi_s,\psi_s\}}\mathcal{L}_{c_s\circ g_s\circ u_s}^{s} =  -\frac{1}{n_s}\sum_{i=1}^{n_s}\sum_{k=1}^{K}\bar{\boldsymbol{l}}_{i,k}^s\log \boldsymbol{p}_{i,k}^s.
\end{equation}
In Eqn.~\eqref{equ:opt_source}, $\boldsymbol{p}_{i,k}^s$ is the $k$-th element of $\boldsymbol{p}_{i}^s$; $\bar{\boldsymbol{l}}_{i,k}^s$ is the $k$-th element of  ${\bar{\boldsymbol{l}}}_i^s=(1-\gamma)~\boldsymbol{l}_i^s + \gamma/K$, i.e., the smooth label \cite{muller2019does}, where $\boldsymbol{l}_i^s$ is a one-hot encoding of label $y_i^s$.

\textbf{Adaptation Phase}. 
In this phase, we learn $f_t$ by self-training a target model in an epoch-wise manner. Fig.~\ref{fig:fw} presents the pipeline for this self-training.
The target model has a similar structure as the source model except for a newly introduced module named geometry constructor. 
As shown at the top of Fig.~\ref{fig:fw}, the target model includes four modules. They are \textbf{(\romannumeral1)} a deep feature extractor $u_t(\cdot;\theta_t)$, \textbf{(\romannumeral2)} the geometry constructor, \textbf{(\romannumeral3)} a bottleneck $g_t(\cdot;\phi_t)$, and \textbf{(\romannumeral4)} a classifier $c_t(\cdot;\psi_t)$ where $\{\theta_t,\phi_t,\psi_t\}$ are the model parameters. 
We also write them as $u_t(\cdot)$, $g_t(\cdot)$ and $c_t(\cdot)$ for simplicity and use $c_t \circ g_t \circ u_t$ to denote the target model $f_t$. 
To implement the self-training and geometry building, as shown at the bottom of Fig.~\ref{fig:fw}, we propose another branch, including an auxiliary data preparation module and a self-supervised signal generator.
\begin{algorithm}[h]
	\caption{Pseudo-code of the target model training.}
	\label{alg:traing}
	\vspace{2pt}
	\textbf{Input}: Pre-trained source model $f_s(\theta_s,\phi_s,\psi_s)$; target samples $\mathcal{X}_t$; max epoch number $T_m$; max iteration number $N_b$ of each epoch. \\%trade-off parameters $\alpha, \beta$.\\
	\textbf{Initialization}: Initialize $\{\theta_t,\phi_t,\psi_t\}$ using $\{\theta_s,\phi_s,\psi_s\}$. \\
	\vspace{-1.2em}
	\begin{algorithmic}[1]
		%\STATE Use $f_s$ to initialize the target model $f_t$ where the end classifier $h_t = h_s$ is fixed.
		\FOR{Epoch = 1 to ${T_m}$} 
		%\STATE Obtain confident group using the \textit{confident group splitter in HNN generator} formulated by Eq.\ref{eqn:ce} $\sim$ Eq.\ref{eqn:c}.
		%\STATE Generate deep features of $\mathcal{X}_t$ by $u_t$ to facilitate the NNH generation.
		%\STATE Generate low-dimensional features of $\mathcal{X}_t$ by $g_t \circ u_t$ to facilitate the pseudo-labels generation.
		\STATE Generate auxiliary data by the module of auxiliary data preparation.
		\STATE Generate semantic-fused pseudo-labels for $\mathcal{X}_t$ by the self-supervised signal generator.
		\FOR{Iter = 1 to ${N_b}$}
		\STATE Sample a batch and get their pseudo-labels.
		\STATE Construct dynamical NNH for the data in this batch by the geometry constructor. %formulated by Eqn.~\ref{eqn:samslf}$\sim$\ref{eq:mix}.
		\STATE Update model parameters $\{\theta_t,\phi_t,\psi_t\}$ by optimizing the SCNNH-based objective. %objective formulated by  Eqn.~\ref{eq:objctive}.
		\ENDFOR 
		\ENDFOR
	\end{algorithmic}
\end{algorithm}

%For each input instance $\boldsymbol{x}_i^t$, when it goes through the target model, we will get a serial of module outputs in turn. 
The modules mentioned above work in two parallel stages of a training epoch.     
In the initialization stage, the self-supervised signal generator conducts self-labelling to output semantic-fused pseudo-labels for all target data. The auxiliary data preparation outputs data to facilitate the other two modules, including the geometry constructor and the self-supervised signal generator. 
As for the iteration stage, these modules in the target model transform input data from pixel space to the logits space, as shown at the top of Fig.~\ref{fig:fw}.
%Specifically, for each input instance $\boldsymbol{x}_i^t$, the target model maps it from pixel space to a semantic space.
Firstly, the deep feature extractor transforms $\boldsymbol{x}_i^t$ into a deep feature ${\boldsymbol{h}}_i^t$. 
Secondly, the geometry constructor builds NNH for all data at the deep feature space. 
Using this module, we switch the clustering unit from individual data ${\boldsymbol{h}}_i^t$ to its NNH. 
We use $\mathcal{H}_i^t=\{{\boldsymbol{h}}_i^t, {\boldsymbol{h}}_{in}^t\}$ to denote NNH where ${\boldsymbol{h}}_{in}^t$ is the nearest neighbor of ${\boldsymbol{h}}_i^t$. Thirdly, the bottleneck maps the constructed NNH $\mathcal{H}_i^t$ into a low-dimensional feature space. 
We use $\mathcal{B}_i^t=\{{\boldsymbol{b}}_i^t, {\boldsymbol{b}}_{in}^t\}=\{g_t\left({\boldsymbol{h}}_{i}^t\right), g_t\left({\boldsymbol{h}}_{in}^t\right)\}$ to represent the output of $g_t$. 
%where ${\boldsymbol{b}}_{i}^t=g_t\left({\boldsymbol{z}}_{i}^t\right)$ and ${\boldsymbol{b}}_{in}^t=g_t\left({\boldsymbol{z}}_{in}^t\right)$. 
Finally, the classifier further maps $\mathcal{B}_i^t$ into a final semantic space. We use $\mathcal{V}_i^t=\{{\boldsymbol{v}}_i^t, {\boldsymbol{v}}_{in}^t\}=\{c_t\left({\boldsymbol{b}}_{i}^t\right), c_t\left({\boldsymbol{b}}_{in}^t\right)\}$ to represent the ending output.
%where ${\boldsymbol{h}}_{i}^t=c_t\left({\boldsymbol{b}}_{i}^t\right)$ and ${\boldsymbol{h}}_{in}^t=c_t\left({\boldsymbol{b}}_{in}^t\right)$.

Alg.~\ref{alg:traing} summarizes the training process of the target model $f_t$.
Before training, we initialize $f_t$ using the pre-trained source model $f_s$. 
Specifically, we use $u_s$, $g_s$ and $c_s$ in $f_s$ to initialize the corresponding parts, i.e., $u_t$, $g_t$ and $c_t$, of the target model, respectively. 
Subsequently, we freeze the parameters of $c_t$ during the succeeding training. 
After completing the model initialization, we start the self-training, which runs in an epoch-wise manner.

\textbf{Note:} In the inference time, we do not need to keep the full training set in memory to construct and update the local neighborhood for every incoming input. We simply obtain the category prediction by passing the input data through the three network modules, i.e., $u_t$, $g_t$ and $c_t$.  

%To train the target model in the deep clustering's manner, we also perform a epoch-wise strategy.  

%Regarding the self-training, we slice into $T_m$ epochs where each epoch consists of two stages.The first stage, presented in line 1$\sim$4, is to generate the auxiliary information and pseudo-labels. In contrast, the second stage performs end-to-end training, as presented in line 5$\sim$9. 
%in the initialization stage of an upcoming epoch. 
%These pre-obtained auxiliary information is computed by the trained model from lastest epoch and fixed during the whole upcoming epoch. 
%\textbf{Note that} these pre-obtained facilitating information is computed by the trained model from lastest epoch and fixed during the whole upcoming epoch. 
%In Alg.~\ref{alg:traing}, we summarize the whole training process of N2O Net where the NNH generation (perfomed by NNCDescriptor) and the pseudo-labels generation (performed by the cluster block) are the key components. In the following, we will first detail the two modules, and then give the details of the objective for self-training.
%\vspace{-0pt}

\subsection{Auxiliary Data Preparation}
%For the all target data $\{{\mathbf{x}}_{j}^t\}_{j=1}^{n}$, when they go through the taraget model, we can obtain two intermediate features, namely the deep features $\{{\mathbf{z}}_{j}^{t'}\}_{j=1}^{n}$ 
%and low-dimensional features $\{{\mathbf{b}}_{j}^{t'}\}_{j=1}^{n}$. We takes these features as the auxiliary information and maintain tehm during the whole new epoch. 
%For an input instance x, the network computes feature representation z = ϕ(x), and outputs a probability vector p = softmax(f(z)) ∈ [0, 1]K after the final softmax operation.
At the beginning of an epoch, we prepare the auxiliary data by making all target data go through the target model. 
For all target data $\{{\boldsymbol{x}}_{i}^t\}_{i=1}^{n_t}$, the target model first outputs the deep features $\{{\boldsymbol{h}}_{i}^{t}\}_{i=1}^{n_t}$, then outputs the low-dimensional features $\{{\boldsymbol{b}}_{i}^{t}\}_{i=1}^{n_t}$, and finally outputs the logits features $\{{\boldsymbol{v}}_{i}^{t}\}_{i=1}^{n_t}$. 
Considering that the obtained information is frozen during the upcoming epoch, we rewrite these features as $\{\bar{\boldsymbol{h}}_{i}^t\}_{i=1}^{n_t}$, $\{\bar{\boldsymbol{b}}_{i}^{t}\}_{i=1}^{n_t}$ and $\{\bar{\boldsymbol{v}}_{i}^{t}\}_{i=1}^{n_t}$ to indicate this distinction. For simplicity, we also collectively write them as $\bar{\boldsymbol{H}}_t$, $\bar{\boldsymbol{B}}_t$, and $\bar{\boldsymbol{V}}_t$, respectively.

%when they go through the target model, we obtain two kinds of intermediate features in turn. One is the deep features $\{\bar{\mathbf{z}}_{i}^{t}\}_{i=1}^{n}$ that we collectively write as $\bar{\mathbf{Z}}_t$. The other is the low-dimensional features $\{\bar{\mathbf{b}}_{i}^{t}\}_{i=1}^{n}$ that we collectively write as $\bar{\mathbf{B}}_t$. We take these features as auxiliary information and maintain them during the whole new epoch. 
%the target model outputs the deep features $\{{\mathbf{z}}_{j}^{t'}\}_{j=1}^{n}$ 
%and low-dimensional features $\{{\mathbf{b}}_{j}^{t'}\}_{j=1}^{n}$. 
%Our method takes the features of all target data, outputted by the deep feature extractor and the bottleneck, as the auxiliary information.  
%generated by the trained intermediate model from the latest epoch. 
%Auxiliary information refers to the intermedia features of all target data outputted by the deep feature extractor and the botteleneck. We generate these feature using the trained model from the latest epoch. 

\subsection{Semantic-Fused Self-labelling}\label{sec:pl}
The semantic-fused self-labelling is to output the semantic-fused pseudo-labels.
We adopt two steps to generate this kind of pseudo-labels including 1) NNH construction and 2) pseudo-labels generation. 

\textbf{Static NNH construction}.
%This step constructs the NNH based on  $\bar{\boldsymbol{H}}_t$. 
In this step, we use a similarity-comparison-based method over the pre-computed deep features $\bar{\boldsymbol{H}}_t$ to find the nearest neighbor of $\bar{\boldsymbol{h}}_i^t$ and build the NNH.
Because this geometry building is only executed once in the initiation stage, we term this construction as the static NNH construction.

Without loss of generality, for given data $\boldsymbol{u}$, we use $\boldsymbol{u}_{in}=\mathrm{F}(\boldsymbol{u}; {\boldsymbol{U}})$ to represent its nearest neighbor detected in set ${\boldsymbol{U}}$. We make this function $\mathrm{F}(\cdot;\cdot)$ equivalent to an optimization problem formulated by Eqn.~\eqref{eqn:nnh-pl} where $\mathrm{D}_{sim}(\cdot,\cdot)$ is the cosine distance function computing the similarity of two vectors. 
%Alg.~\ref{alg:cons-nnh} provides the details.
\begin{equation}
	\label{eqn:nnh-pl}
	\begin{split}
		{\boldsymbol{u}}_{in}&={\boldsymbol{u}}_{{i^{'}}}, ~~{i^{'}}= \arg \min \limits_{i} \mathrm{D}_{sim}({\boldsymbol{u}}, {\boldsymbol{u}}_{i}),\\ 
		s.t.&~~i=1, 2, \cdots, |{\boldsymbol{U}}|;~{\boldsymbol{u}} \neq {\boldsymbol{u}}_{{i}};~{\boldsymbol{u}}_{{i}}, {\boldsymbol{u}}_{i^{'}} \in {\boldsymbol{U}}.\\
	\end{split}
\end{equation}
%In Eqn.~\eqref{eqn:nnh-pl}, . In this paper, we adopt the cosine-distance to measure the similarity.

Thus, we obtain $\bar{\boldsymbol{h}}_i^t$'s nearest neighbor $\bar{\boldsymbol{h}}_{in}^t = \mathrm{F}(\bar{\boldsymbol{h}}_i^t; \bar{\boldsymbol{H}}_t)$ using the method represented by Eqn.~\eqref{eqn:nnh-pl} and build the NNH on $\bar{\boldsymbol{H}}_t$, denoted by $\bar{\mathcal{H}}_i^t=\{\bar{\boldsymbol{h}}_i^t, \bar{\boldsymbol{h}}_{in}^t\}$.

\textbf{Pseudo-label generation}.
This step performs a semantic fusion on $\bar{\mathcal{H}}_i^t$. To facilitate the semantic fusion, we firstly give the method to compute a similarity-based logits of data using the pre-obtained features $\bar{\boldsymbol{B}}_t$ and $\bar{\boldsymbol{V}}_t$. We arrive at the similarity-based logits $\bar{\boldsymbol{q}}_{i}^{t}$ of any target data ${\boldsymbol{x}}_{i}^{t}$ by the following computation.
\begin{equation}
	\label{eqn:logits}
	\bar{\boldsymbol{q}}_{i}^{t}=\frac{1}{2}\left( 1+\frac{\bar{\boldsymbol{b}}_i^{t\top}\boldsymbol{\mu}_k}{\Vert\bar{\boldsymbol{b}}_i^{t}\Vert  \Vert\boldsymbol{\mu}_k \Vert} \right) \in [0,1],~~\bar{\boldsymbol{b}}^t_i \in \bar{\boldsymbol{B}}_t.
\end{equation}

In Eqn.~\eqref{eqn:logits}, we perform a weighted k-means clustering over $\bar{\boldsymbol{B}}_t$  to get $\boldsymbol{\mu}_k$, the $k$-th cluster centroid.  
Let $\{\bar{\boldsymbol{p}}_{i}^{t}\}_{i=1}^{n_t}$ be the probability vectors of $\{\bar{\boldsymbol{v}}_{i}^{t}\}_{i=1}^{n_t}$, i.e., $\bar{\boldsymbol{V}}_t$, after a softmax operation.
The computation of $\boldsymbol{\mu}_k$ is expressed to the equation~\eqref{eqn:ok} where $\bar{{p}}_{i,k}$ is the $k$-th element of $\bar{\boldsymbol{p}}_{i}$.
\begin{equation}
	\label{eqn:ok}
	\boldsymbol{\mu}_k= \frac{\sum_{i=1}^{n}\bar{{p}}_{i,k}^{t}\bar{\boldsymbol{b}}_i^t}{\sum_{i=1}^{n}\bar{{p}}_{i,k}^{t}},~~\bar{\boldsymbol{b}}^t_i \in \bar{\boldsymbol{B}}_t.
\end{equation}

It is known that $\bar{\boldsymbol{b}}_i^t$ and $\bar{\boldsymbol{h}}_i^t$ satisfy the mapping of $g_t$, such that, using the method introduced above, we can get $\bar{\boldsymbol{Q}}_i^t=\{\bar{\boldsymbol{q}}_{i}^{t}, \bar{\boldsymbol{q}}_{in}^{t}\}$ that is the similarity-based logits of $\bar{\mathcal{H}}_i^t$. %$\{\bar{\boldsymbol{h}}_i^t, \bar{\boldsymbol{h}}_{in}^t\}$.  
Based on $\bar{\boldsymbol{Q}}_i^t$, we perform a dynamical fusion-based method to generate the pseudo-label. 
Assigning $\bar{y}_i^t$ as the pseudo-label of target data ${\boldsymbol{x}}_{i}^{t}$, we get it by optimizing the objective as follows. 
\begin{equation}
	\label{eqn:fusion-pl}
	\begin{split}
	\min \limits_{k} {\mathrm{M}_k}\left(\bar{\boldsymbol{Q}}_i^t\right) &= \lambda_k\bar{{q}}_{i,k}^{t} + \left(1-\lambda_k\right)\bar{{q}}_{in,k}^{t},\\
	s.t.~k&=1, 2, \cdots, K,
	\end{split}
\end{equation}   
where ${\mathrm{M}_k}\left(\cdot\right)$ stands for the $k$-th element of function ${\mathrm{M}}\left(\cdot\right)$ output, $\bar{{q}}_{i,k}^{t}$ and $\bar{{q}}_{in,k}^{t}$ are the $k$-th element of $\bar{\boldsymbol{q}}_{i}^{t}$ and $\bar{\boldsymbol{q}}_{in}^{t}$ respectively, $\lambda_k$ is the $k$-th element of random vector $\lambda\sim \mathrm{N}\left(\alpha,\delta\right)$, in which $\delta=1-\alpha$.
During the consecutive iteration stage, we fix these pseudo-labels and take them as a self-supervised signal to regulate the self-training.

\subsection{Dynamical NNH Construction}\label{sec:dymmh}

In the iteration stage of an epoch, we dynamically build NNH for every input instance based on the pre-computed deep feature $\bar{\boldsymbol{H}}_t$. Our idea also is to construct NNH by detecting the nearest neighbor of the input instance from $\bar{\boldsymbol{H}}_t$. For an input instance $\boldsymbol{x}_i^t$ whose deep feature is ${\boldsymbol{h}}_i^t$, we form its NNH by the following steps. 
\begin{itemize}
	\item Find the nearest neighbor ${\boldsymbol{h}}_{in}^t = \mathrm{F}({\boldsymbol{h}}_i^t, \bar{\boldsymbol{H}}_t)$.
	\item Construct the NNH $\mathcal{H}_i^t$ combining ${\boldsymbol{h}}_{i}^t$ and ${\boldsymbol{h}}_{in}^t$.  
\end{itemize}

As shown in~Fig.~\ref{fig:fw}, after we complete the NNH geometry building, the subsequent calculations are performed based on the local geometry instead of individual data. In other words, we perform a switch for the fundamental clustering unit.

\noindent \textbf{Remarks.} 
The NNH construction in the iteration stage is different from the operation in the initiation stage (refer to the static NNH construction operation in Section~\ref{sec:pl}). 
Due to the update of model parameters, for instance $\boldsymbol{x}_i^t$, the target model outputs different deep features ${\boldsymbol{h}}_{i}^t$ that lead to the various nearest neighbors ${\boldsymbol{h}}_{in}^t$. Therefore, we term the construction in the iteration stage as a dynamical NNH construction. 
%In addition, our NNH is the simplest local geometry directly formed by spatial information.     
% In this subsection, we present the generation method for the simplest NNH formed by spatial information, while in Section~\ref{sec:method-ext}, we introduce another advanced implementation.  

\subsection{SCNNH-based Regularization for Model Adaptation}\label{sec:obj}
To drive NNH-based deep clustering, inspired by~\cite{2020shot}, this paper adopts a joint objective represented by Eqn.~\eqref{eqn:objctive} to regulate the clustering.
%including an IM component and self-supervised component.
\begin{equation}
	\label{eqn:objctive}
	\mathop{\min}_{\{\theta_t,\phi_t\}} \mathcal{L}_{g_t\circ u_t}^{t}\left({\mathcal{H}}_i^t\right) = \mathcal{L}_{im}^{t}\left({\mathcal{H}}_i^t\right) + \beta \mathcal{L}_{ss}^{t}\left({\mathcal{H}}_i^t\right).
\end{equation}
In the Eqn. above, $\mathcal{L}_{im}^t$ is an NNH-based information maximization ~(IM)~\cite{krause2010discriminative} regularization, $\mathcal{L}_{ss}^t$ is an NNH-based self-supervised regularization, $\beta$ is a trade-off parameter. In the clustering process, $\mathcal{L}_{im}^t$ mainly drives global clustering while $\mathcal{L}_{ss}^t$ provides category-based adjustment to correct the wrong move. Different from~\cite{2020shot}, our objective is built on NNH instead of individual data.

This paper proposes the SCNNH constraint, introduced in Section~\ref{sec:scnnh}, to encourage robust deep clustering taking NNH as the fundamental unit. Aiming at the \textit{same semantic representation on NNH}, we implement this constraint in both regularizations $\mathcal{L}_{im}^t$ and $\mathcal{L}_{ss}^t$. 
For input instance $\boldsymbol{x}_i^t$, the target model outputs its NNH ${\mathcal{H}}_i^t$, and finally outputs ${\mathcal{V}}_i^t$. We write the probability vectors of ${\mathcal{V}}_i^t$  as $\boldsymbol{P}_i^t=\{\boldsymbol{p}_i^t, \boldsymbol{p}_{in}^t\}$ after the softmax operation. 
To ensure the data in NNH have similar semantics, we make them move concurrently to a specific cluster centroid, such that NNH moves as a whole. To achieve this goal, we perform a semantic fusion operation $\mathrm{G}\left({\boldsymbol{P}}_i^t\right)$, as shown below. 
\begin{equation}
	\label{eqn:im-fusion}
	\begin{split}
		\mathrm{G}\left({\boldsymbol{P}}_i^t\right) &= \omega_{i}{\boldsymbol{p}}_{i}^{t} + \omega_{in}{\boldsymbol{p}}_{in}^{t},
	\end{split}
\end{equation}
where $\omega_{i}$ and $\omega_{in}$ are weight parameters. 
Let $\hat{\boldsymbol{p}}_{i}^{t}=\mathrm{G}\left({\boldsymbol{P}}_i^t\right)$, we formulate the IM term $\mathcal{L}_{im}^t$ by the following objective.
\begin{equation}
	\label{eqn:im}
	\begin{split}
		\mathcal{L}_{im}^{t}\left({\mathcal{H}}_i^t\right) =-\frac{1}{n_t}\sum_{i=1}^{n_t} \sum_{k=1}^{K}  {\hat{p}}_{i,k}^t \log {\hat{p}}_{i,k}^t + \sum_{k=1}^K{{\varrho}}_k^t\log{{\varrho}}_k^t,
	\end{split}
\end{equation}
where ${\hat{p}}_{i,k}^t$ is the $k$-th element of the fused probability vector $\hat{\boldsymbol{p}}_{i}^{t}$, ${{\varrho}}_k^t=\frac{1}{n_t} \sum_{i=1}^{n_t}{\hat{p}}_{i,k}^t$ is a mean in the $k$-th dimension over the dataset. In Eqn.~\ref{eqn:im}, the first term is an information entropy minimization that ensures clustering of NNH and the second term 
balances cluster assignments that encourage aggregation diversity over all clusters.

We achieve the semantic consistency of the data in NNH through enforcing the move of NNH, as represented in $\mathcal{L}_{im}^t$ above. However, this IM regularization cannot absolutely guarantee a move to the true category of the input instance. Therefore, we use the pseudo-labels to impose a category guidance. To this end, we introduce the self-supervised regularization $\mathcal{L}_{ss}^t$ formulated by Eqn.~\eqref{eqn:ss}.
\begin{equation}
	\label{eqn:ss}
	\begin{split}
		\mathcal{L}_{ss}^{t}\left({\mathcal{H}}_i^t\right) &=-\eta_i\left(\frac{1}{n_t}\sum_{i=1}^{n_t} \sum_{k=1}^{K}  \mathrm{I}[k=\bar{y}_i^t]\log {{p}}_{i,k}^t\right)\\
		&-\eta_{in}\left(\frac{1}{n_t}\sum_{i=1}^{n_t} \sum_{k=1}^{K} \mathrm{I}[k=\bar{y}_i^t]\log {{p}}_{in,k}^t\right),
	\end{split}
\end{equation}
where $\eta_i$ and $\eta_{in}$ are weight parameters, $\mathrm{I}[\cdot]$ is the function of indicator, ${{p}}_{i,k}^{t}$ and ${{p}}_{in,k}^{t}$ are the $k$-th elements of probability vectors ${\boldsymbol{p}}_{i}^{t}$ and ${\boldsymbol{p}}_{in}^{t}$ respectively.

\section{Deep Clustering based on Semantic Hyper-nearest Neighborhood}\label{sec:method-ext}

In category learning, the similarity-based cognition strategy will work well when the semantics of the most similar object is reliable. 
However, in the NNH based on the spatial information, as represented in Section~\ref{sec:pl}, we suppose all nearest neighbors' semantic credibilities are equal. 
To better mimic the working mechanism in category learning, we propose a more SCNNH-compliant version of NNH by introducing semantic credibility constraints. 
We term this new structure as semantic hyper-nearest neighborhood~(SHNNH).
In the following, we firstly introduce the structure of SHNNH, followed by its construction method. In the end, we present how to extend our method to this new geometry.

% chain search
\subsection{Semantic Hyper-nearest Neighborhood~(SHNNH)}
\begin{figure}[t]
	\begin{center}
		\includegraphics[width=0.85\linewidth]{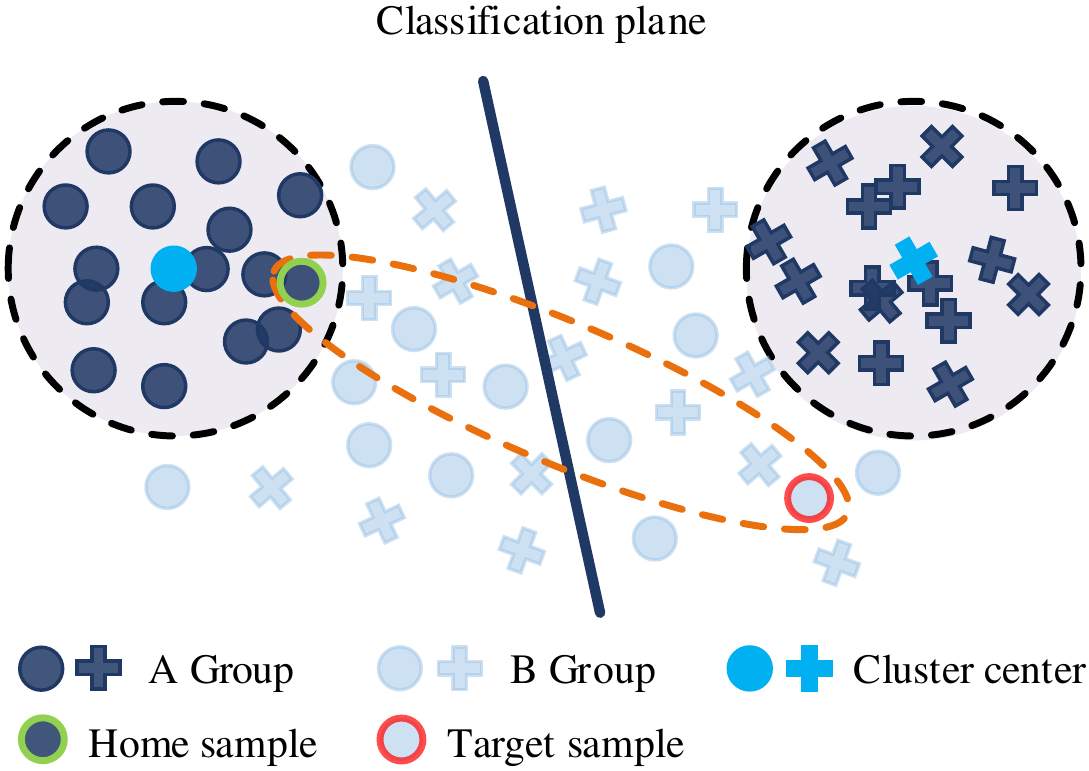}
	\end{center}
	\caption{Structure of SHNNH geometry.} %After being mapped by a deep source model, the target data will have an clustering attribution to some extent. The observation inspires us from two folds. First, essence of SFUDA is to align the two groups. Second, the relatively credible data (A group) may be exploited to regulate the alignment.}
	\label{fig:mov-shnnh}
\end{figure}

Using the feature extractor $u_s$ to map the target data into the deep feature space, we find that the obtained feature data has particular clustering (discriminative) characteristics (see Fig.~\ref{fig:mov-shnnh}), benefiting from the powerful feature extraction capabilities of deep networks. 
For one specific cluster, we can roughly divide the feature samples into two categories: \textbf{\romannumeral1)} data located near the cluster centers (A group) and \textbf{\romannumeral2)} samples far away from the centers (B group). 
Some samples in the B group closely distribute over the classification plane. 
This phenomenon implies that the semantic information of the A group is more credible than  that of the B group. Therefore, we plan to select samples from the A group to construct NNH.

Suppose samples in the A group and B group (see Fig.~\ref{fig:mov-shnnh}) are respectively confident and unconfident data. We model the self-learning process in category learning by defining a hyper-local geometry, i.e., SHNNH (marked by an orange dotted oval in Fig.~\ref{fig:mov-shnnh}). 
We form SHNNH by a target sample (marked by a red circle) and a home sample (marked by a green circle). 
In this geometry, the home sample is the high-confidence data most similar to the target sample. 
%Compared to NNH based on spatial information, the home sample corresponds to the nearest neighbor in the NNH.

\subsection{SHNNH Geometry Construction}\label{sec:shnnh-constr}
As with previous NNH only constructed by the spatial information, we build SHNNH by detecting the home sample that we implement by two steps in turn: \textbf{(\romannumeral1)} confident group splitting and \textbf{(\romannumeral2)} home sample detection. 
Using these in combination, we find the most similar data with high confidence.
%In the two steps, the first operation is performed in the initialization stage of an epoch as a new component of the auxiliary information generation. The second operation is used in two places, including the pseudo-label generation and the iteration stage of an epoch.

%Same as previous mentioned, \
%Same to the NNH contructed by the nearest neighbor, 
%we also construct SHNNH by finding the home sample. To this end, we firstly split confident group in the initialization stage of an epoch. Because the the grouping information is for during whole epoch training, we take this operation as a new component of the auxiliary information generation.

%To construct SHNNH, we firstly split the confident group and then detect the home sample.
%We implement the confident group splitting in the initialization stage of an epoch and take this operation as a new component of the auxiliary information generation. 

%As a component   
%auxiliary information generation
%operation of confident group splitting is 

%We implement the SHNNH construction using two steps: \textbf{1)} confident group splitting and \textbf{2)} home sample detection. 
%In the two steps, the first operation is performed in the initialization stage of an epoch, as a new componen t of the auxiliary information generation. The second operation is used in two places including the pseudo-label generation and the iteration stage of an epoch.

%\noindent \textbf{A. confident group splitter}. 
\textbf{Confident Group Splitting}.
This operation is performed in the initialization stage.
In this step, based on the information entropy of data, we split the target data into a confident group $\mathcal{C}^e$ and unconfident group in the deep feature space. 

Suppose that the auxiliary information, including the deep features $\bar{\boldsymbol{H}}_t$, the low-dimensional features $\bar{\boldsymbol{B}}_t$ and the logits feature $\bar{\boldsymbol{V}}_t$ , is pre-computed. After the softmax operation, the final probability vector is $\{\bar{\boldsymbol{p}}_i^t\}_{i=1}^{n_t}$.  
We adopt a simple strategy to obtain the confident group according to the following equation. 
\begin{equation}
	\label{eqn:ce}
	\begin{split}
	 \bar{\mathcal{C}}^e=\left\{{\bar{\boldsymbol{h}}_i^t}~|~ent_i^t < \gamma_e \right\}, \bar{\boldsymbol{h}}_i^t \in \bar{\boldsymbol{H}}_t
	\end{split}
\end{equation}
where $ent_i^t=-\sum_{k=1}^K{\bar{\boldsymbol{p}}}_{i,k}^t\log{\bar{\boldsymbol{p}}}_{i,k}^t$ is the information entropy corresponding to ${\bar{\boldsymbol{h}}_i^t}$, threshold $\gamma_e$ is the median value of the entropy over all target data.

Also, to obtain more credible grouping, we perform another splitting strategy based on the minimum distance to the cluster centroid in the low-dimensional feature space. We denote the new confident group as $\mathcal{C}^d$.  The new splitting strategy contains the two following steps.

Firstly, we conduct a weighted k-means clustering and get $K$ cluster centroids and compute the similarity-based logits of all target data $\{\bar{\boldsymbol{q}}_{i}^{t}\}_{i=1}^{n_t}$ according to Eqn.~(\ref{eqn:logits}) and Eqn.~(\ref{eqn:ok}). 
Secondly, we obtain the new confident group by Eqn.~(\ref{eqn:new_confi}).
\begin{equation}
	\label{eqn:new_confi}
	\begin{split}
		\bar{\mathcal{C}}^d&=\left\{{\bar{\boldsymbol{h}}_i^t}~|~\bar{d}_i < \gamma_d \right\}, \bar{\boldsymbol{h}}_i^t \in \bar{\boldsymbol{H}}_t\\
		{\bar{d}}_i &= \min\left(\bar{\boldsymbol{q}}_{i}\right),
	\end{split}
\end{equation}
where $\min(\cdot)$ is a function that outputs the minimum of the input vector, and threshold $\gamma_d$ is also the median value of a measure-set $\bar{\boldsymbol{D}}_t=\{{\bar{d}}_i\}_{i=1}^{n_t}$.
Combining $\bar{\mathcal{C}}^e$ and $\bar{\mathcal{C}}^d$, we get the final confident group $\bar{\mathcal{C}}$ by conducting an intersection operation represented by Eqn.~(\ref{eqn:c}).
\begin{equation}
	\label{eqn:c}
	\bar{\mathcal{C}} = \bar{\mathcal{C}}^e  \cap \bar{\mathcal{C}}^d.
\end{equation}
\begin{figure}[b]
	\begin{center}
		\includegraphics[width=0.85\linewidth]{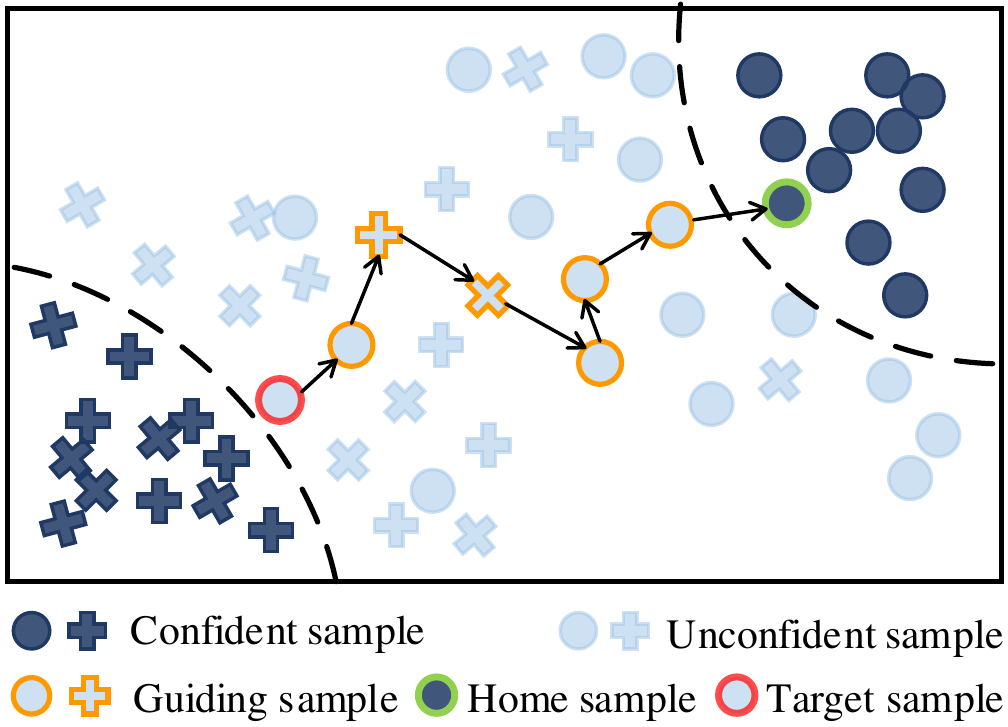}
	\end{center}
	\caption{Illustration of home sample detection by chain-search in the deep feature space.}
	\label{fig:home-sam}
\end{figure}

%\noindent \textbf{B. home sample detector}. 
\textbf{Home Sample Detection}.
Without loss of generality, we present our detecting method focusing on the geometry construction in pseudo-label generation.
In the implementation, we do not directly find the home sample from the confident group $\bar{\mathcal{C}}$ but propose a chain-search method that fully considers the nearest neighbor constraint.

As shown in Fig.~\ref{fig:home-sam}, the method starts with a target sample (with a red circle) in the deep feature space. 
Subsequently, it searches a serial of guiding samples (with an orange circle) one by one without repeating itself, using most-similarity comparison, until the home sample (with a green circle) is detected. 
In the detection chain, only the ending home samples belong to the confident group.  

%in an embedding space, starting with a tray sample (with green circle) we continue searching a serail of guide samples (with orange circumference) using a similarity comparision one by one until the home sample (with green circumferenc) is detected, i.e., the found sample belongs to the confidence group. Furthmore, in a intermediate step of the search, the previous guide samples do not be included in the guide sample search of next iteration. 
%Suppose in the embedding space defined by $f_t$ the confident sample of $\mathcal{X}_t$ is We map the target data and the confident group into an embedding space defined by $f_t$ and obtain two embedding sets ${\mathcal{Z}_t}=\{{\mathbf{z}_i^t} | i=1,2,\cdots,n\}$ and $\mathcal{Z}_t^c=f_t({\mathcal{C}})$.  

Given the $(k-1)$-th guiding sample ${\bar{\boldsymbol{h}}_{ig}^{t\left\langle {k-1} \right\rangle}}$ and a temporary set $\boldsymbol{I}_k$ storing the found guiding samples. We formulate the search over the deep features $\bar{\boldsymbol{H}}_t$ for the $k$-th new guiding sample ${\bar{\boldsymbol{h}}_{ig}^{t\left\langle {k} \right\rangle}}$ by the following optimization where $\left\langle {k} \right\rangle$ means the $k$-th iteration.
\begin{equation}
	\label{eqn:gudingdata}
	\begin{split}
		{\bar{\boldsymbol{h}}_{ig}^{t{\left\langle {k} \right\rangle}}}&={\bar{\boldsymbol{h}}_{j^{'}}^{t}}, ~~{j^{'}}= \arg \min \limits_{j}\mathrm{D}_{sim}({\bar{\boldsymbol{h}}_{ig}^{t{\left\langle {k-1} \right\rangle}}}, {\bar{\boldsymbol{h}}_{j}^{t}}),\\ 
		s.t. ~~&j=1,2,\cdots, n_t;~{ig} \neq j;~\bar{\boldsymbol{h}}_{j^{'}}^t \in \bar{\boldsymbol{H}}_t;~\bar{\boldsymbol{h}}_{j^{'}}^t \notin I_k;\\
	\end{split}
\end{equation}
%Given the $(j-1)$-th guiding sample ${\bar{\boldsymbol{h}}_{ig}^{t\left\langle {j-1} \right\rangle}}$ and a temporary set $\boldsymbol{I}_j$ storing the found guiding samples. We formulate the search over the deep features $\bar{\boldsymbol{H}}_t$ for the $j$-th new guiding sample ${\bar{\boldsymbol{h}}_{ig}^{t\left\langle {j} \right\rangle}}$ by the following optimization where $\left\langle {j} \right\rangle$ means the $j$-th iteration.
where $\mathrm{D}_{sim}\left(\cdot,\cdot\right)$ is the cosine distance function.
We use function $R(\cdot;\bar{\boldsymbol{H}}_t,\bar{\mathcal{C}})$ to express the home sample search process in general. 
For deep feature $\bar{\boldsymbol{h}}_i^t$, corresponding to any target data ${\boldsymbol{x}}_i^t$, the search for the home sample $\bar{\boldsymbol{h}}_{ih}^t = R(\bar{\boldsymbol{h}}_i^t;\bar{\boldsymbol{H}}_t,\bar{\mathcal{C}})$ can be expressed by Alg.~\ref{alg:alg_confi}.
\begin{algorithm}[h]
	\caption{Pseudo-code of home sample detection.}
	\label{alg:alg_confi}
	\textbf{Input}: $\bar{\boldsymbol{h}}_i^t$, $\bar{\boldsymbol{H}}_t$ and $\bar{\mathcal{C}}$\\
	%\textbf{Parameter}: Optional list of parameters\\
	\textbf{Output}: ${\bar{\boldsymbol{h}}_{ih}^t}$\\
	\textbf{Initialization}: $k=1$, ${\bar{\boldsymbol{h}}_{ig}^{t{\left\langle {k-1} \right\rangle}}}=\bar{\boldsymbol{h}}_{i}^t$ and $\boldsymbol{I}_k=\varnothing$.\\
	\vspace{-1.2em}
	\begin{algorithmic}[1] %[1] enables line numbers
		%\STATE Let $k=1$, ${\bar{\boldsymbol{h}}_i^{t(k-1)}}=\bar{\boldsymbol{h}}_{ig}^t$, $\boldsymbol{I}_k=\varnothing$.
		%\WHILE{${\boldsymbol{h}_{ig}^{t(k)}} \notin {\mathcal{C}}$}
		\STATE \textbf{do}
		\STATE ~~~Find the guiding sample ${\bar{\boldsymbol{h}}_{ig}^{t{\left\langle {k} \right\rangle}}}$ by Eqn.~\ref{eqn:gudingdata}.
		\STATE ~~~Add ${\bar{\boldsymbol{h}}_{ig}^{t{\left\langle {k} \right\rangle}}}$ into set $\boldsymbol{I}_k$.
		\STATE ~~~Update ${\bar{\boldsymbol{h}}_{ig}^{t{\left\langle {k-1} \right\rangle}}}$ by ${\bar{\boldsymbol{h}}_{ig}^{t{\left\langle {k-1} \right\rangle}}}={\bar{\boldsymbol{h}}_{ig}^{t{\left\langle {k} \right\rangle}}}$.
		\STATE ~~~Update searching counter $k+=1$.
		%\ENDWHILE
		\STATE \textbf{while} ${\bar{\boldsymbol{h}}_{ig}^{t{\left\langle {k} \right\rangle}}} \notin \bar{\mathcal{C}}$.
		\STATE \textbf{return} The home sample $\bar{\boldsymbol{h}}_{ih}^t=\bar{\boldsymbol{h}}_{ig}^{t{\left\langle {k} \right\rangle}}$.
	\end{algorithmic}
\end{algorithm}

%\begin{algorithm}[H]
%\caption{Algorithm}
%\begin{algorithmic}[1]
%\renewcommand{\algorithmicrequire}{\textbf{Input:}}
%\renewcommand{\algorithmicensure}{\textbf{Output:}}
%\REQUIRE $Graph\ G(V, E)$

%\STATE $\textbf{\textit{function}}\ $
%  \Do
%    \State Something
%  \doWhile
%\end{algorithmic}
%\label{algo1}
%\end{algorithm}

\subsection{Extending Our Method to SHNNH}
As shown in Fig.~\ref{fig:fw}, the adaptation method proposed in this paper is a general framework that we can smoothly extend to the new geometry through simple actions. 
For SHNNH, we extend our method to this new geometry by updating three modules.  
Specifically, we update the auxiliary data constructor by integrating the confident group splitting operation introduced in Section~\ref{sec:shnnh-constr}. In the meantime, we let $\bar{\boldsymbol{h}}_{in}^t=\bar{\boldsymbol{h}}_{ih}^t$ and ${\boldsymbol{h}}_{in}^t={\boldsymbol{h}}_{ih}^t=R({\boldsymbol{h}}_i^t;\bar{\boldsymbol{H}}_t,\bar{\mathcal{C}})$, by which we perform the update for the static NNH construction (see Section~\ref{sec:pl}) and the dynamical NNH construction (see Section~\ref{sec:dymmh}), respectively.

In the rest of this paper, we denote the original method, based on the NNH without semantic credibility constraint, as N2DC. Meanwhile, we term the extended method, based on SHNNH, as N2DC-EX for clarity.

\section{Experiments}\label{sec:rlt}
\subsection{Datasets}
% saenko2010adapting, venkateswara2017deep,  peng2017visda
\textbf{Office-31}~\cite{saenko2010adapting} 
is a small benchmark that has been widely used in visual domain adaptation. The dataset includes 4,652 images of 31 categories, all of which are of real-world objects in an office environment. There are three domains in total, i.e., Amazon (A), Webcam (W), and DSLR (D). Images in (A) are online e-commerce pictures from the Amazon website. (W) consists of low-resolution pictures collected by webcam. (D) are of high-resolution images taken by SLR cameras.

\textbf{Office-Home}~\cite{venkateswara2017deep} 
is another challenging medium-sized benchmark released in 2017. It is mainly used for visual domain adaptation and consists of 15,500 images, all of which are from a working or family environment. There are 65 categories in total, covering four distinct domains, i.e., Artistic images (Ar), Clipart (Cl), Product images (Pr), and Real-World images (Rw).
%is released in 2017 contains 15,500 images belonging to 65 categories come from office and home circumstances. As a challenging medium-sized benchmark, it is mainly used for the research in the domain adaptation field, which consists of four distinct domains, including Artistic images (Ar), Clip Art (Cl), Product images (Pr), and Real-World images (Rw).

\textbf{VisDA-C}~\cite{peng2017visda} 
is the third dataset used in this paper. Different from Office-31 and Office-Home, the dataset is a large benchmark for visual domain adaptation, including target classification and segmentation, and 12 types of synthetic to real target recognition tasks. The source domain contains 152,397 composite images generated by rendering 3D models, while the target domain has 55,388 real object images from Microsoft COCO.

\subsection{Experimental Settings}
\textbf{Network setting}. In experiments, we do not train the source model from scratch. Instead, the feature extractor is transferred from pre-trained deep models. Similar to the work in~\cite{long2018conditional,xu2019larger}, ResNet-50 is used as the feature extractor in the experiments on small and medium-sized datasets (i.e., Office-31 and Office-Home). For the experiments on VisDA-C, ResNet-101 is adopted as the feature extractor. For the bottleneck and classifier, we adopted the structure proposed in~\cite{2020shot,yang2020unsupervised}.
In the bottleneck and the classifier, the fully-connect layers have a size of $2048\times 256$ and $256\times K$, respectively, in which $K$ differs from one dataset to another.   

%Specifically, the bottleneck consists of a batch-normalization layer and a fully-connect layer with a size of $2048\times 256$. The classifier consists of a weight-normalization layer and a fully-connect layer with a size of $256\times K$, in which $K$ differs from one dataset to another. 

%\vspace{2pt}
%\noindent\textbf{B. Training setting}.~
\textbf{Training setting}.
On all experiments, we adopt the same setting for the hyper-parameter $\gamma$, $(\alpha,\delta)$, $\beta$,  $(\omega_{i},\omega_{in})$, and $(\eta_{i},\eta_{in})$.
Among these parameters, $\gamma$ is a weight parameter that indicates the  degree of relaxation for the smooth label (see Eqn.~\eqref{equ:opt_source}). Same as \cite{szegedy2016rethinking}, we chose $\gamma=0.1$.
$(\alpha,\delta)$ as the mean and variance of the random variable $\lambda$ used in semantic fusion for pseudo-label generation (see Eqn.~\eqref{eqn:fusion-pl}). We set $(\alpha,\delta)=(0.85,0.15)$ to ensure that the fused pseudo-label of NNH contains more semantics from the input instance.
$\beta$ is a trade-off parameter representing the confidence for the self-supervision from the semantic consistency loss $\mathcal{L}_{ss}^t$ (see Eqn.~\eqref{eqn:objctive}). In this paper, we choose a relaxed value of 0.2 for $\beta$. 
$(\omega_{i},\omega_{in})$ in Eqn.~\eqref{eqn:im-fusion} and $(\eta_{i},\eta_{in})$ in Eqn.~\eqref{eqn:ss} are weight coefficients representing the effects of the two samples forming the NNH. Due to equal importance to NNH construction, we set them to the same value.   
%Regarding the distance function, i.e., $\mathrm{D}_{sim}(\cdot, \cdot)$ in Eqn.~\ref{eqn:nnh-pl} and Eqn.~\ref{eqn:gudingdata}, we select Cosine Distance to compute the similarity used in our method. 
Under random seed 2020, we run the codes repeatedly for 10 rounds and then take the average value as the final result.

%Because our method is closely related to SHOT, to be fair, we also re-ran all experiments using SHOT on the same test bed chosen for our method. %Moreover, to exclude random factors' influence, as also done in SHOT, we run the code at three different random seeds ($2019, 2020, 2021$). 
 
%The results of other methods are directly cited from the published papers.  

%$\alpha$ is the weight parameter for the mix-up operation (see Eqn.~\eqref{eq:mix}). From the view of geometry, $\alpha$ determines the location of the augmented sample in the inner area of the NNH. Because the distances of the master sample and the servant sample to the classification plane are unknown, $\alpha$ takes a fair value of 0.5. That is, the augmented sample is located in the midpoint of the NNH.

%In our method, the NNH orientation comes from an adjustment caused by the servant sample and the augmented sample. In order to guarantee the relative intensity of the adjusting forces, we weakened the influence of the master sample by setting $\omega_{i}$ and $\lambda_{i}$ to a small value. Typically,  $\omega_i=\lambda_{i}=0.5$ is half of the other two values.

\subsection{Baseline methods}
%In the this section, the evaluations were carried out on three domain adaptation benchmarks mentioned in previous subsection. 
%\vspace{2pt}
%\noindent\textbf{A. Baseline methods}.
We carry out the evaluation on three domain adaptation benchmarks mentioned in the previous subsection. 
To verify the effectiveness of our method, we selected 18 comparison methods and divided them into the following three groups. 

\begin{itemize}
	\item The first group includes 13 state-of-the-art UDA methods requiring access to the source data, i.e.,  DANN\cite{ganin2015unsupervised}, CDAN~\cite{long2018conditional}, CAT~\cite{deng2019cluster}, BSP~\cite{chen2019transferability}, SAFN~\cite{xu2019larger}, SWD~\cite{lee2019sliced}, ADR~\cite{saito2017adversarial}, TN~\cite{wang2019transferable}, IA~\cite{jiang2020implicit},  BNM~\cite{cui2020towards}, BDG~\cite{yang2020bi}, MCC~\cite{jin2020minimum} and SRDC~\cite{tang2020unsupervised}.
	\item The second group comprises four state-of-the-art methods for UDA without access to the source data. They are SHOT~\cite{2020shot}, SFDA~\cite{kim2020domain}, MA~\cite{2020MA} and BAIT~\cite{yang2020unsupervised}. 
	\item The third group includes the pre-trained deep models, namely ResNet-50 and ResNet-101~\cite{he2016deep}, that are used to initiate the feature extractor of the source model before training on the source domain.
\end{itemize}

%Because our method is closely related to SHOT, to be fair, we also re-ran all experiments using SHOT on the same test bed chosen for our method. %Moreover, to exclude random factors' influence, as also done in SHOT, we run the code at three different random seeds ($2019, 2020, 2021$). 
%Under random seed 2020, the codes repeatedly runs for 10 rounds. We take the average value as the final result. The results of other methods are directly cited from the published papers.  

%The code for N2O Net is available on: \url{https://github.com/DouMM/N2O-Net/}.
\begin{table}[h]
	\caption {Classification accuracies (\%) of 6 transfer tasks on the small Office-31 data set. The bold means the best result, the underline means the second-best result, and \textbf{SDA} means Source Data-Absent.}
	\label{tab:office}
	\renewcommand\tabcolsep{2pt}
	\renewcommand\arraystretch{1.0}
	%\small
	\footnotesize%
	%\scriptsize 
	\centering
	\begin{tabular}{ l c c c c c c c c}
		\toprule
		Method~$(\mathcal{S}\to\mathcal{T})$  &\textbf{SDA} &A$\to$D &A$\to$W &D$\to$A &D$\to$W &W$\to$A &W$\to$D &Avg.\\
		\midrule
		ResNet-50~\cite{he2016deep}        &$\times$  &68.9 &68.4 &62.5 &96.7 &60.7 &99.3 &76.1\\
		DANN~\cite{ganin2015unsupervised}  &$\times$  &79.7 &82.0 &68.2 &96.9 &67.4 &99.1 &82.2\\
		CDAN~\cite{long2018conditional}    &$\times$  &92.9 &94.1 &71.0 &98.6 &69.3 &\textbf{100.} &87.7\\
		CAT~\cite{deng2019cluster}         &$\times$  &90.8 &94.4 &72.2 &98.0 &70.2 &\textbf{100.} &87.6\\
		SAFN~\cite{xu2019larger}           &$\times$  &90.7 &90.1 &73.0 &98.6 &70.2 &\uline{99.8} &87.1\\
		BSP~\cite{chen2019transferability} &$\times$  &93.0 &93.3 &73.6 &98.2 &72.6 &\textbf{100.} &88.5\\
		TN~\cite{wang2019transferable}     &$\times$  &94.0 &\uline{95.0} &73.4 &98.7 &74.2 &\textbf{100.} &89.3\\
		IA~\cite{jiang2020implicit}        &$\times$  &92.1 &90.3 &75.3 &98.7 &74.9 &\uline{99.8} &88.8\\
		BNM~\cite{cui2020towards}          &$\times$  &90.3 &91.5 &70.9 &98.5 &71.6 &\textbf{100.} &87.1\\
		BDG~\cite{yang2020bi}              &$\times$  &93.6 &93.6 &73.2 &\uline{99.0} &72.0 &\textbf{100.} &88.5\\
		MCC~\cite{jin2020minimum}          &$\times$  &95.6 &95.4 &72.6 &98.6 &73.9 &\textbf{100.} &89.4\\
		SRDC~\cite{tang2020unsupervised}   &$\times$  &\uline{95.8} &\textbf{95.7} &\textbf{76.7} &\textbf{99.2} &\uline{77.1} &\textbf{100.} &\textbf{90.8}\\
		\midrule
		SHOT~\cite{2020shot}       &\checkmark  &93.9 &91.3 &74.1 &98.2 &74.6 &\textbf{100.} &88.7\\
		SFDA~\cite{kim2020domain}  &\checkmark  &92.2 &91.1 &71.0 &98.2 &71.2 &99.5 &87.2\\
		MA~\cite{2020MA}           &\checkmark  &92.7 &93.7 &75.3 &98.5 &\textbf{77.8} &\uline{99.8} &89.6\\
		BAIT~\cite{yang2020unsupervised}         &\checkmark  &92.0 &94.6 &74.6 &98.1 &75.2 &\textbf{100.} &89.1\\   
		\midrule
		Source model only  &\checkmark  &80.7 &77.0 &60.8 &95.1 &62.3 &98.2 &79.0\\
		N2DC $($ours$)$  &\checkmark  &93.9 &89.8 &74.7 &98.6 &74.4 &\textbf{100.} &88.6 \\
		N2DC-EX $($ours$)$  &\checkmark  &\textbf{97.0} &93.0 &\uline{75.4} &98.9 &75.6 &\uline{99.8} &\uline{90.0} \\
		\bottomrule
	\end{tabular}
\end{table} 
\begin{table*}[t]
	\caption{Classification accuracies (\%) of 12 transfer tasks on the medium Office-Home dataset. The bold means the best result, the underline means the second-best result, and \textbf{SDA} means Source Data-Absent.}
	\label{tab:oh}
	\renewcommand\tabcolsep{2.3pt}
	\renewcommand\arraystretch{1.0}
	\centering
	\small
	%\footnotesize
	%\smallsize 
	\begin{tabular}{ l c c c c c c c c c c c c c c}
		\toprule
		Method~$(\mathcal{S}\to\mathcal{T})$ & \textbf{SDA}
		&Ar$\to$Cl &Ar$\to$Pr &Ar$\to$Rw
		&Cl$\to$Ar &Cl$\to$Pr &Cl$\to$Rw 
		&Pr$\to$Ar &Pr$\to$Cl &Pr$\to$Rw  
		&Rw$\to$Ar &Rw$\to$Cl &Rw$\to$Pr 
		&Avg.\\
		\midrule
		ResNet-50~\cite{he2016deep}        &$\times$  &34.9 &50.0 &58.0 &37.4 &41.9 &46.2 &38.5 &31.2 &60.4 &53.9 &41.2 &59.9 &46.1\\
		DANN~\cite{ganin2015unsupervised}  &$\times$  &45.6 &59.3 &70.1 &47.0 &58.5 &60.9 &46.1 &43.7 &68.5 &63.2 &51.8 &76.8 &57.6\\
		CDAN~\cite{long2018conditional}    &$\times$  &50.7 &70.6 &76.0 &57.6 &70.0 &70.0 &57.4 &50.9 &77.3 &70.9 &56.7 &81.6 &65.8\\
		BSP~\cite{chen2019transferability} &$\times$  &52.0 &68.6 &76.1 &58.0 &70.3 &70.2 &58.6 &50.2 &77.6 &72.2 &59.3 &81.9 &66.3\\
		SAFN~\cite{xu2019larger}           &$\times$  &52.0 &71.7 &76.3 &64.2 &69.9 &71.9 &63.7 &51.4 &77.1 &70.9 &57.1 &81.5 &67.3\\
		TN~\cite{wang2019transferable}     &$\times$  &50.2 &71.4 &77.4 &59.3 &72.7 &73.1 &61.0 &53.1 &79.5 &71.9 &59.0 &82.9 &67.6\\
		IA~\cite{jiang2020implicit}        &$\times$  &56.0 &77.9 &79.2 &64.4 &73.1 &74.4 &64.2 &54.2 &79.9 &71.2 &58.1 &83.1 &69.5\\
		BNM~\cite{cui2020towards}          &$\times$  &52.3 &73.9 &80.0 &63.3 &72.9 &74.9 &61.7 &49.5 &79.7 &70.5 &53.6 &82.2 &67.9\\
		BDG~\cite{yang2020bi}              &$\times$  &51.5 &73.4 &78.7 &65.3 &71.5 &73.7 &65.1 &49.7 &81.1 &\uline{74.6} &55.1 &84.8 &68.7\\
		SRDC~\cite{tang2020unsupervised}   &$\times$  &52.3 &76.3 &81.0 &\uline{69.5} &76.2 &78.0 &\textbf{68.7} &53.8 &81.7 &\textbf{76.3} &57.1 &85.0 &71.3\\
		\midrule
		SHOT~\cite{2020shot}         &\checkmark  &56.6 &78.0 &80.6 &68.4 &78.1 &\uline{79.4} &68.0 &54.3 &82.2 &74.3 &58.7 &84.5 &71.9\\
		SFDA~\cite{kim2020domain}    &\checkmark &48.4 &73.4 &76.9 &64.3 &69.8 &71.7 &62.7 &45.3 &76.6 &69.8 &50.5 &79.0 &65.7\\
		BAIT~\cite{2020MA}  &\checkmark &\textbf{57.4} &77.5 &\textbf{82.4} &68.0 &77.2 &75.1 &67.1 &\uline{55.5} &81.9 &73.9 &\uline{59.5} &84.2 &71.6\\
		\midrule
		Source model only  &\checkmark  &44.0 &67.0 &73.5 &50.7 &60.3 &63.6 &52.6 &40.4 &73.5 &65.7 &46.2 &78.2 &59.6\\
		N2DC $($ours$)$ 
		&\checkmark  &\uline{57.1} &\uline{79.1} &\uline{82.1} &69.2 &\uline{78.6} &\textbf{80.3} &\uline{68.3} &54.9 &\uline{82.4} &74.5 &59.2 &\uline{85.1} &\uline{72.6}\\
		N2DC-EX $($ours$)$ 
		&\checkmark  &\textbf{57.4} &\textbf{80.0} &\uline{82.1} &\textbf{69.8} &\textbf{79.6} &\textbf{80.3} &\textbf{68.7} &\textbf{56.5} &\textbf{82.6} &74.4 &\textbf{60.4} &\textbf{85.6} &\textbf{73.1}\\
		\bottomrule
	\end{tabular}
\end{table*}
\begin{table*}[t]
	\caption{Classification accuracies (\%) on the large VisDA-C dataset. The bold means the best result, the underline means the second-best result, and \textbf{SDA} means Source Data-Absent.}  
	\label{tab:vc}
	\renewcommand\tabcolsep{5pt}
	\renewcommand\arraystretch{1.0}
	\centering
	\small
	%\footnotesize%
	%\scriptsize 
	\begin{tabular}{ l c c c c c c c c c c c c c c}
		\toprule
		Method~(Syn. $\to$ Real)&\textbf{SDA} &plane &bcycl &bus &car &horse &knife &mcycl &person &plant &sktbrd &train &truck &Per-class\\
		\midrule
		ResNet-101~\cite{he2016deep}       &$\times$ &55.1 &53.3 &61.9 &59.1 &80.6 &17.9 &79.7 &31.2 &81.0 &26.5 &73.5 &8.5  &52.4\\ 
		DANN~\cite{ganin2015unsupervised}  &$\times$ &81.9 &77.7 &82.8 &44.3 &81.2 &29.5 &65.1 &28.6 &51.9 &54.6 &82.8 &7.8  &57.4\\ 
		ADR~\cite{saito2017adversarial}    &$\times$ &94.2 &48.5 &84.0 &72.9 &90.1 &74.2 &\textbf{92.6} &72.5 &80.8 &61.8 &82.2 &28.8 &73.5\\      
		CDAN~\cite{long2018conditional}    &$\times$ &85.2 &66.9 &83.0 &50.8 &84.2 &74.9 &88.1 &74.5 &83.4 &76.0 &81.9 &38.0 &73.9\\ 
		IA~\cite{jiang2020implicit}        &$\times$ &-    &-    &-    &-    &-    &-    &-    &-    &-    &-    &-    &-    &75.8\\
		BSP~\cite{chen2019transferability} &$\times$ &92.4 &61.0 &81.0 &57.5 &89.0 &80.6 &90.1 &77.0 &84.2 &77.9 &82.1 &38.4 &75.9\\     
		SAFN~\cite{xu2019larger}           &$\times$ &93.6 &61.3 &84.1 &70.6 &94.1 &79.0 &\uline{91.8} &79.6 &89.9 &55.6 &89.0 &24.4 &76.1\\
		SWD~\cite{lee2019sliced}           &$\times$ &90.8 &82.5 &81.7 &70.5 &91.7 &69.5 &86.3 &77.5 &87.4 &63.6 &85.6 &29.2 &76.4\\
		MCC~\cite{jin2020minimum}          &$\times$ &88.7 &80.3 &80.5 &71.5 &90.1 &93.2 &85.0 &71.6 &89.4 &73.8 &85.0 &36.9 &78.8\\
		\midrule
		SHOT~\cite{2020shot}        &\checkmark &95.0 &87.5 &81.0 &57.6 &93.9 &94.1 &79.3 &80.5 &90.9 &89.8 &85.9 &\uline{57.4} &82.7   \\
		SFDA~\cite{kim2020domain} &\checkmark &86.9 &81.7 &\uline{84.6} &63.9 &93.1 &91.4 &86.6 &71.9 &84.5 &58.2 &74.5 &42.7 &76.7\\
		MA~\cite{2020MA}          &\checkmark &94.8 &73.4 &68.8 &\uline{74.8} &93.1 &95.4 &88.6 &\textbf{84.7} &89.1 &84.7 &83.5 &48.1 &81.6\\
		BAIT~\cite{yang2020unsupervised}        &\checkmark &93.7 &83.2 &84.5 &65.0 &92.9 &95.4 &88.1 &80.8 &90.0 &89.0 &84.0 &45.3 &82.7\\
		\midrule
		Source model only   &\checkmark &62.1 &21.2 &48.8 &\textbf{77.8} &63.1 &5.0  &72.9 &25.9 &66.1 &44.1 &80.9 &5.3  &47.8\\
		N2DC $($ours$)$ &\checkmark &\uline{95.5} &\uline{88.1} &82.2 &58.7 &\uline{95.5} &\uline{95.8} &85.4 &81.4 &\uline{92.2} &\textbf{91.2} &\uline{89.7} &\textbf{58.4} &\uline{84.5}\\
		N2DC-EX $($ours$)$ &\checkmark  &\textbf{96.6} &\textbf{90.6} &\textbf{87.1} &62.6 &\textbf{95.7} &\textbf{96.1} &86.0 &\uline{82.5} &\textbf{93.8} &\textbf{91.3} &\textbf{90.4} &56.8 &\textbf{85.8}\\
		\bottomrule
	\end{tabular}
\end{table*}

%\noindent\textbf{B. Quantitative results}.~
\subsection{Quantitative results}
Table~\ref{tab:office}$\sim$\ref{tab:vc} report the experimental results on the three datasets mentioned above. 
On Office-31 (Table~\ref{tab:office}), N2DC's performance is close to SHOT. 
Compared to SHOT, N2DC decreases 0.1\% in average accuracy because there is 
a 1.5\% gap in task $A \to W$. 
By contrast, N2DC-EX achieves competitive results.
Compared to the SAUDA methods, N2DC-EX obtain the best performance in half the tasks and surpass MA, the previous best SAUDA method, by 0.4\% on average. 
In all comparison methods, N2DC-EX obtain the best result in task $A \to D$ and the second-best result in average accuracy. Considering that the best method SRDC has to work with the source data, we believe that the gap of 0.8\% is acceptable.
As the results have shown, N2DC and N2DC-EX do not show apparent advantages.
We argue that it is reasonable since the small dataset Office-31 cannot support the end-to-end training on our method's deep network. 
The results on Office-Home and VisDA-C in the following will confirm our expectation.

\begin{figure*}[t]
	\begin{center}
	    \begin{subfigure}{.24\linewidth}
			\centering
			\includegraphics[width=1\linewidth]{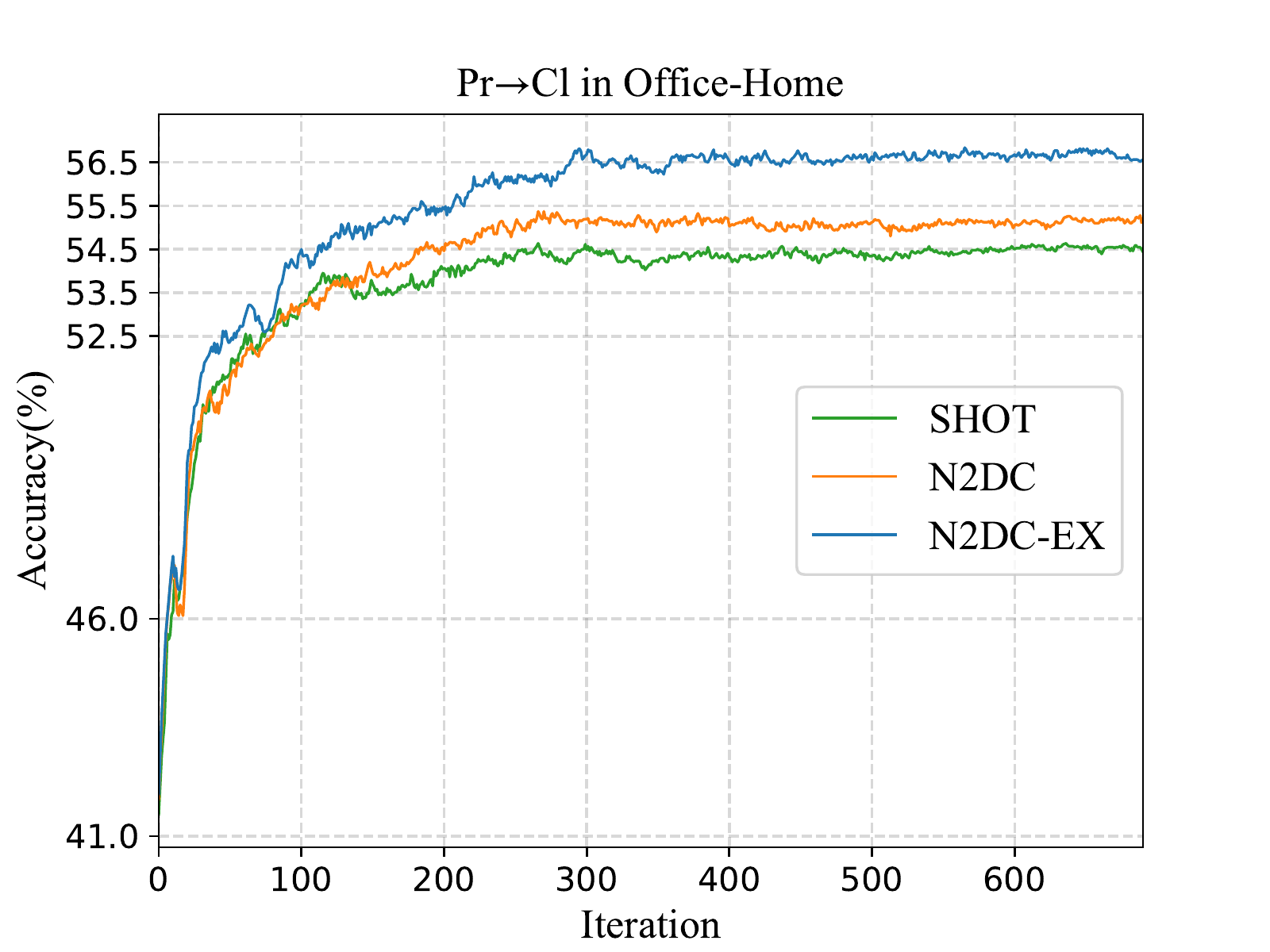}
			\caption{}
		\end{subfigure}
		\begin{subfigure}{.24\linewidth}
			\centering
			\includegraphics[width=1\linewidth]{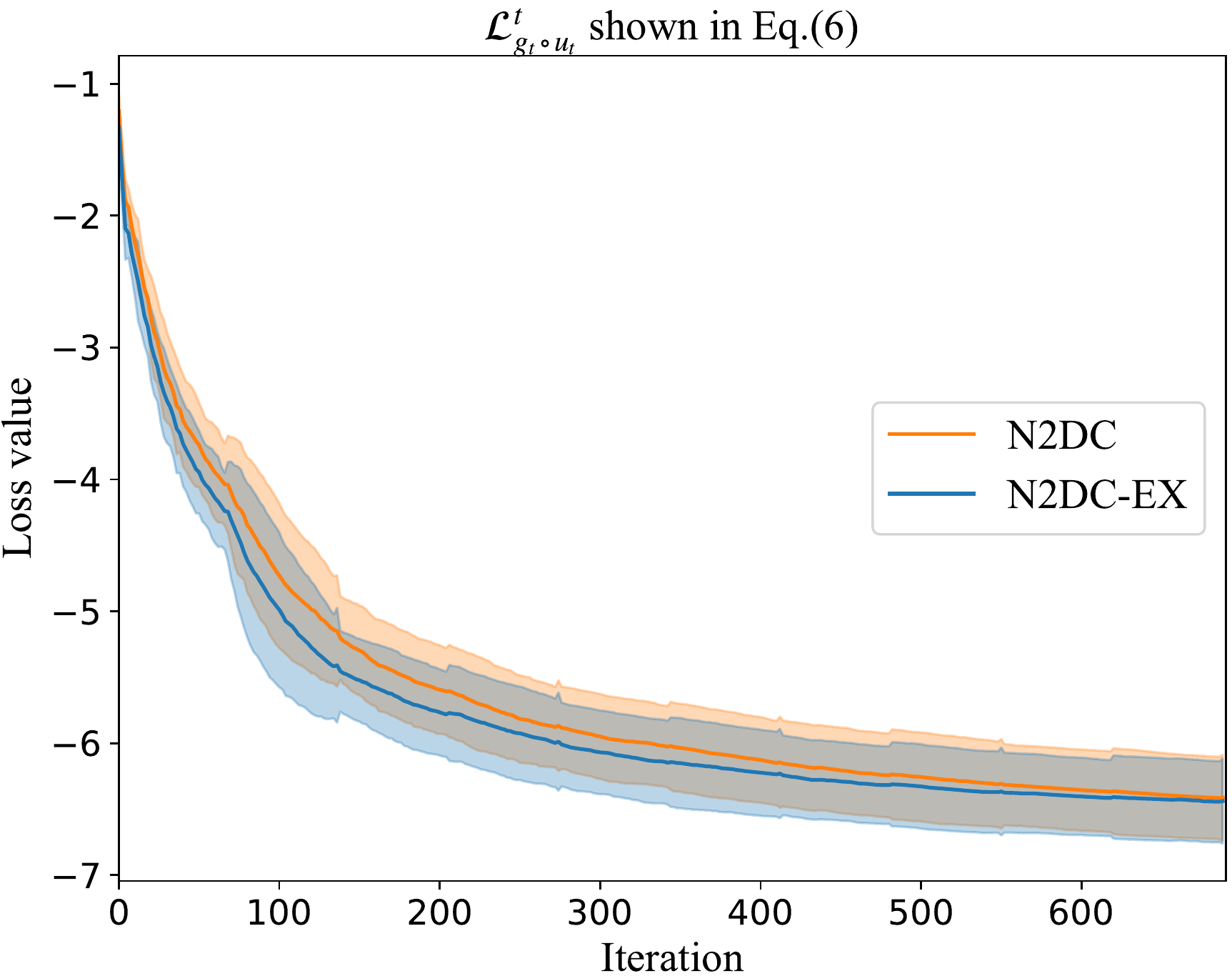}
			\caption{}
		\end{subfigure}
		\begin{subfigure}{.24\linewidth}
			\centering
			\includegraphics[width=1\linewidth]{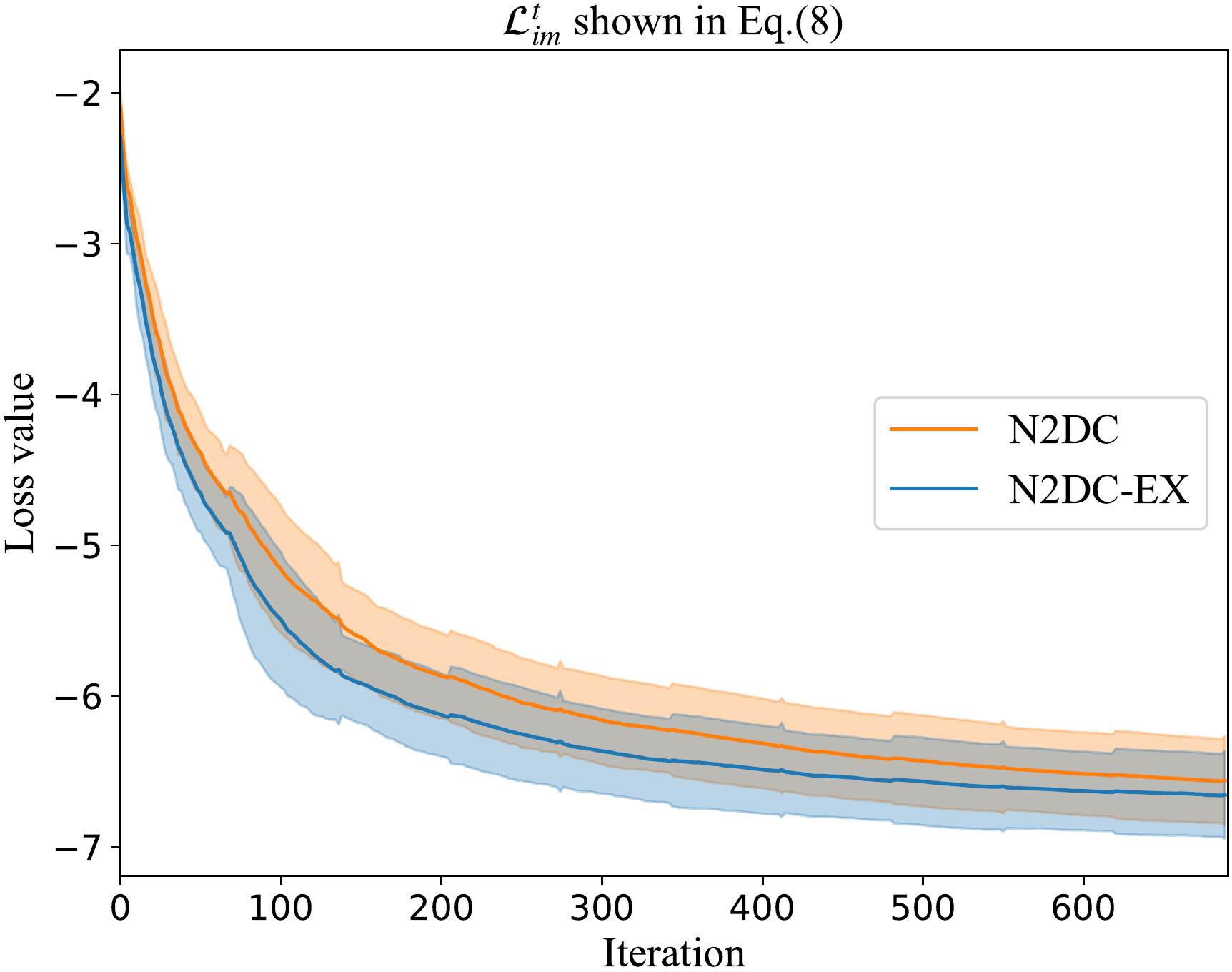}
			\caption{}
		\end{subfigure}
		\begin{subfigure}{.24\linewidth}
			\centering
			\includegraphics[width=1\linewidth]{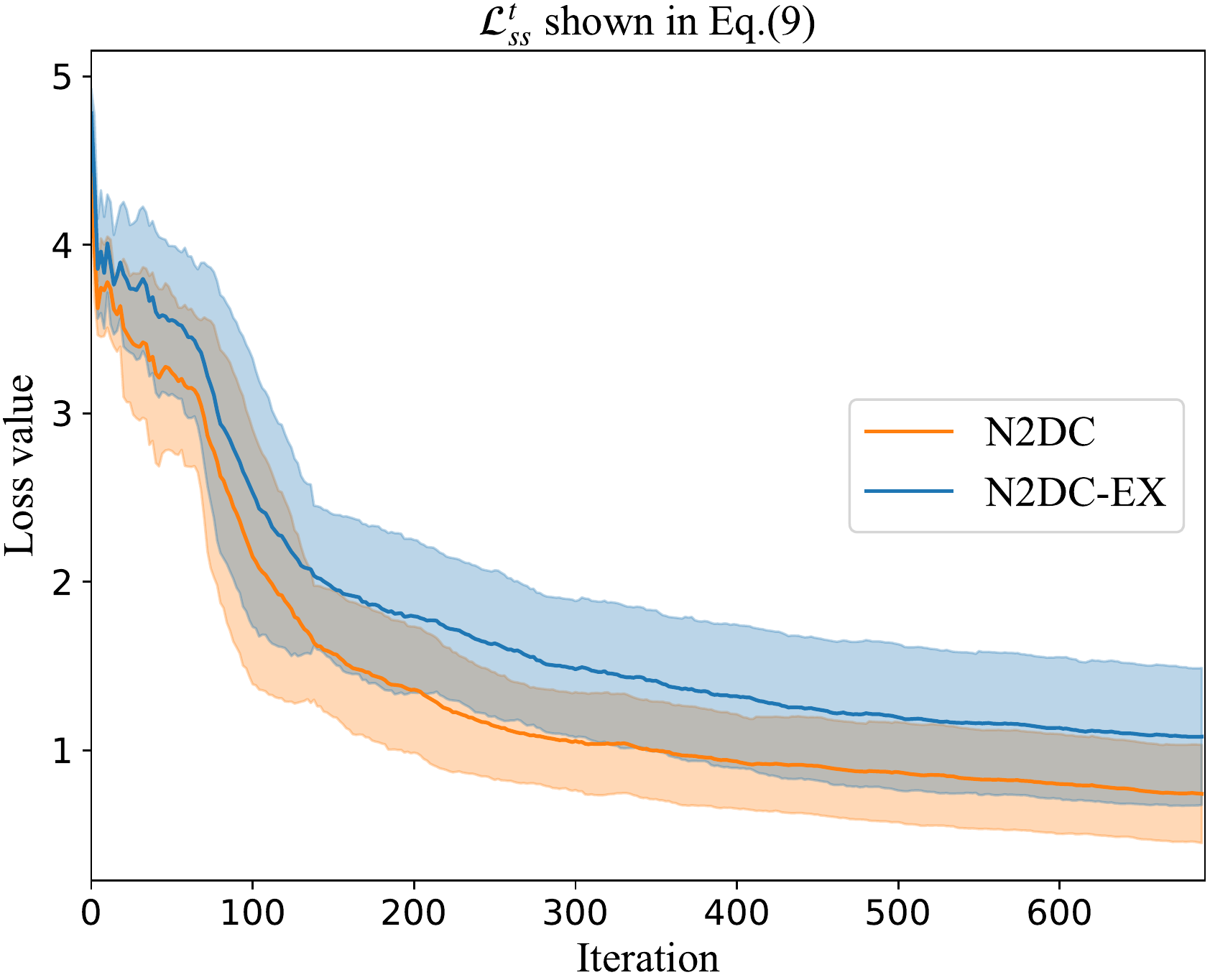}
			\caption{}
		\end{subfigure}
	\end{center}
	\setlength{\abovecaptionskip}{0mm}
	\setlength{\belowcaptionskip}{0cm}
	\caption{The accuracy and loss value of our objective during the model adaptation for task Pr$\to$Cl on Office-Home dataset. (a) Accuracies (\%) where SHOT is the baseline. (b), (c) and (d) present the loss value of $\mathcal{L}_{g_t\circ u_t}^{t}$, $\mathcal{L}_{im}^t$ and $\mathcal{L}_{ss}^t$, respectively.} 
	\label{fig:trn-stab}
\end{figure*}

%by comparing the improvements on three datasets (improved by -0.1\%, 0.6\%, and 1.7\% for ), it is observed that the performance is positively correlated to the dataset size, which further confirms the discussion in the previous paragraph about SRDC and MA.
\begin{figure*}[t]
	\begin{center}
		\begin{subfigure}{.27\linewidth}
			\centering
			\includegraphics[width=1\linewidth]{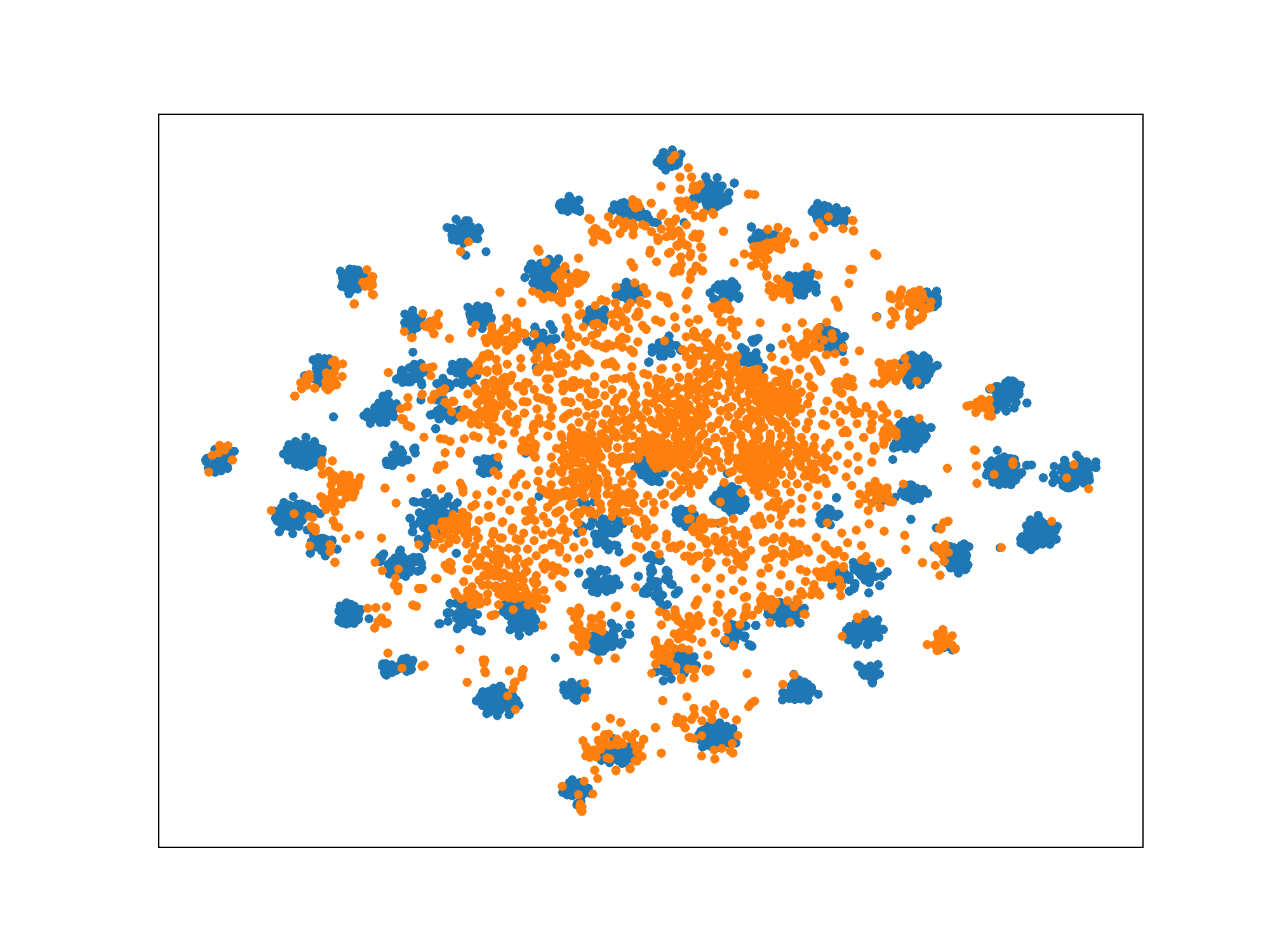}
			\caption{}
		\end{subfigure}
		\begin{subfigure}{.27\linewidth}
			\centering
			\includegraphics[width=1\linewidth]{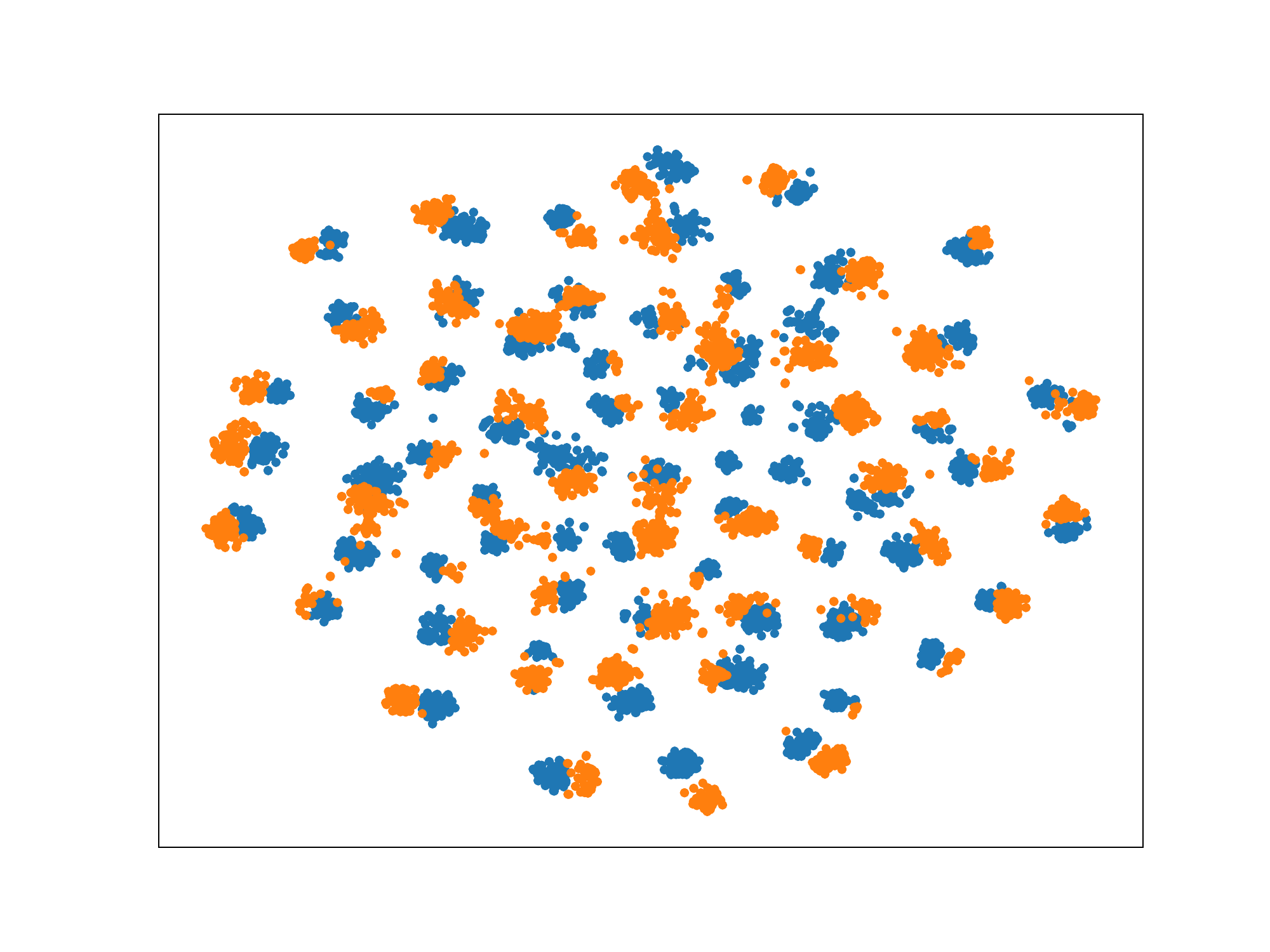}
			\caption{}
		\end{subfigure}	
		\begin{subfigure}{.27\linewidth}
			\centering
			\includegraphics[width=1\linewidth]{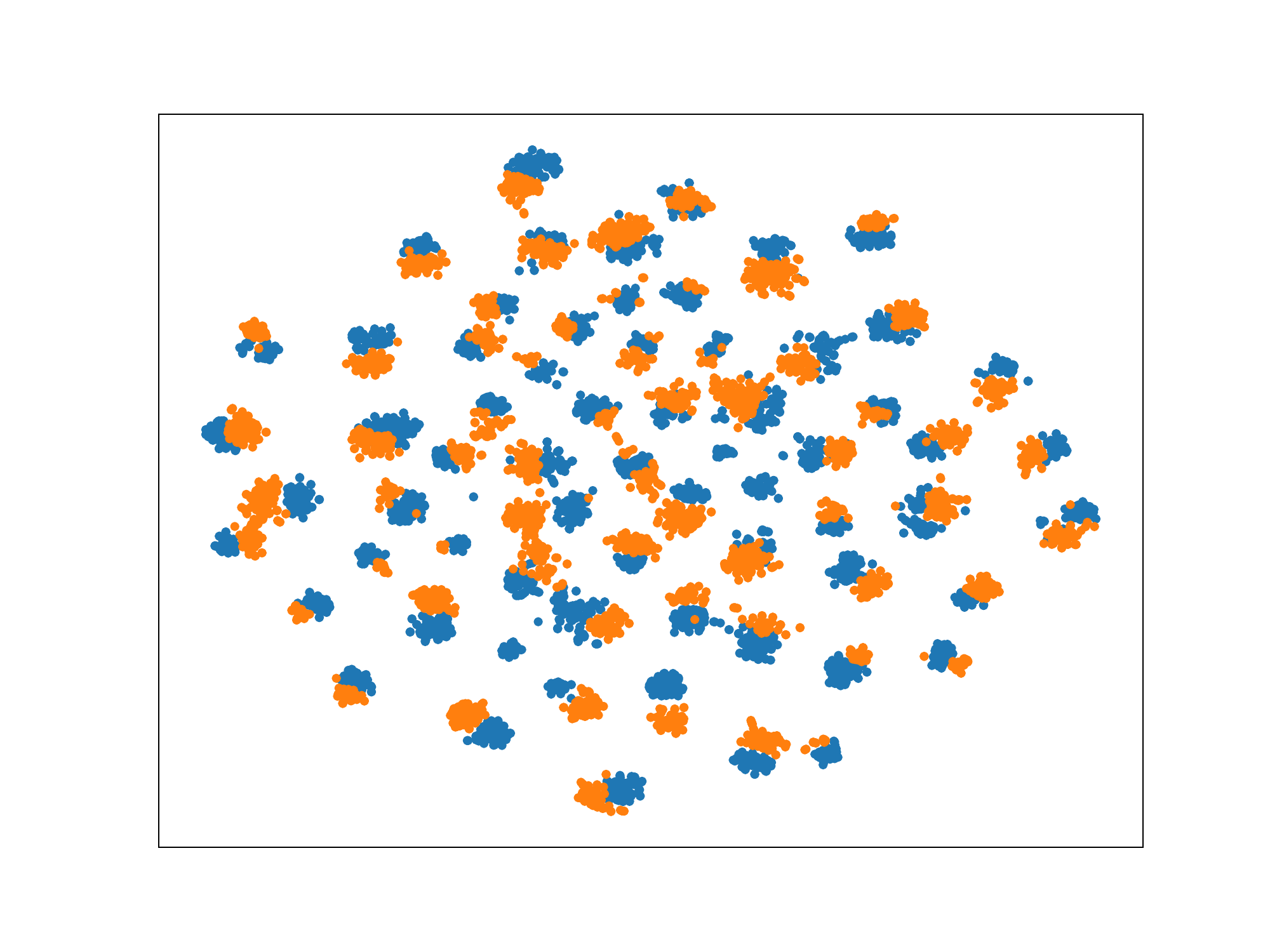}
			\caption{}
		\end{subfigure}\\
		\begin{subfigure}{.27\linewidth}
			\centering
			\includegraphics[width=1\linewidth]{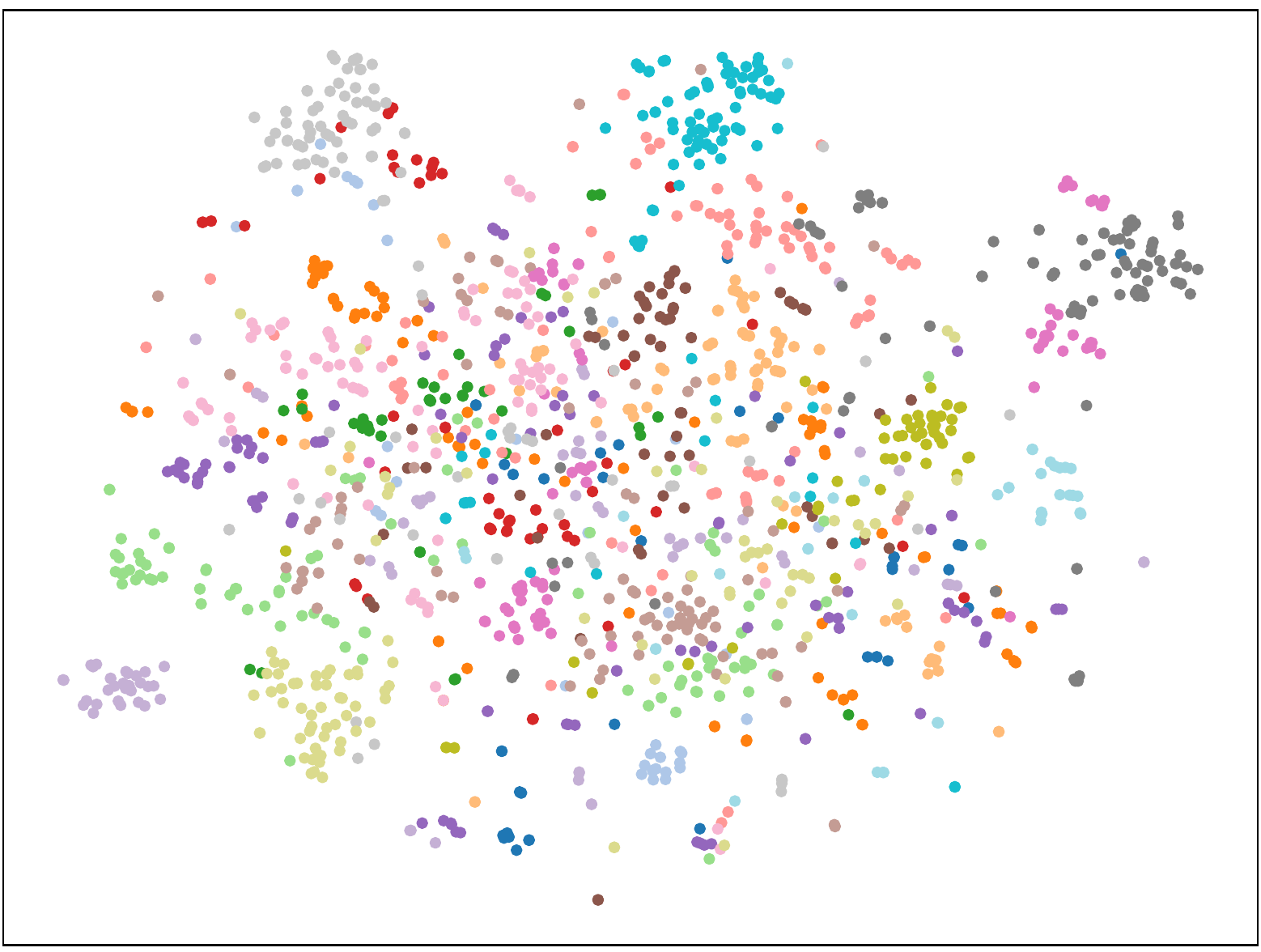}
			\caption{}
		\end{subfigure}	
		\begin{subfigure}{.27\linewidth}
			\centering
			\includegraphics[width=1\linewidth]{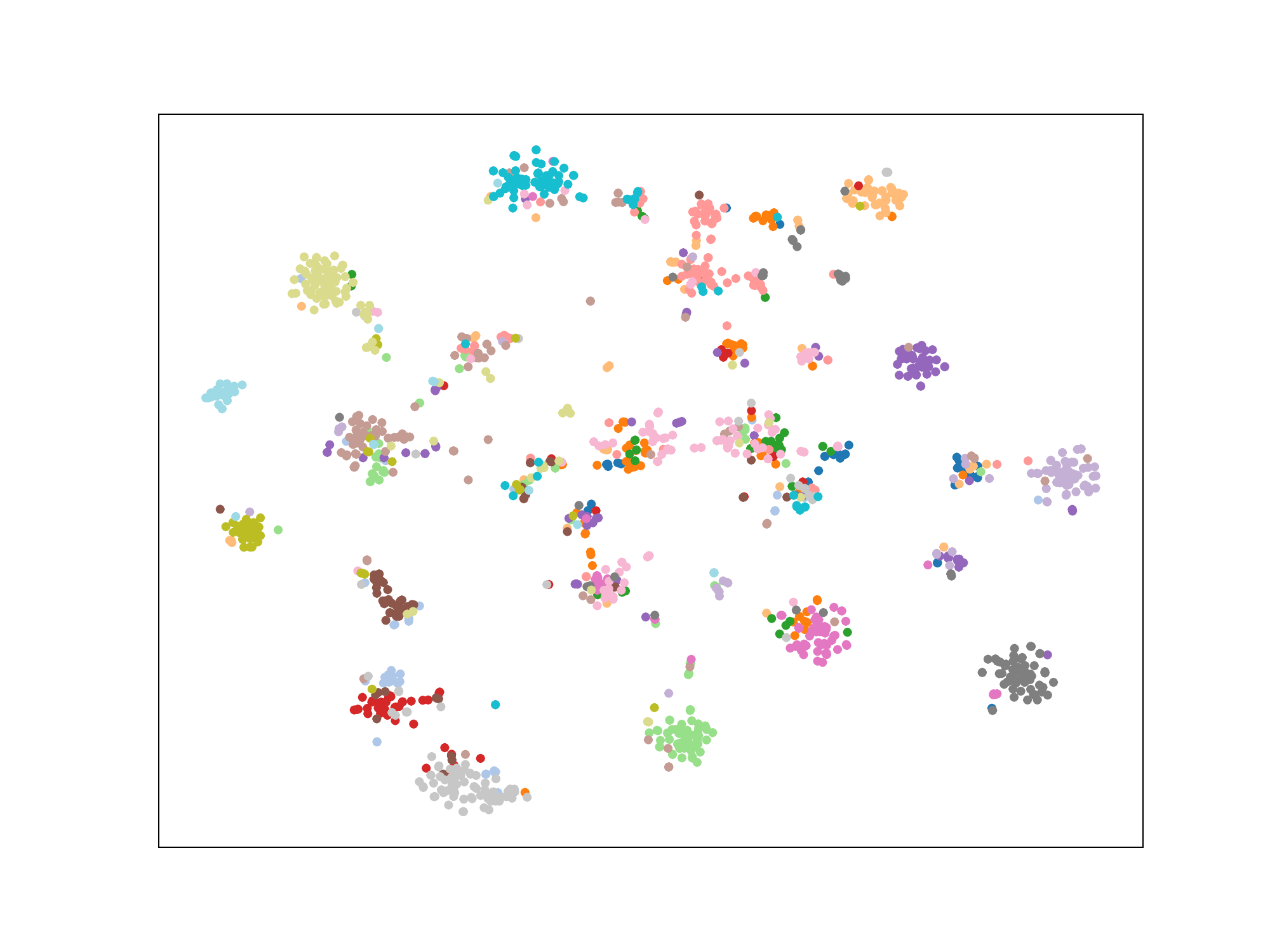}
			\caption{}
		\end{subfigure}
		\begin{subfigure}{.27\linewidth}
			\centering
			\includegraphics[width=1\linewidth]{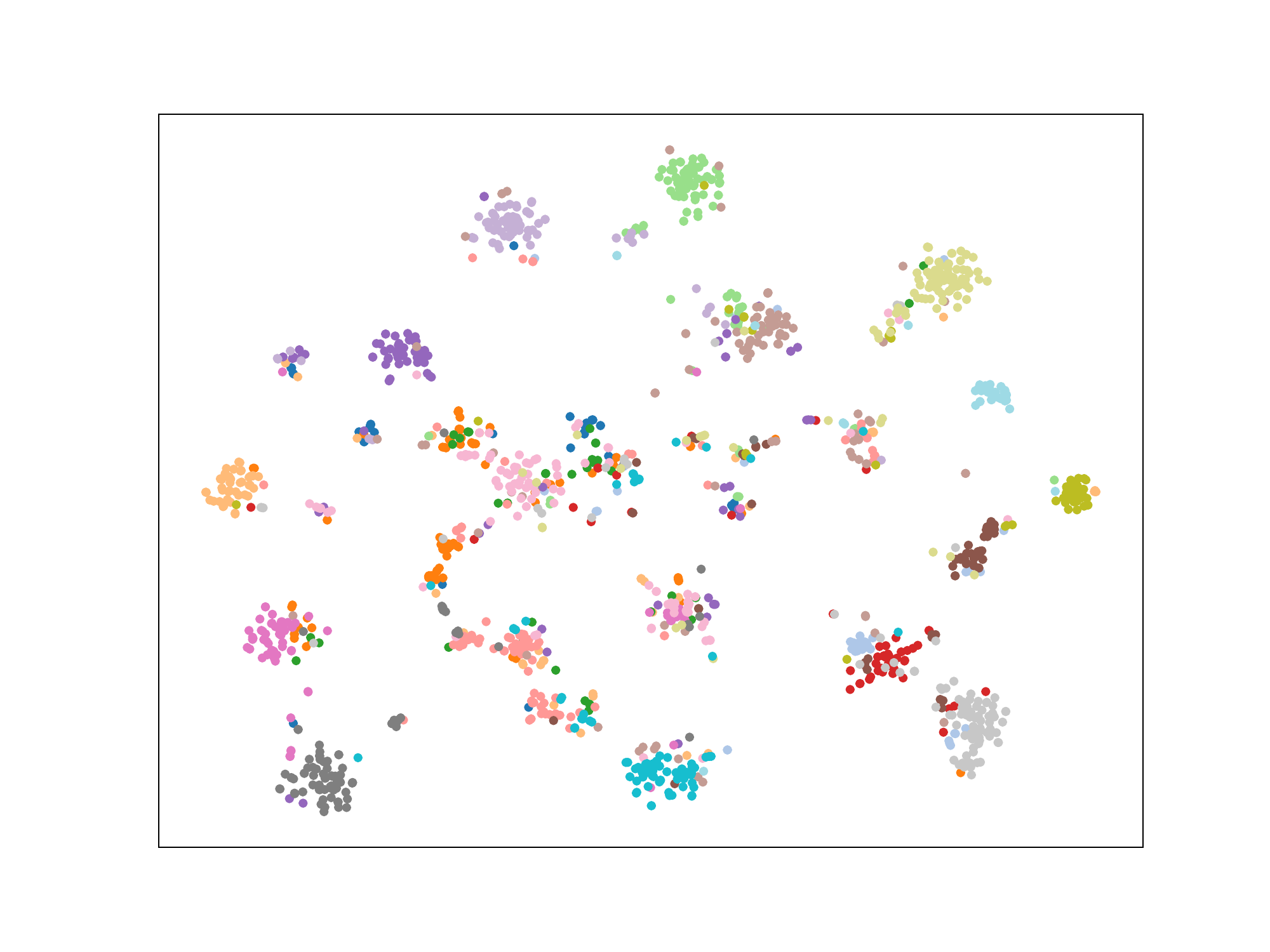}
			\caption{}
		\end{subfigure}
	\end{center}
	\caption{The t-SNE feature visualizations for task Pr$\to$Cl on Office-Home. (a), (b) and (c) present feature alignment between the source data and the target data by the source model, N2DC and N2DC-EX, respectively. (d), (e) and (f) present deep clustering with category information by the source model, N2DC and N2DC-EX, respectively. In (a), (b) and (c), blue circles denote the features of the absent source data, and orange circles denote the target data's features. In (d), (e) and (f), only the first 20 categories in each domain are selected for better illustration, and a different color denotes a different category.}
	\label{fig:vis}
\end{figure*}

On Office-Home (Table~\ref{tab:oh}), our methods exceed other methods. 
Concerning average accuracy, N2DC and N2DC-EX respectively improve by 0.7\% and 1.2\% compared to the previous second-best method SHOT. 
N2DC-EX achieves the best results on 10 out of 12 transfer tasks, while N2DC achieves the second-best results on 7 out of 12 tasks.
Compared to SRDC that is the best UDA method on Office-31, our two methods have an evident improvement of 1.3\% and 1.8\% The results were consistent with our expectations that the larger the dataset used, the better our model's performance.
%and for all the tasks,  our method was significantly better

Experiments on VisDA-C further verified the above trends. As shown in Table~\ref{tab:vc}, both N2DC and N2DC-EX further defeat other methods.
N2DC and N2DC-EX obtain best performance on 10 out of 12 tasks in total. 
N2DC-EX obtains the best results on 8 out of 12 classes and reaches the best accuracy of 85.8\% on average. Also, N2DC ranks second on half the tasks.  
Compared to the second-best method SHOT and BAIT, the average improvement increases to 1.8\% and 3.1\% for N2DC and N2DC-EX. 
Compared to the best SAUDA method MA on Office-31, our two methods improve the average accuracy by at least 2.9\% from 81.6\%.
%N2O Net surpasses the best source data-absent method MA on \textbf{Office-31} by $2.8\%$ in average accuracy.    
%Compared to the previous best UDA method MCC, N2O Net improved $5.6\%$ in average accuracy. 
In our opinion, the evident advantage on VisDA-C is reasonable. On large datasets, due to the increased amount of data, more comprehensive semantic information and more finely portrayed geometric information is available to support our NNH-based deep cluster.

Compared with the SHOT results reported in the three tables, our two methods are equivalent to or better than SHOT on all tasks except for four situations in total. 
For N2DC, the exceptions include $A \to W$ and $W \to A$ on Office-31, and $W \to D$ on Office-31 and the class 'truck' on VisDA-C for N2DC-EX.  
This indicates that the deep cluster based on NNH is more robust than the cluster taking individual data as the fundamental clustering unit. 

Also, N2DC-EX surpasses N2DC on the three datasets in average accuracy. On Office-Home, N2DC-EX improves by 0.5\% and improves by at least 1.3\% on the other two datasets.
N2DC beat N2DC-EX in only three situations, including $W \to D$ on Office-31, Rw $\to$ Ar on Office-Home and class 'truck' on VisDA-C. Besides class 'truck' with a gap of 2.4\%, there are narrow gaps (up to 0.2\%) in two other situations. 
This comparison between N2DC and N2DC-EX indicates that introducing the semantic credibility used in SHNNH is an effective means to boost our NNH-based deep cluster.

%All the experiments were implemented by \textbf{Pytorch} and run on a single \textbf{NVIDIA RTX TITAN}.
%In order to exclude the influence of random factors, we run the code\footnote{The code is available on: https://github.com/DouMM/N2GONet/} at three different random-seeds ($2019, 2020, 2021$) as done in~\cite{2020Do}. Moreover, under each seed, the method repeatedly runs for 10 rounds. We took the average value as the final result.
%The method repeatedly runs for 10 rounds. We took the average value as the final results.

\subsection{Analysis}
In this subsection, we analyze our method from the following three aspects for a complete evaluation.
To support our analysis, we select two hard transfer tasks as toy experiments, including task Pr$\to$Cl on Office-Home and task W$\to$A on Office-31. The two tasks have the worst accuracies in their respective datasets.
\begin{figure}[h]
	\begin{center}
		\begin{subfigure}{.46\linewidth}
			\centering
			\includegraphics[width=1\linewidth]{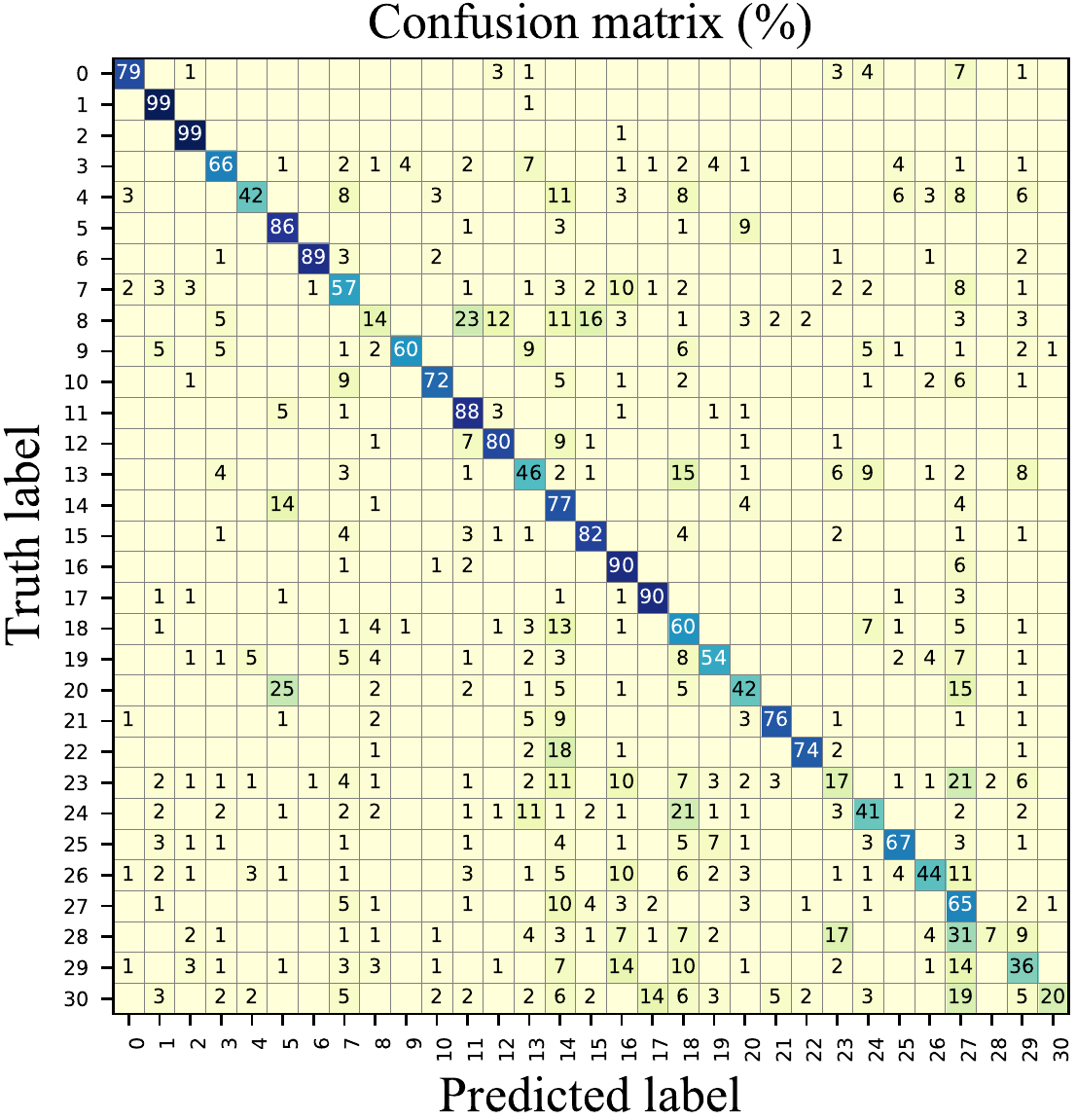}
			\caption{}
		\end{subfigure}
		\begin{subfigure}{.46 \linewidth}
			\centering
			\includegraphics[width=1\linewidth]{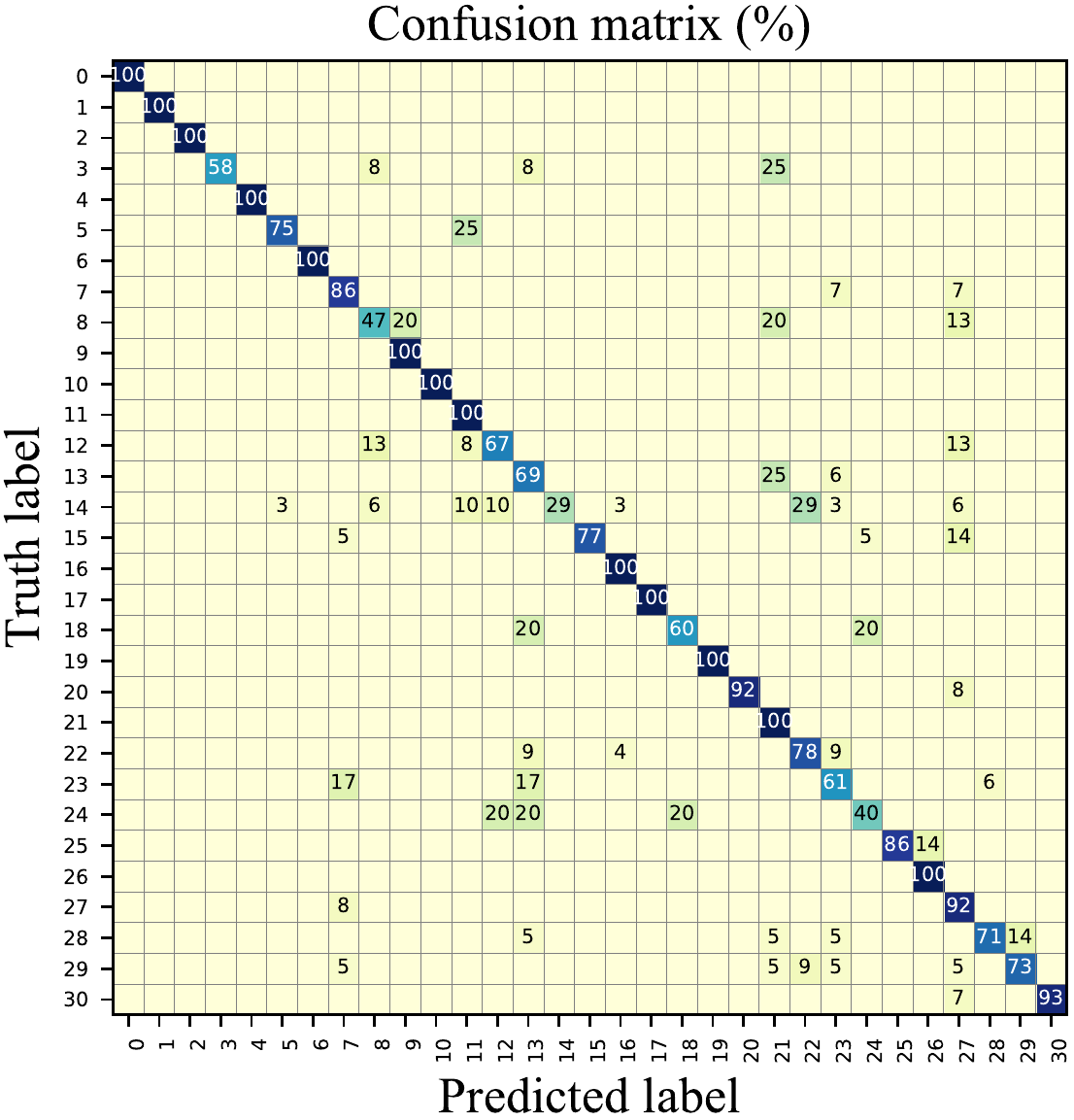}
			\caption{}
		\end{subfigure}\\
		\begin{subfigure}{.46\linewidth}
			\centering
			\includegraphics[width=1\linewidth]{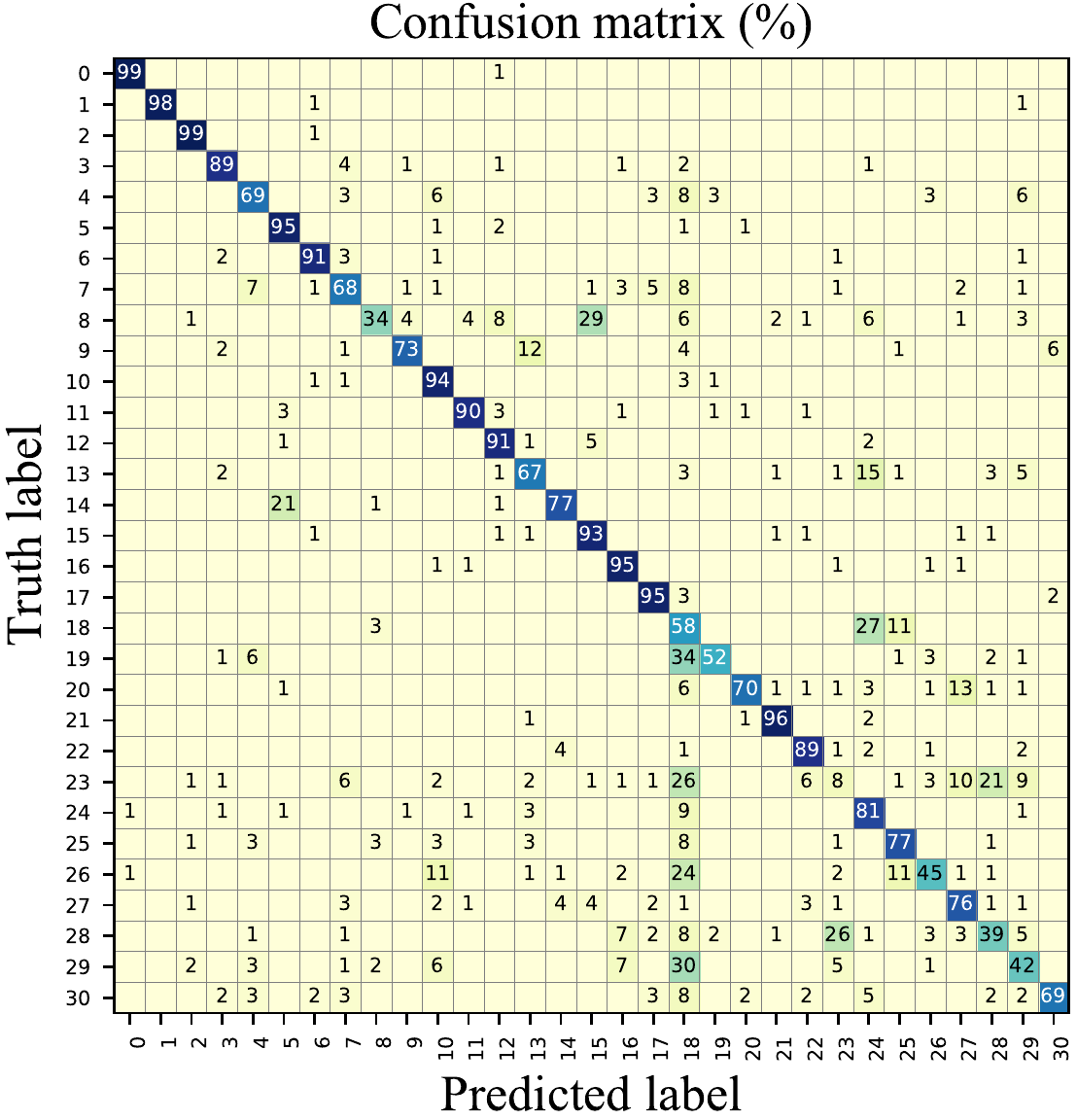}
			\caption{}
		\end{subfigure}
		\begin{subfigure}{.46\linewidth}
			\centering
			\includegraphics[width=1\linewidth]{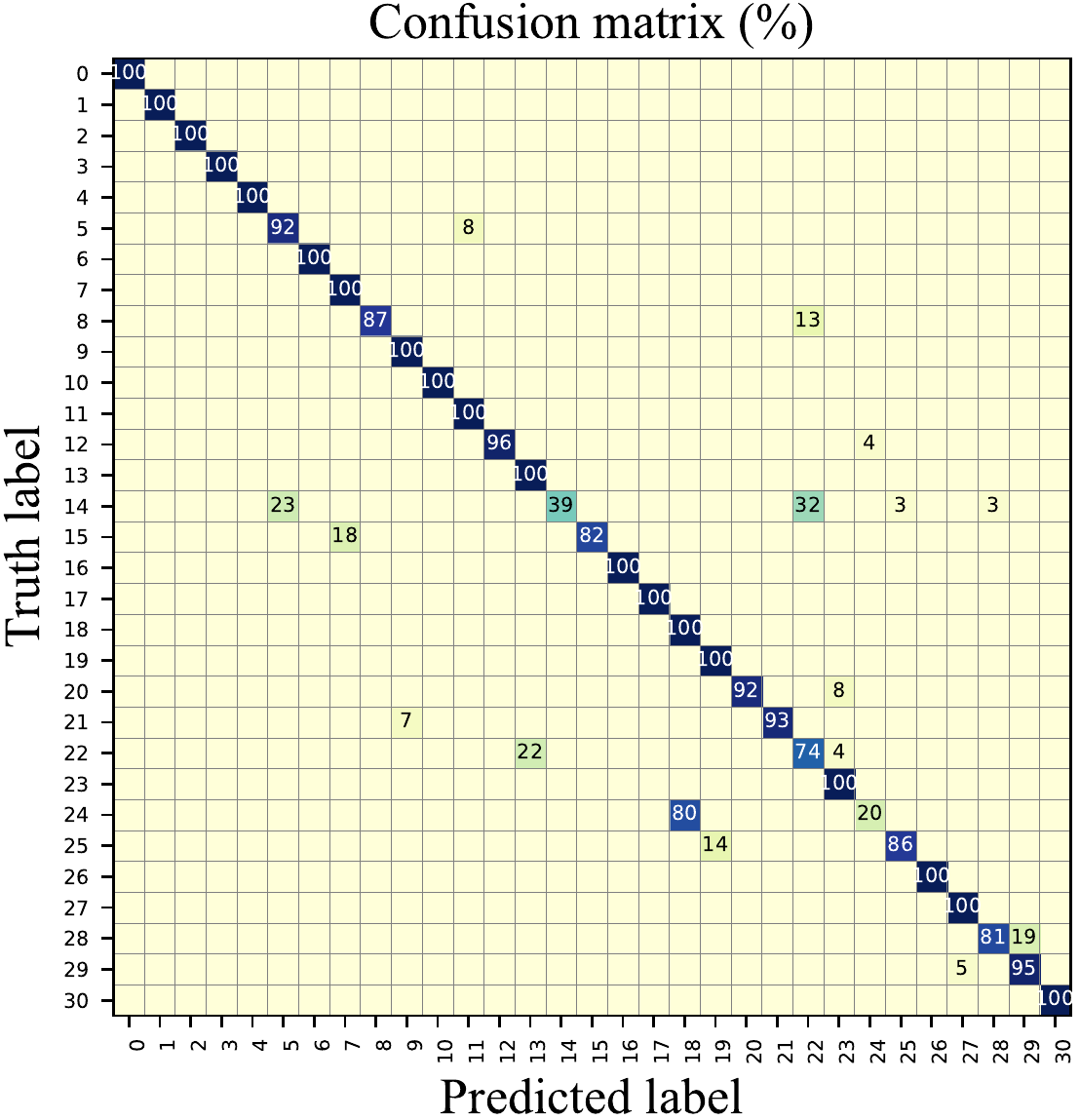}
			\caption{}
		\end{subfigure}\\
		\begin{subfigure}{.46\linewidth}
			\centering
			\includegraphics[width=1\linewidth]{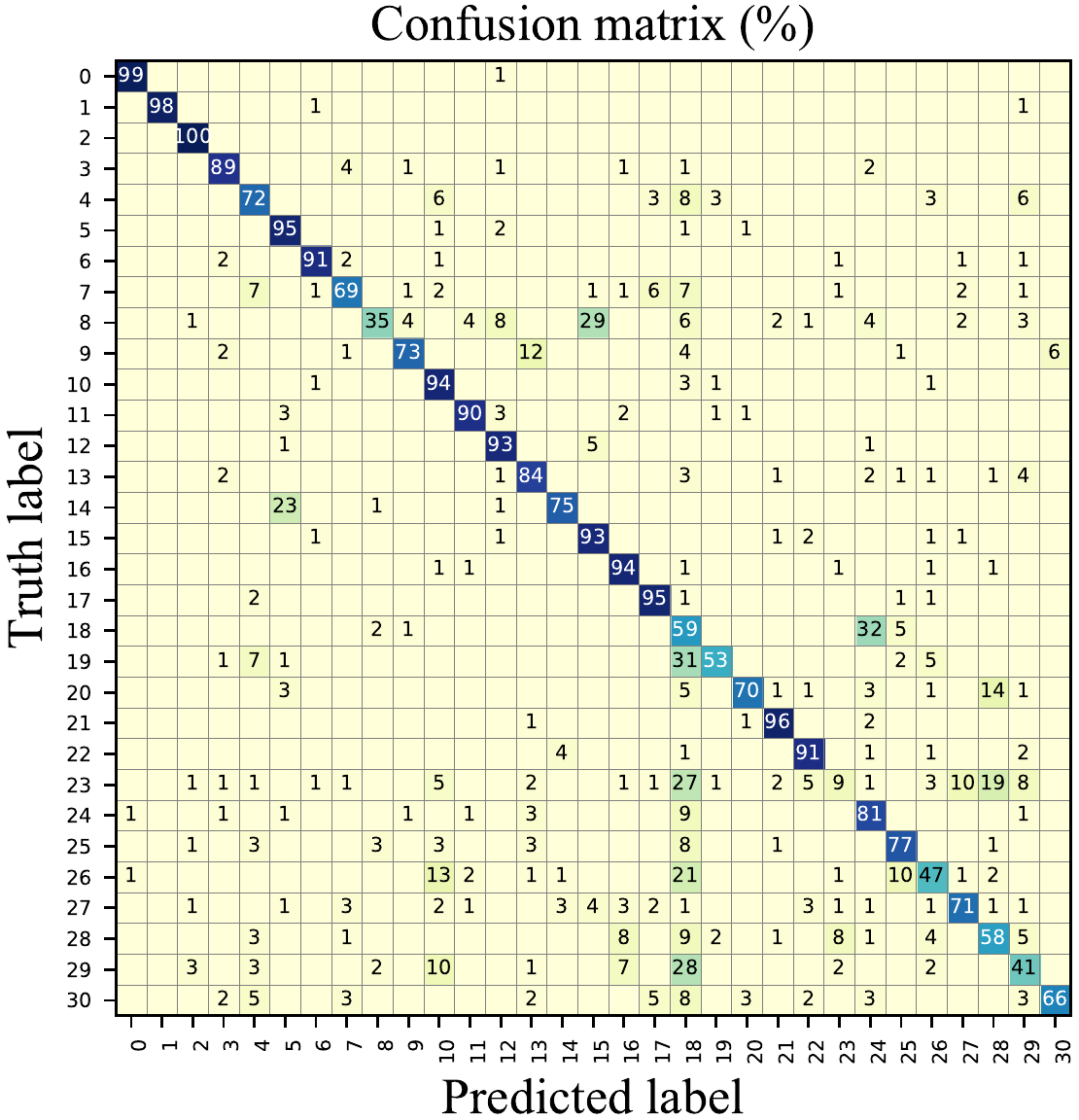}
			\caption{}
		\end{subfigure}
		\begin{subfigure}{.46\linewidth}
			\centering
			\includegraphics[width=1\linewidth]{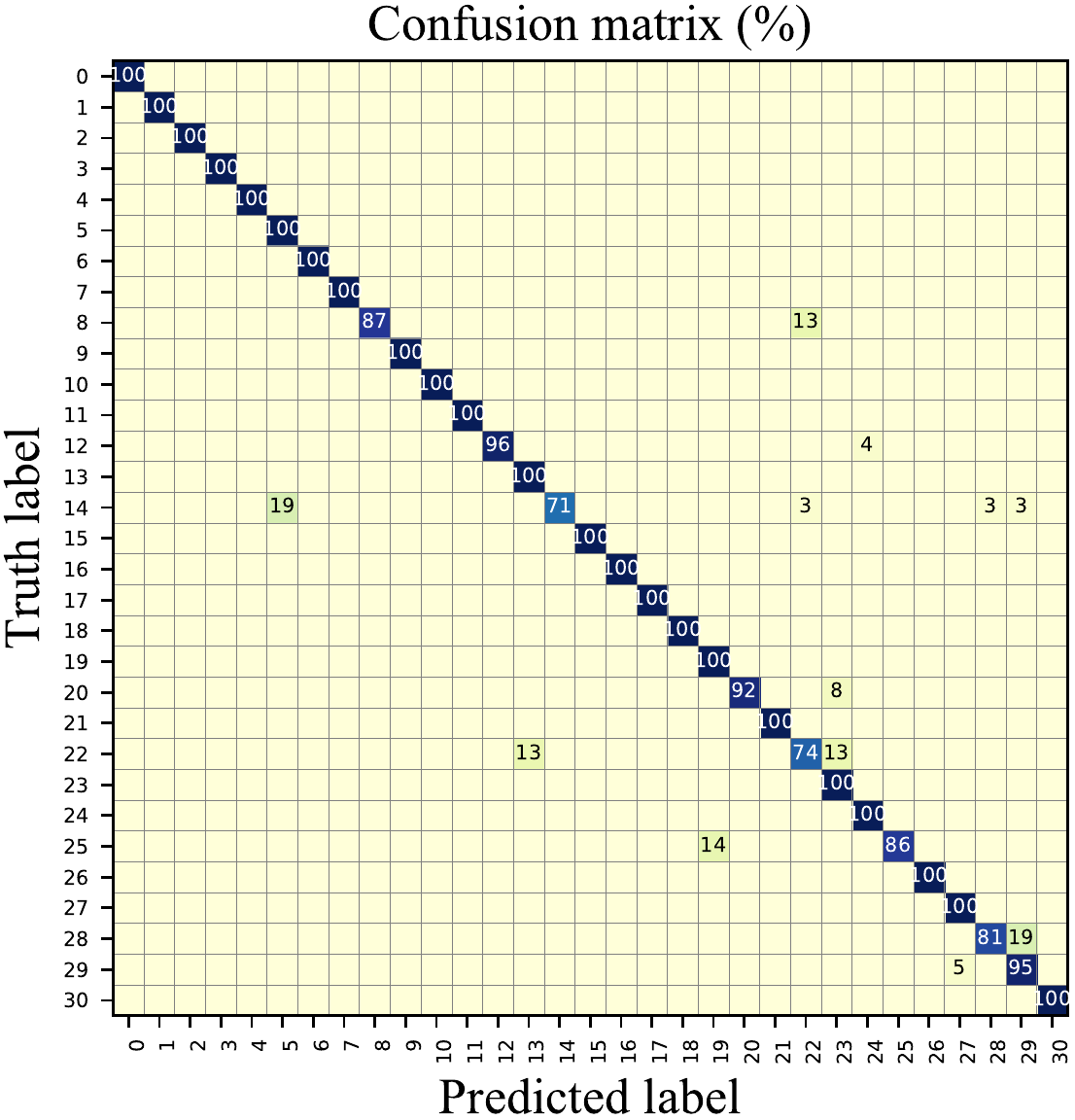}
			\caption{}
		\end{subfigure}
	\end{center}
	\caption{The confusion matrix for 31-way classification task W$\to$A and A$\to$W on Office-31 dataset. \textbf{Left column}: (a), (c) and (e) present the results of the source model, N2DC and N2DC-EX in task W$\to$A, respectively. \textbf{Right column}: (b), (d) and (f) present the results of the source model, N2DC and N2DC-EX in task A$\to$W, respectively.}
	\label{fig:confu-mtx}
\end{figure}

\textbf{Training stability}.
In Fig.~\ref{fig:trn-stab}(a), we display the evolution of the accuracy of our two methods during the model adaptation for task Pr$\to$Cl where SHOT is the baseline.  
As the iteration increases, N2DC and N2DC-EX stably climb to their best performance. 
%It is also seen that N2DC-EX not only has better accuracy but also has faster convergence speed.
It is also seen that the accuracy of N2DC surpasses SHOT when the iteration is greater than 110, and N2DC-EX beat SHOT at an early phase (at about the iteration of 20).
Correspondingly, the loss value of $\mathcal{L}_{g_t\circ u_t}^{t}$ (Fig.~\ref{fig:trn-stab}(b)), $\mathcal{L}_{ss}^t$ (Fig.~\ref{fig:trn-stab}(c)) and $\mathcal{L}_{im}^t$ (Fig.~\ref{fig:trn-stab}(d)) continuously decreases. This is compliant with the accuracy variety shown in Fig.~\ref{fig:trn-stab}(a).

\textbf{Feature visualization}.
Based on the 65-way classification results of task Pr$\to$Cl, we visualize the feature distribution in the low-dimensional feature space by t-SNE tool.
As shown in Fig.~\ref{fig:vis}(b)(c), compared to the results obtained by the source model (Fig.~\ref{fig:vis}(a)), N2DC and N2DC-EX align the target features to the source features. Moreover, the features learned by both N2DC (Fig.~\ref{fig:vis}(e)) and N2DC-EX (Fig.~\ref{fig:vis}(f)) perform a deep clustering with evident category meaning.

\textbf{Confusion matrix}.
For a clear view of the figure, we change our experiment to task W$\to$A that has half the categories of Office-Home.
In addition, we present the results of symmetrical task A$\to$W as comparison. 
Fig.~\ref{fig:confu-mtx} investigates the confusion matrices of the source model, N2DC and N2DC-EX, for the two tasks.
From the left column of Fig.~\ref{fig:confu-mtx}, we observe that both N2DC and N2DC-EX have evidently fewer misclassifications than the source model on task W$\to$A. 
In the right column of Fig.~\ref{fig:confu-mtx}, N2DC-EX exposes its advantages over N2DC.  
From Fig.~\ref{fig:confu-mtx}(d)(f), it is seen that N2DC-EX maintains the performance of N2DC in all categories and further improves in some hard categories on task A$\to$W. For example, as shown in the $24$-th category, N2DC-EX improves the accuracy from 20\% to 100\%.

\textbf{Parameter sensitivity}.
In our method, there are three vital parameters.  
The first is $\beta$ in Eqn.~\eqref{eqn:objctive} that reflects the adjustment from the self-supervision based on the pseudo-labels.
The second is $(\omega_{i},\omega_{in})$ in Eqn.~\eqref{eqn:im-fusion} and $(\eta_{i},\eta_{in})$ in Eqn.~\eqref{eqn:ss} that determines the 
impact intensity of the data constructing NNH in the two regularization items $\mathcal{L}_{im}^t$ and $\mathcal{L}_{ss}^t$, respectively.
The third is the random variable $\lambda$ that describes the semantic fusion on NNH for  pseudo-label generation.
\begin{figure}[t]
	\begin{center}
		\begin{subfigure}{.49\linewidth}D
			\centering
			\includegraphics[width=1\linewidth]{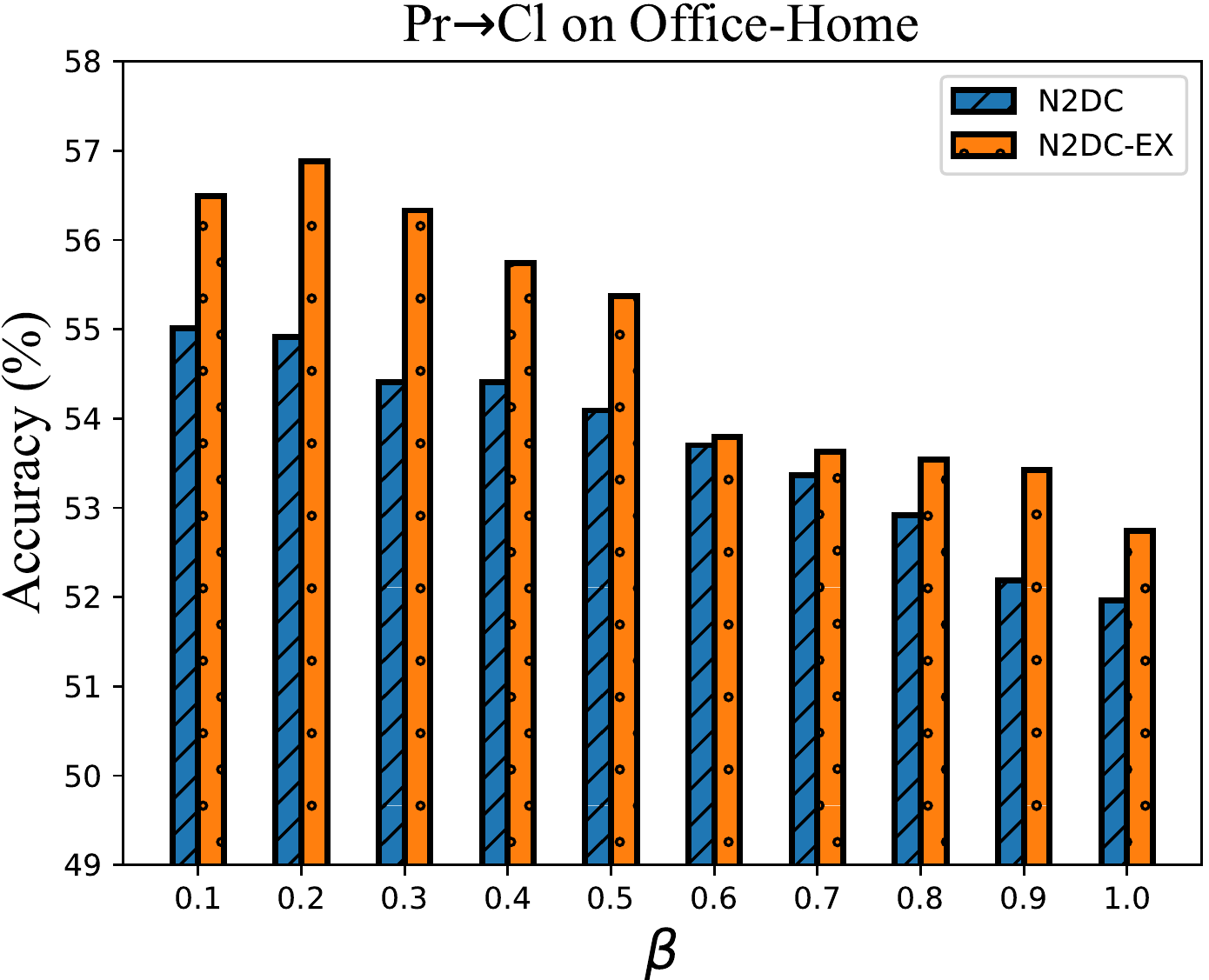}
			\caption{}
		\end{subfigure}
		\begin{subfigure}{.49\linewidth}
			\centering
			\includegraphics[width=1\linewidth]{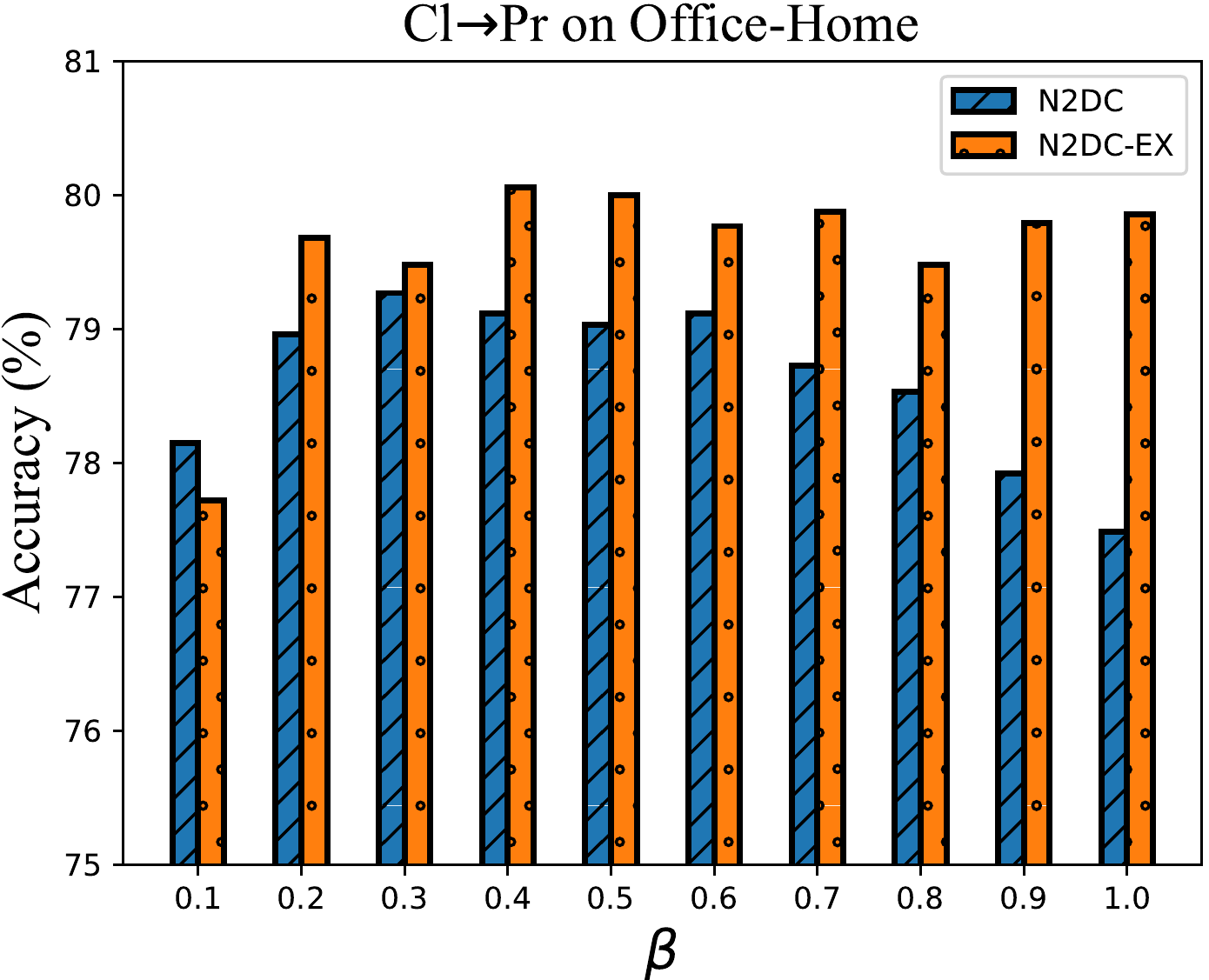}
		    \caption{}
		\end{subfigure}
	\end{center}
	\setlength{\abovecaptionskip}{0mm}
	\setlength{\belowcaptionskip}{-0.2cm}
	\caption{Performance sensitivity of the trade-off parameter $\beta$ in our objective. (a) and (b) present the results for hard task Pr$\to$Cl and easy task Cl$\to$Pr on Office-Home, respectively.}
	\label{fig:para-beta}
\end{figure}
\begin{figure}[b]
	\begin{center}
		\begin{subfigure}{.49\linewidth}
			\centering
			\includegraphics[width=1\linewidth]{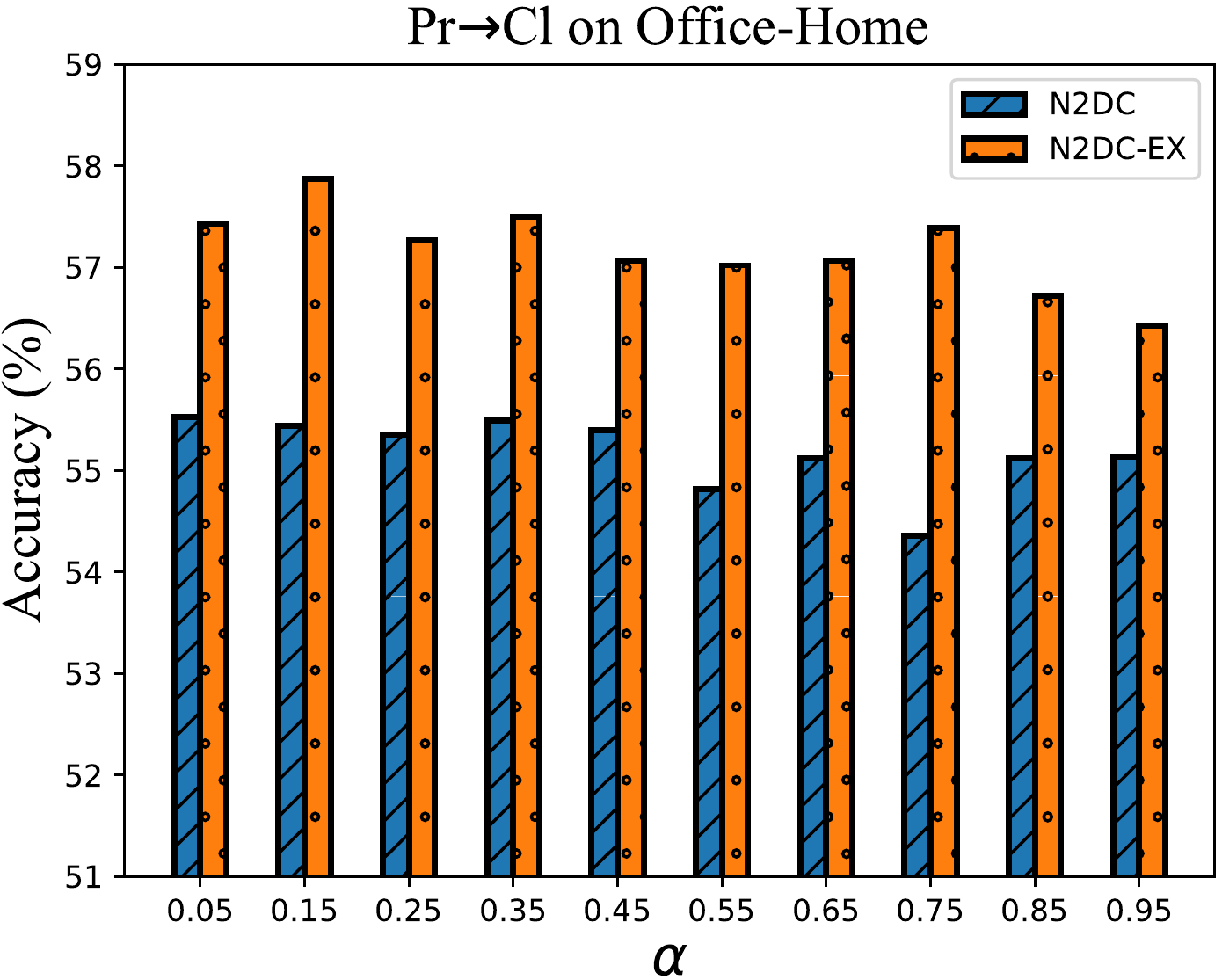}
			\caption{}
		\end{subfigure}
		\begin{subfigure}{.49\linewidth}
			\centering
			\includegraphics[width=1\linewidth]{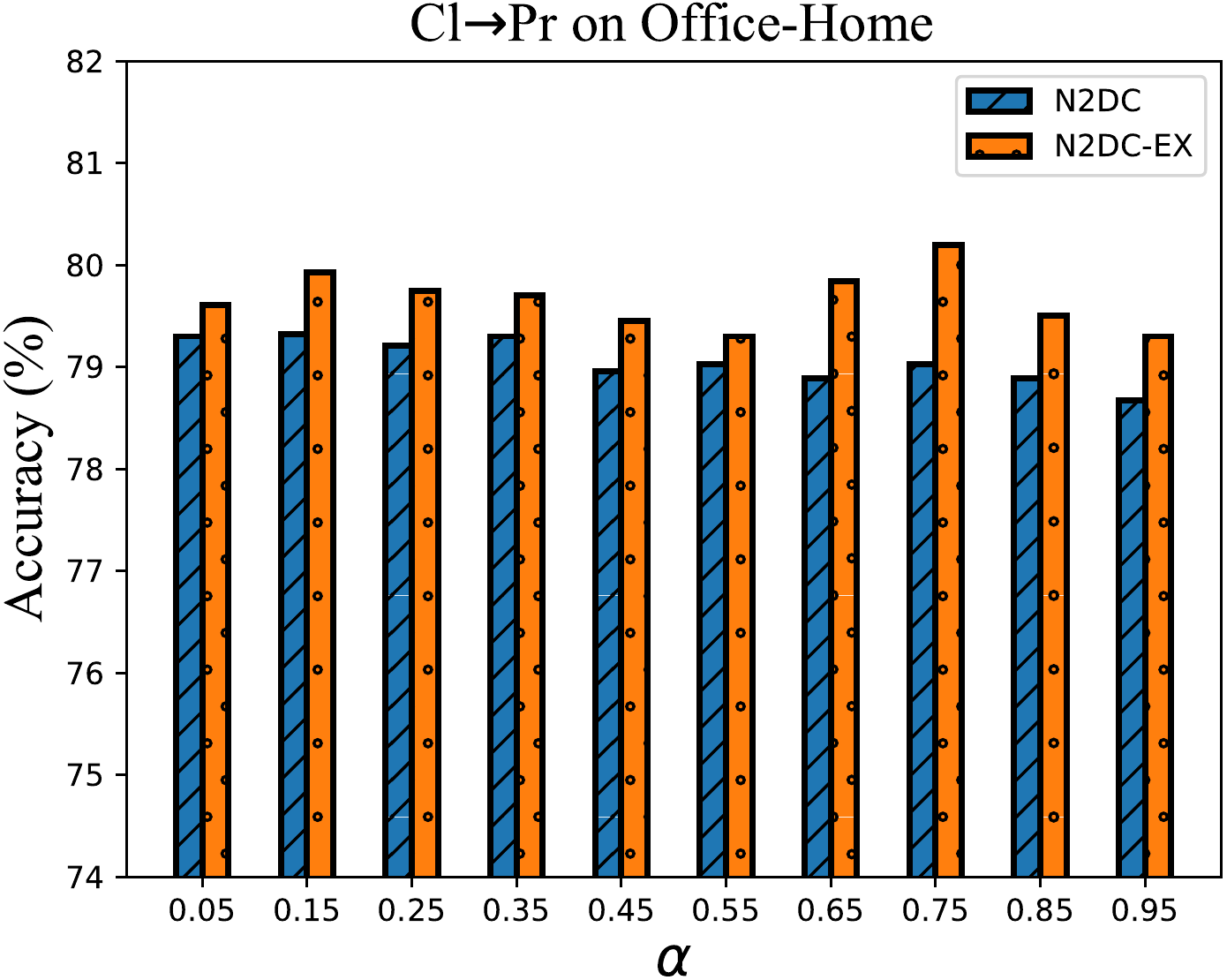}
			\caption{}
		\end{subfigure}
	\end{center}
	\setlength{\abovecaptionskip}{0mm}
	\caption{Performance sensitivity of $\lambda\sim\mathrm{Norm}(\alpha,\delta)$ as the mean $\alpha$ varying with variance $\delta=1-\alpha$. (a) and (b) present the results for hard task Pr$\to$Cl and easy task Cl$\to$Pr on Office-Home, respectively.}
	\label{fig:para-gamma}
\end{figure}

We perform a sensitivity analysis of the parameter $\beta$ in Fig.~\ref{fig:para-beta}(a) for task Pr$\to$Cl. 
N2DC and N2DC-EX respectively reach the best accuracy at $\beta=0.1$ and $\beta=0.2$.
After that, their accuracies decrease gradually as the value of $\beta$ increases.
This phenomenon shows that for this challenging task, the adjustment from the pseudo-labels is very weak.
When the pseudo-labels cannot offer credible category information, an enhancement on self-supervision will deteriorate the final performance.
For comparison, we provide the results of an easy case, the symmetry task Cl$\to$Pr with an accuracy close to 80\%, as shown in Fig.~\ref{fig:para-beta}(b).
N2DC climb to the maximum at a bigger value $\beta=0.3$ and then gradually decrease because the source model has a much better accuracy than the Pr$\to$Cl task that results in pseudo-labels with more credible category information. 
Different from N2DC, N2DC-EX still reaches a robust accuracy at $\beta=0.2$ as shown in Fig.~\ref{fig:para-beta}(a) and holds the performance after that. 
This variation shows that N2DC-EX has a better robustness of parameter $\beta$ compared to N2DC on the easy task. 
%This phenomenon is understandable because SHNNH adopt in N2DC-EX has considered semantic credibility.
%Based on the analysis above, considering that the transfer tasks in SAUDA are without any background knowledge, in practice, we set $\beta$ by small value for low risk. 

Fig.~\ref{fig:para-gamma} investigates the performance sensitivity of parameter $\lambda\sim\mathrm{Norm}(\alpha,\delta)$ when the mean $\alpha$ changes from 0.05 to 0.95 (the variance $\delta=1-\alpha$). On both hard and easy tasks, our two methods do not have a large accuracy decrease. This observation indicates that the semantic fusion in our method for pseudo-label generation is a robust operation.
\begin{figure*}[t]
	\begin{center}
		\begin{subfigure}{.245\linewidth}
			\centering
			\includegraphics[width=1\linewidth]{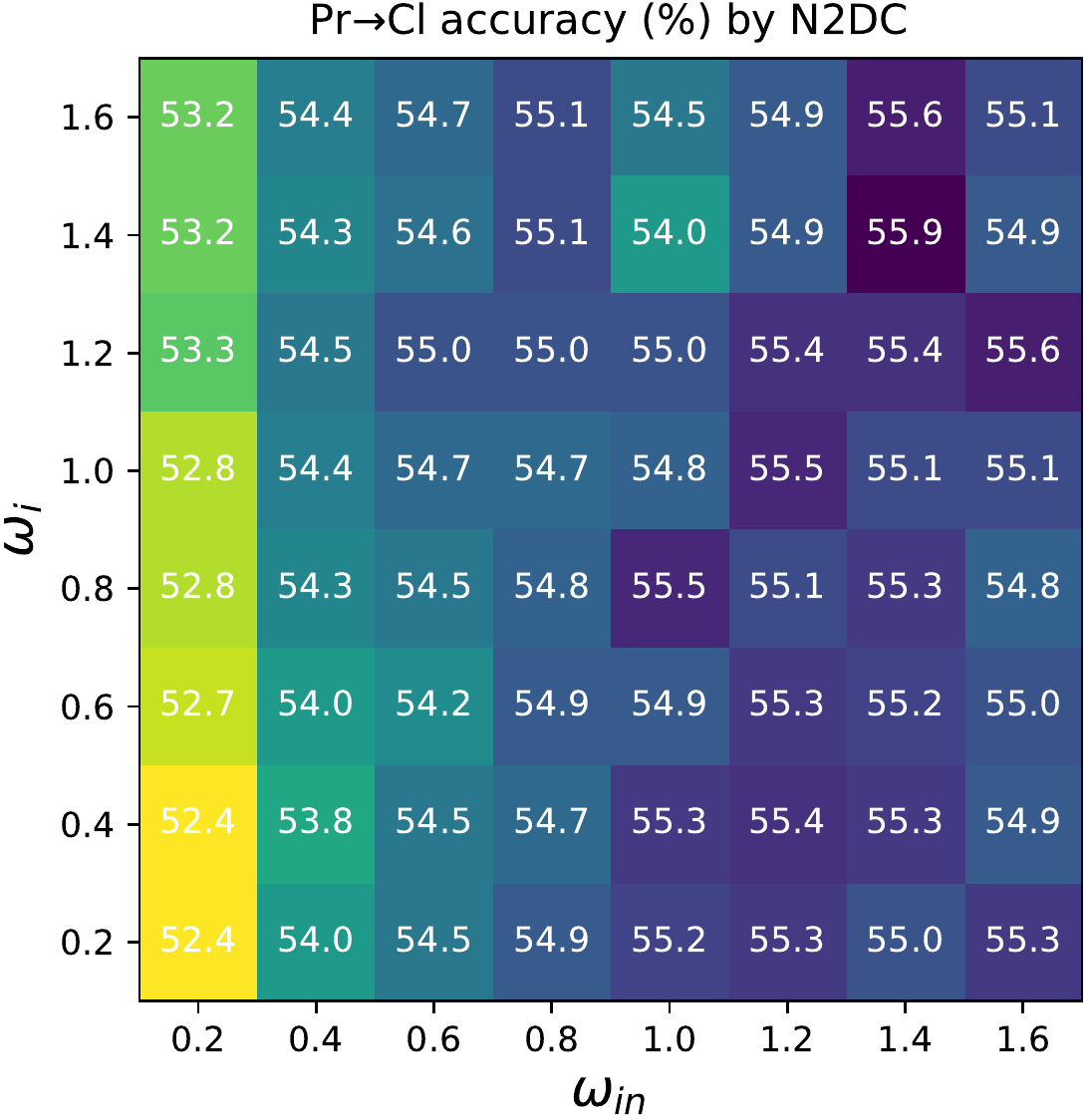}
			\caption{}
		\end{subfigure}
		\begin{subfigure}{.245\linewidth}
			\centering
			\includegraphics[width=1\linewidth]{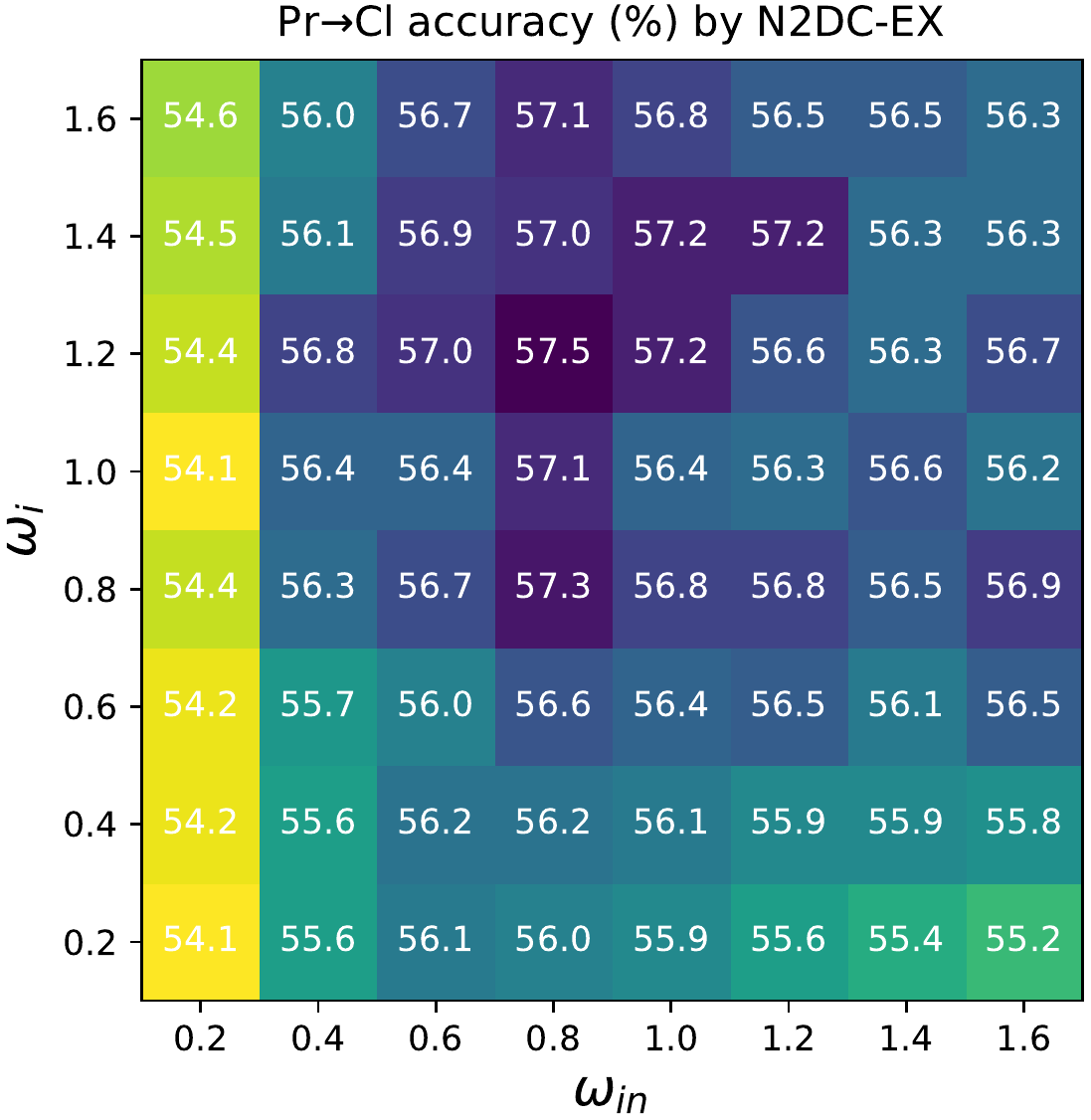}
			\caption{}
		\end{subfigure}	
		\begin{subfigure}{.245\linewidth}
			\centering
			\includegraphics[width=1\linewidth]{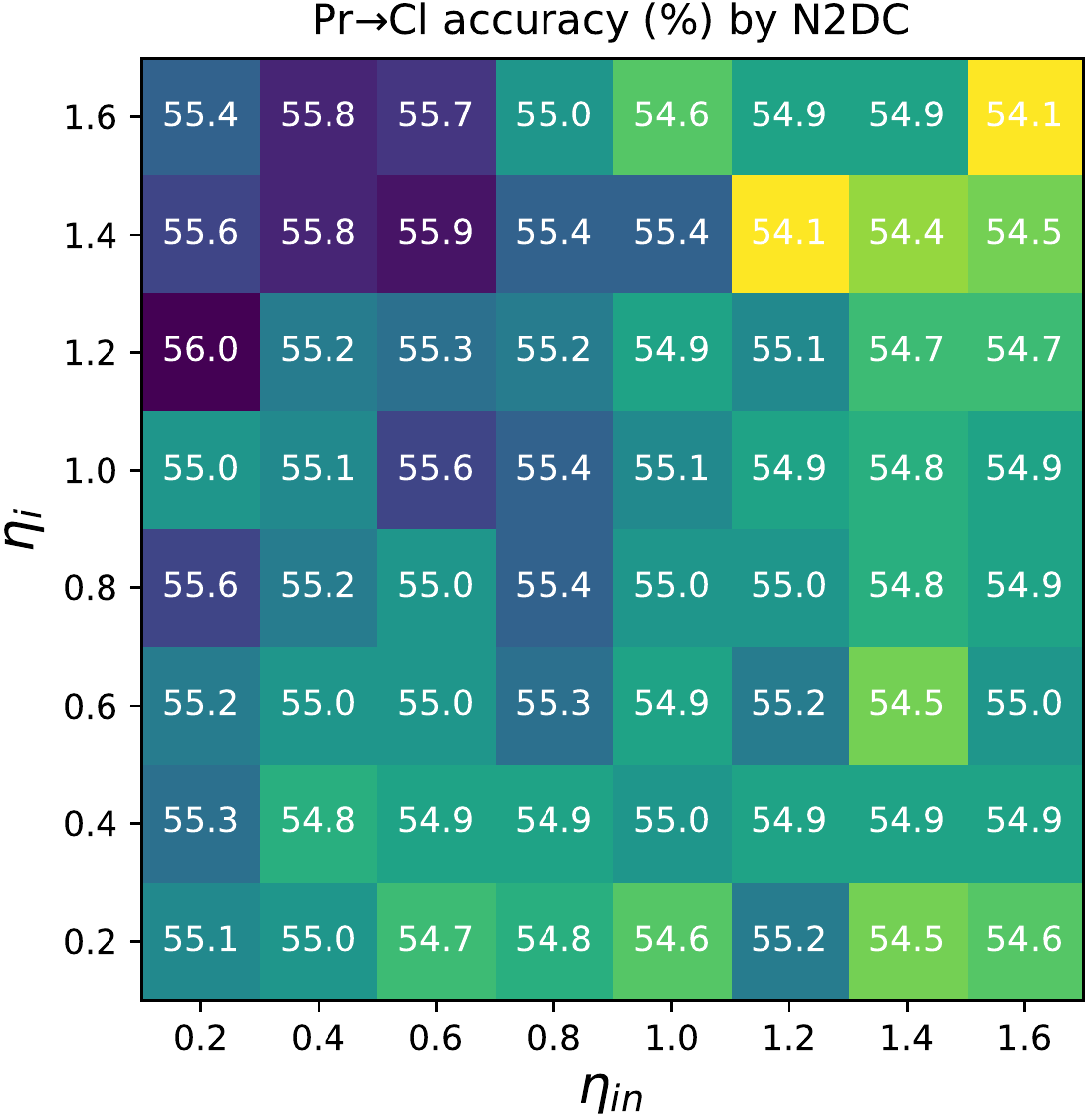}
			\caption{}
		\end{subfigure}
		\begin{subfigure}{.245\linewidth}
			\centering
			\includegraphics[width=1\linewidth]{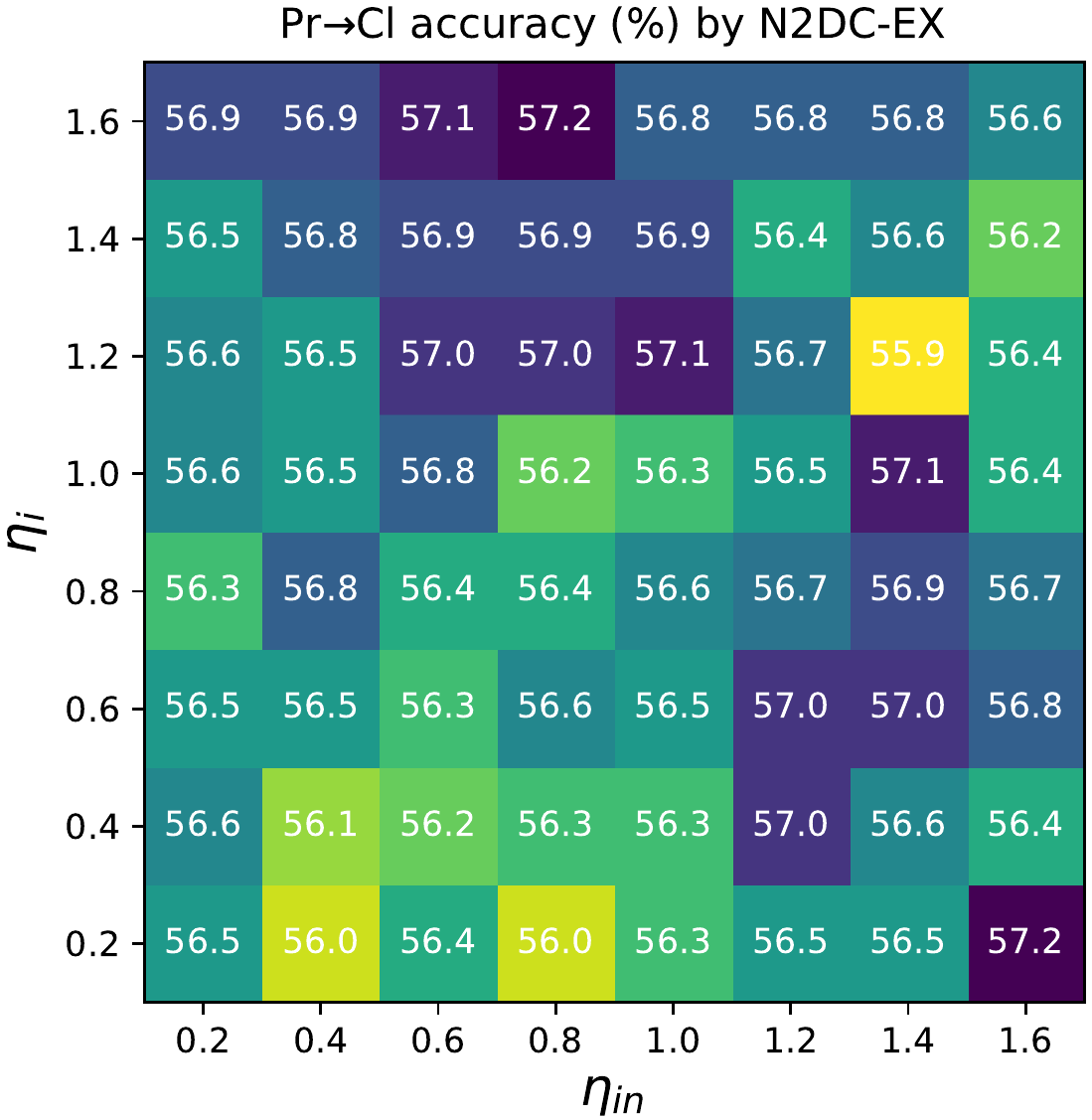}
			\caption{}
		\end{subfigure}\\
		\begin{subfigure}{.245\linewidth}
			\centering
			\includegraphics[width=1\linewidth]{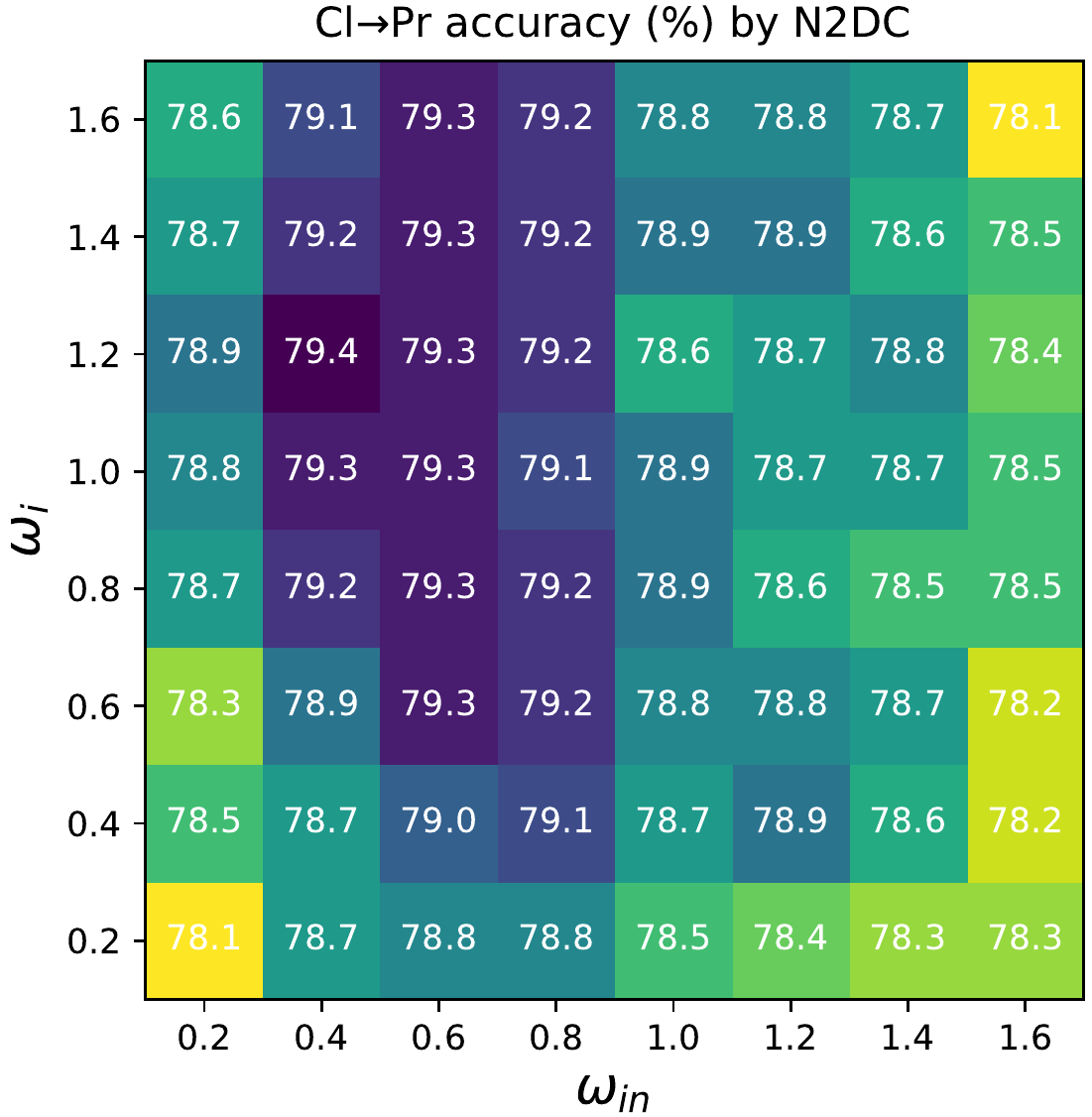}
			\caption{}
		\end{subfigure}	
		\begin{subfigure}{.245\linewidth}
			\centering
			\includegraphics[width=1\linewidth]{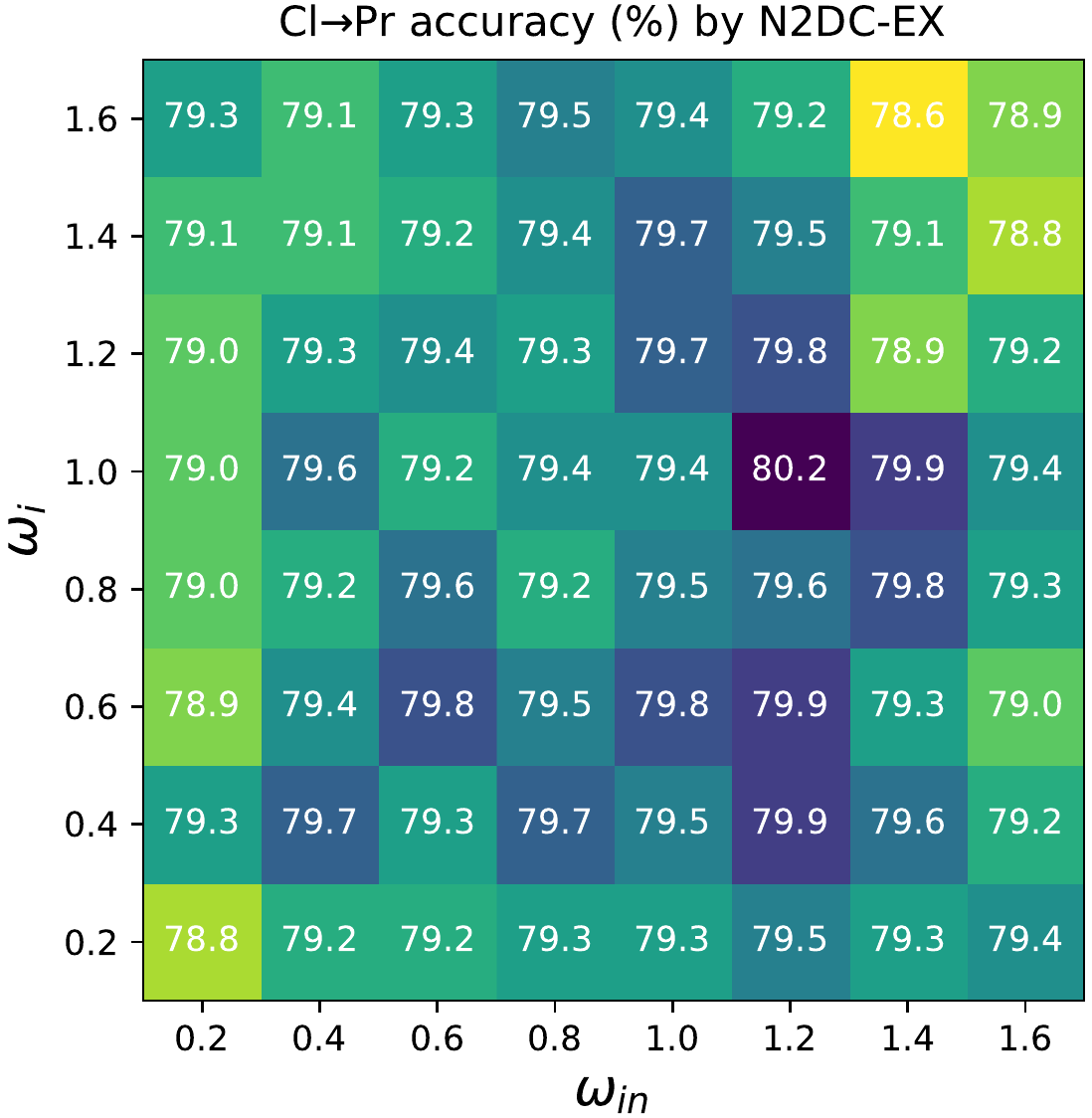}
			\caption{}
		\end{subfigure}	
		\begin{subfigure}{.245\linewidth}
		    \centering
			\includegraphics[width=1\linewidth]{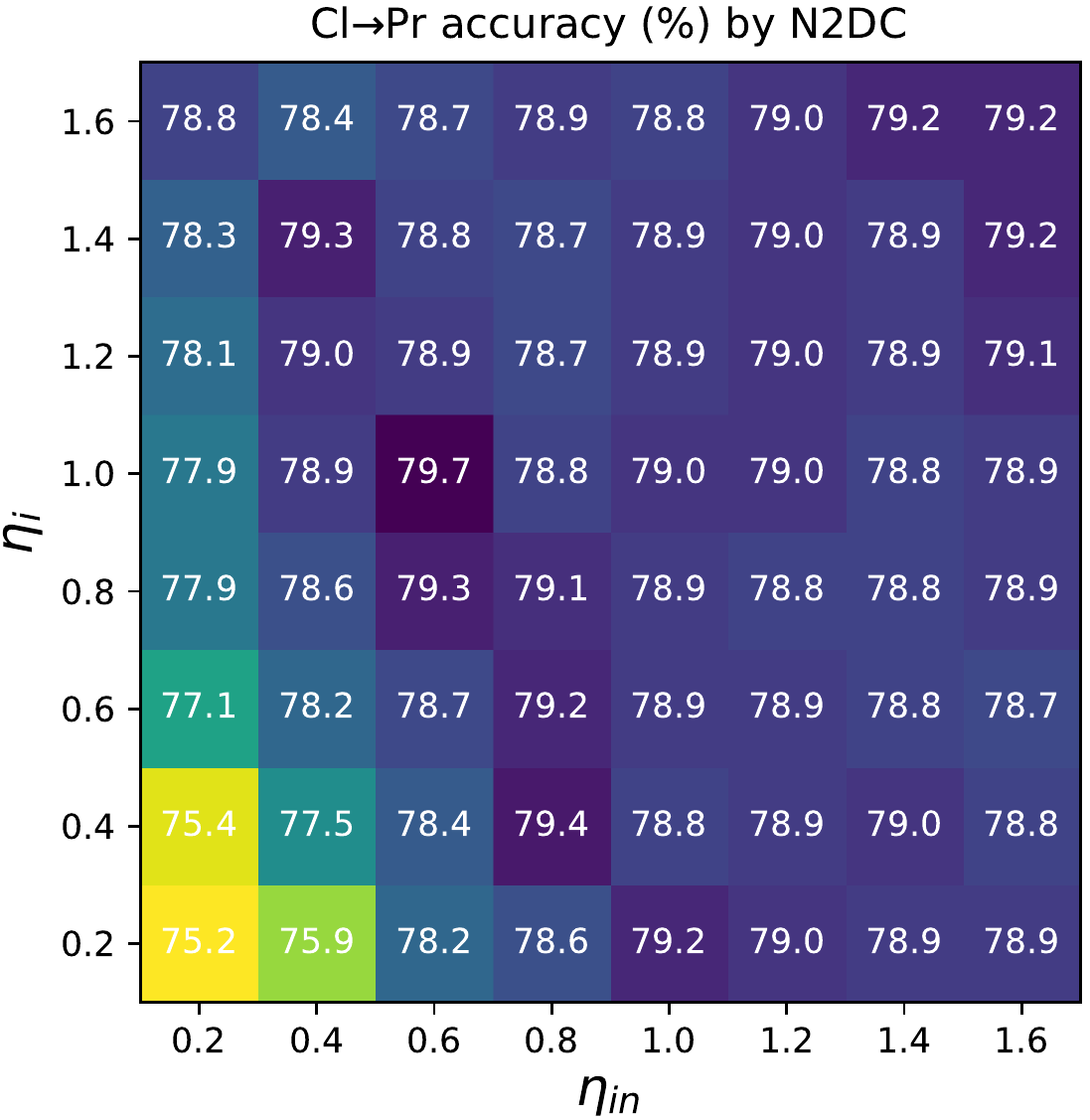}
			\caption{}
		\end{subfigure}
		\begin{subfigure}{.245\linewidth}
			\centering
			\includegraphics[width=1\linewidth]{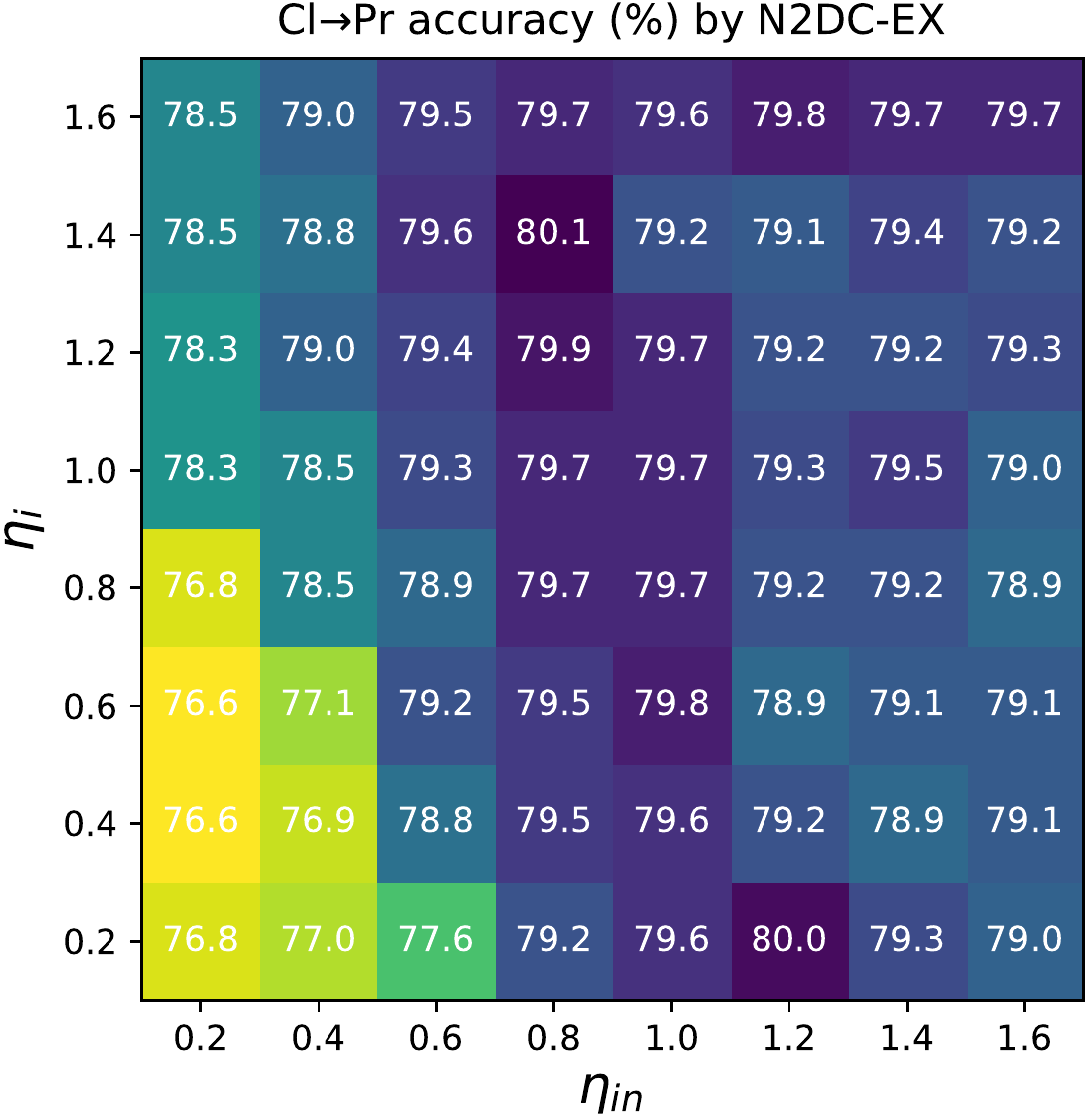}
			\caption{}
		\end{subfigure}
	\end{center}
	\setlength{\abovecaptionskip}{0mm}
	\setlength{\belowcaptionskip}{0cm}
	\caption{Performance sensitivity of $(\omega_{i},\omega_{in})$ and $(\eta_{i},\eta_{in})$. \textbf{Top}: the results for task Pr$\to$Cl on Office-Home. (a)(b) show the results of our two methods with $(\omega_{i},\omega_{in})$ varying. The results when $(\eta_{i},\eta_{in})$ vary are presented in (c)(d).
	\textbf{Bottom}: the results for task Cl$\to$Pr. 
	(e)(f) show the results as $(\omega_{i},\omega_{in})$ vary while (g)(h) show the results as $(\eta_{i},\eta_{in})$ vary.}
	\label{fig:para-sen}
\end{figure*}

To better understand the effects of $(\omega_{i},\omega_{in})$, and $(\eta_{i},\eta_{in})$, we present their performance sensitivity based on task pair Pr$\to$Cl and Cl$\to$Pr. 
Letting $\omega_{i}$ and $\omega_{in}$ vary from 0.2 to 1.6, the top of Fig.~\ref{fig:para-sen} presents the accuracy-matrix of N2DC and N2DC-EX with the obtained 64 parameter pairs on hard task Pr$\to$Cl.
For N2DC, the accuracy-matrix in Fig.~\ref{fig:para-sen}(a) is split by the diagonal, below which these parameter pairs have high accuracies. 
In contrast, for N2DC-EX, the better zone locates upon the diagonal as shown in Fig.~\ref{fig:para-sen}(b). 
Meanwhile, as long as we do not significantly weaken the impact of the input instance, for example, $\omega_i=0.2$, our method's performance will not have an evident decrease.
In Fig.~\ref{fig:para-sen}(c)(d), we give the performance sensitivity results for $(\eta_{i},\eta_{in})$.
The high-performance zone of N2DC in Fig.~\ref{fig:para-sen}(c) is upon the diagonal of the accuracy-matrix, especially the region $\eta_{in}>1$, while N2DC-EX's high-performance zone is symmetric to the diagonal as shown in Fig.~\ref{fig:para-sen}(d). 
Although being smaller than the high accuracy zone of $(\omega_{i},\omega_{in})$, the high accuracy region of $(\eta_{i},\eta_{in})$ is distributed in patches rather than  in isolated parameter pairs.

Similarly, we present the sensitivity results on easy task Cl$\to$Pr at the bottom of Fig.~\ref{fig:para-sen}.
Except for the case shown in Fig.\ref{fig:para-sen}(b) that has relatively weaker parameter robustness, the other three situations are not sensitive to the changing of parameters. 
Meanwhile, it is also seen that on the easy task, our method has a better performance sensitivity than on the hard task.

\begin{table}[h]
    \caption{Ablation study of N2DC on Office-Home dataset. The bold means the best result.}
	\label{tab:ablation}
	\renewcommand\tabcolsep{8pt}
	\renewcommand\arraystretch{1.1}
	\centering
	\small
	%\footnotesize
	%\scriptsize 
	\begin{tabular}{l l c}
		\toprule
		Method   & Ablation operation  & Avg.\\
		\midrule
		Source model only            &---                                              & 59.6\\
		\midrule
		N2DC-no-im &Let $\mathcal{L}_{g_t\circ u_t}^{t}=\beta\mathcal{L}_{ss}^t$ & 69.9\\
		N2DC-no-ss &Let $\mathcal{L}_{g_t\circ u_t}^{t}=\mathcal{L}_{im}^t$      & 71.5\\
		
		N2DC-no-NNH-in-im &Set $\omega_{in}=0$ in Eqn.~\eqref{eqn:im-fusion}  & 72.2\\
		N2DC-no-NNH-in-ss &Set $\eta_{in}=0$ in Eqn.~\eqref{eqn:ss}   & 72.5\\
		
		N2DC-no-fused-pl   &Fix $\lambda_k=1.0$ in Eqn.~\eqref{eqn:fusion-pl}    & 72.4\\
		%N2DC-no-rand-in-pl &Fix $\lambda_k=0.85$ in Eqn.~\eqref{eqn:fusion-pl}   & 72.7\\
		\midrule
		N2DC          &---                                            & \textbf{72.6}\\
		\hline
	\end{tabular}
\end{table}

\subsection{Ablation study}
%\noindent\textbf{A. Ablation study}.~
The ablation study is to isolate the effect of the skills adopted in our methods from three aspects.
For N2DC, to evaluate the effectiveness of the two regularization components $\mathcal{L}_{im}^t$ and $\mathcal{L}_{ss}^t$, of our objective, we respectively delete them from the objective and denote the two edited methods by N2DC-no-im and N2DC-no-ss.
To evaluate the effectiveness of changing the clustering unit from individual data to NNH $\mathcal{H}_i^t$, we give two comparison methods N2DC-no-NNH-in-im and N2DC-no-NNH-in-ss. 
In N2DC-no-NNH-in-im, the influence of $\mathcal{H}_i^t$ on the $\mathcal{L}_{im}^t$ is eliminated by setting $\omega_{in}=0$ in Eqn.~\eqref{eqn:im-fusion} while in N2DC-no-NNH-in-ss the influence on $\mathcal{L}_{ss}^t$ is canceled by setting $\eta_{in}=0$ in Eqn.~\eqref{eqn:ss}.
To evaluate the effectiveness of the semantic-fused pseudo-labels, we cancel the fusion operation presented in Eqn.~\eqref{eqn:fusion-pl} by letting $\lambda_k=1$ for $k=1,2,\cdots,K$.
We denote the new method by N2DC-no-fused-pl. 
\begin{table}[h]
	\caption{Ablation study of N2DC-EX on an Office-Home dataset. The bold means the best result.}
	\label{tab:ablation-ex}
	\renewcommand\tabcolsep{5pt}
	\renewcommand\arraystretch{1.1}
	\centering
	\small
	%\footnotesize
	%\scriptsize 
	\begin{tabular}{l l c}
		\toprule
		Method   & Ablation operation  & Avg.\\
		\midrule
		Source model only            &---                                              & 59.6\\
		\midrule
		N2DC-EX-no-im &Let $\mathcal{L}_{g_t\circ u_t}^{t}=\beta\mathcal{L}_{ss}^t$ & 70.0\\
		N2DC-EX-no-ss &Let $\mathcal{L}_{g_t\circ u_t}^{t}=\mathcal{L}_{im}^t$      & 72.3\\
		
		N2DC-EX-no-NNH-in-im &Set $\omega_{in}=0$ in Eqn.~\eqref{eqn:im-fusion}  & 72.4\\
		N2DC-EX-no-NNH-in-ss &Set $\eta_{in}=0$ in Eqn.~\eqref{eqn:ss}   & 73.0\\
		
		N2DC-EX-no-fused-pl   &Fix $\lambda_k=1.0$ in Eqn.~\eqref{eqn:fusion-pl}    & 72.9\\
		%N2DC-EX-no-rand-in-pl &Fix $\lambda_k=0.85$ in Eqn.~\eqref{eqn:fusion-pl}   & 73.2\\
		\midrule
		N2DC-EX-no-chain  &Find $\bar{\boldsymbol{h}}_{ih}^t$, ${\boldsymbol{h}}_{ih}^t$ by  $\mathrm{F}(\cdot;\bar{\mathcal{C}})$   & 72.9\\
		N2DC-EX-Ce  &Let $\bar{\mathcal{C}} = \bar{\mathcal{C}}^e$   & \textbf{73.1}\\
		N2DC-EX-Cd  &Let $\bar{\mathcal{C}} = \bar{\mathcal{C}}^d$   & \textbf{73.1}\\
		\midrule
		N2DC-EX          &---                                          & \textbf{73.1}\\
		\hline
	\end{tabular}
\end{table}

Similarly, we carry out all ablation experiments above for N2DC-EX. 
We use similar notations like N2DC,  replacing the 'N2DC' with 'N2DC-EX', to denote these methods. 
Also, we add three other experiments. 
Concretely, to verify the effectiveness of the chain-search method for the home sample detection, we directly find $\bar{\boldsymbol{h}}_{ih}^t$ and ${\boldsymbol{h}}_{ih}^t$ from $\bar{\mathcal{C}}$ using the minimal distance rule presented by $\mathrm{F}(\cdot;\bar{\mathcal{C}})$
(refer to Eqn.~\eqref{eqn:nnh-pl}). We denote this method by N2DC-EX-no-chain.
To verify the effectiveness of the confident group detection strategy using intersection, as done in Eqn.~\eqref{eqn:c}, we let $\bar{\mathcal{C}} = \bar{\mathcal{C}}^e$ and $\bar{\mathcal{C}} = \bar{\mathcal{C}}^d$ respectively and denote the two methods by N2DC-EX-Ce and N2DC-EX-Cd.  
\begin{table}[h]
	\caption{Supplemental experiment results (Avg.\%) for ablation study on Office-31 (OC), Office-Home (OH), and VisDA-C (VC) datasets. The bold means the best result.}
	\label{tab:ablation-add}
	\renewcommand\tabcolsep{10pt}
	\renewcommand\arraystretch{1.1}
	\centering
	\small
	%\footnotesize
	%\scriptsize 
	\begin{tabular}{l c c c}
		\toprule
		Method                 & OC   & OH   & VC \\
		\midrule
		N2DC-EX-Ce             & 89.9  & \textbf{73.1}  & 84.6\\
		N2DC-EX-Cd             & 89.7  & \textbf{73.1}  & 77.8\\
		\midrule
		N2DC-EX                & \textbf{90.0}  & \textbf{73.1}  & \textbf{85.8}\\
		\hline
	\end{tabular}
\end{table}

As reported in Tab.~\ref{tab:ablation} and Tab.~\ref{tab:ablation-ex}, on Office-Home dataset, the performance of all comparison methods decreases by varying degrees as they lack specific algorithm components, compared to the full version, i.e., N2DC and N2DC-EX.
This result indicates that the skills aforementioned are all practical.
At the same time, we see that the full versions have no improvement compared to N2DC-EX-Ce and N2DC-EX-Cd. 
To avoid biased evaluation, Tab.~\ref{tab:ablation-add} gives the supplemental experiment results of the two comparisons on the other two datasets.
On the small Office-31, N2DC-EX surpasses both N2DC-EX-Ce and N2DC-EX-Cd. This advantage of N2DC-EX is more obvious on the large VisDA-C.

%Excluding $\mathcal{L}_{im}^t$ or $\mathcal{L}_{ss}^t$ from the objective function causes considerable performance deterioration over 1.0\%.
%Without introducing NNH, $\mathcal{L}_{im}^t$ is more affected than $\mathcal{L}_{ss}^t$.

\section{Conclusion}\label{sec:con}
Deep clustering is a promising method to address the SAUDA problem because it bypasses the absence of source data and the target data's labels by self-supervised learning.  
However, the individual data-based clustering in the existing DC methods is not a robust process.
Aiming at this weakness, we exploit the constraints hidden in the local geometry between data to encourage robust gathering in this paper. 
To this end, we propose the new semantic constraint SCNNH inspired by a cognitive law named category learning. 
Focusing on this proposed constraint, we develop a new NNH-based DC method that regards SAUDA as a model adaptation. 
In the proposed network, i.e., the target model, we add a geometry construction module to switch the basic clustering unit from the individual data to NNH. 
In the training phase, we initialize the target model with a given source model trained on the labeled source domain.
After this, the target model is self-trained using a new objective building upon NNH.
As for geometry construction, besides the standard version of NNH that we only construct based on spatial information, we also give an advanced implementation of NNH, i.e., SHNNH.
State-of-the-art experiment results on three challenging datasets confirm the effectiveness of our method.

Our method achieves competitive results by only using simple local geometry. 
This implies that the local geometry of data is meaningful for end-to-end DC methods. 
We summarize three possibilities for this phenomenon: 
\textbf{(\romannumeral1)} These local structures are inherent constraints, which have robust features and are easy to maintain in non-linear transformation; 
\textbf{(\romannumeral2)} Compared with other regularizations, the regularization in our method is easy to understand for its obvious geometrical meaning; 
\textbf{(\romannumeral3)} The local structure is usually linear, thus can be easily modeled. 

In our method, the used geometries are all constructed in the Euclidean space. 
This assumption does not always hold. 
For example, manifold learning has proved that data locate on a manifold embedded in a high-dimensional Euclidean space. 
Therefore, our future work will focus on mining geometry with more rich semantic information in more natural data space and form new constraints to boost self-unsupervised learning such as deep clustering.

% or


%\appendices

%\section{}

%\section*{Acknowledgment}

\ifCLASSOPTIONcaptionsoff
  \newpage
\fi



%\bibliographystyle{IEEEtran}
%\begin{thebibliography}{1}


%\end{thebibliography}

{
	\bibliographystyle{IEEEtran}
	\normalem
	\bibliography{bare_jrnl}
}
\vfill


%\newpage



%\vfill


\end{document}